\documentclass[sigconf]{acmart}
\usepackage{booktabs} 
\usepackage{CJK}
\usepackage{subfigure}
\usepackage{sistyle}
\SIthousandsep{,}
\usepackage{makecell}
\usepackage{multirow}
\usepackage{enumitem}
\usepackage{filecontents}
\usepackage{bbm}
\usepackage{algorithm}
\usepackage{algorithmic}
\usepackage{pgfplots}
\usepackage{color}
\usetikzlibrary{patterns}

\copyrightyear{2020}
\acmYear{2020}
\setcopyright{acmcopyright}
\acmConference[CIKM '20]{Proceedings of the 29th ACM International Conference on Information and Knowledge Management}{October
19--23, 2020}{Virtual Event, Ireland}
\acmBooktitle{Proceedings of the 29th ACM International Conference on Information and Knowledge Management (CIKM '20), October 19--23, 2020, Virtual Event, Ireland} 
\acmPrice{15.00}
\acmDOI{10.1145/3340531.3411999}
\acmISBN{978-1-4503-6859-9/20/10}

\makeatletter
\g@addto@macro\normalsize{%
  \abovedisplayskip 1pt plus1pt 
  \belowdisplayskip 1pt plus1pt
  \abovedisplayshortskip  0pt plus1pt%
  \belowdisplayshortskip  0pt plus1pt
}
\makeatother

\definecolor{c1}{RGB}{255,255,157}
\definecolor{c2}{RGB}{190, 235, 159}
\definecolor{c3}{RGB}{94,180,210}

\begin{document}

\title{Query Understanding via Intent Description Generation}
\author{Ruqing Zhang, Jiafeng Guo, Yixing Fan, Yanyan Lan, and Xueqi Cheng}
\affiliation{
  \institution{
    CAS Key Lab of Network Data Science and Technology, Institute of Computing Technology, \\ Chinese Academy of Sciences, Beijing, China\\
    University of Chinese Academy of Sciences, Beijing, China\\} 
}
\email{{zhangruqing,guojiafeng,fanyixing,lanyanyan,cxq}@ict.ac.cn}

\begin{abstract}

Query understanding is a fundamental problem in information retrieval (IR), which has attracted continuous attention through the past decades. 
Many different tasks have been proposed for understanding users' search queries, e.g., query classification or query clustering. 
However, it is not that precise to understand a search query at the intent class/cluster level due to the loss of many detailed information. 
As we may find in many benchmark datasets, e.g., TREC and SemEval, queries are often associated with a detailed description provided by human annotators which clearly describes its intent to help evaluate the relevance of the documents. 
If a system could automatically generate a detailed and precise intent description for a search query, like human annotators, that would indicate much better query understanding has been achieved. 
In this paper, therefore, we propose a novel Query-to-Intent-Description (Q2ID) task for query understanding. 
Unlike those existing ranking tasks which leverage the query and its description to compute the relevance of documents, Q2ID is a reverse task which aims to generate a natural language intent description based on both relevant and irrelevant documents of a given query. 
To address this new task, we propose a novel Contrastive Generation model, namely CtrsGen for short, to generate the intent description by contrasting the relevant documents with the irrelevant documents given a query. 
We demonstrate the effectiveness of our model by comparing with several state-of-the-art generation models on the Q2ID task. 
We discuss the potential usage of such Q2ID technique through an example application.

\end{abstract}

%
%



\maketitle

\section{Introduction}

Query understanding is a key issue in information retrieval (IR), which aims to predict the search intent given a search query.   
Understanding the intent behind a query offers several advantages: 
1) achieving better performance on query reformulation, query refinement, query prediction and query suggestion, and 
2) improving the document relevance modeling based on the search intent of the query.

\begin{table}[t]
\small
\renewcommand{\arraystretch}{1.3}
   \setlength\tabcolsep{3pt}
  \centering
   \caption{An example from the TREC 2004 Robust dataset.} 
  \begin{tabular}{p{7.9cm}} \hline \hline
    \textbf{Query} (Number: 400): Amazon rain forest \\ \hline
     \textbf{Description}: What measures are being taken by local South American authorities to preserve the Amazon tropical rain forest? \\\hline
   \textbf{Document 1} (DOCNO: LA050789-0087): Eight nations of South America's Amazon basin called on wealthy countries to provide money for the preservation of the world's greatest rain forest and for the economic development of the region. At the first summit meeting on the Amazon ... (Relevance: 1) \\
    \textbf{Document 2} (DOCNO: LA060890-0004): Destruction of the world's tropical forests is occurring nearly 50\% faster than the best previous scientific estimates showed ... Tree burning accounts for an estimated 30\% of worldwide total carbon dioxide emissions ... (Relevance: 1) \\  
    \textbf{Document 3} (DOCNO: LA062589-0034):  For Beverly Revness and Janice Tarr ... Scientists also caution that burning the trees amounts to a double-barreled contribution to global warming the so-called ``greenhouse effect.'' Combustion adds pollutants and carbon dioxide to the atmosphere ... (Relevance: 0) \\  \hline \hline
  \end{tabular}
  \label{example}
\end{table}

There have been many related research topics dedicated to query  understanding over the past decades.  
Early works include a huge amount of human analysis and effort to identify the intent of a search query  \cite{shneiderman1997clarifying}. 
Later, many automated query intent analysis, such as query classification and query clustering, have been proposed to understand users' information needs. 
Query classification \cite{broder2002taxonomy,cao2009context,hu2009understanding} aims to classify queries into one or more pre-defined target categories depending on different types of taxonomies. 
However, most existing research on query classification has focused on the coarse-grained understanding of a search query at the intent category level, which may result in the loss of many detailed information. 
Query clustering \cite{beeferman2000agglomerative,wen2002query,hong2016accurate} attempts to discover search topics/intent by mining clusters of queries.  
Nevertheless, it is a little difficult for a human to clearly understand what each cluster represents. 
Hence, it is not that precise to understand a search query at the intent class/cluster level.

As we may find in many relevance ranking benchmark datasets (e.g., TREC\footnote{https://trec.nist.gov/} and SemEval\footnote{https://en.wikipedia.org/wiki/SemEval}), queries are often associated with a detailed description provided by human annotators which clearly describes its intent.  
As shown in Table \ref{example}, given a short query ``Amazon rain forest'' from the TREC 2004 Robust dataset, the description precisely clarifies its search intent ``what are the measures to preserve the Amazon tropical rain forest''. 
Based on the detailed description, the relevance of the documents to a query can be well evaluated (i.e., relevance score 1 denotes \textit{relevant} and relevance score 0 denotes \textit{irrelevant}). 
From this example, we can find that the intent description is a more accurate and informative representation of the query compared with the intent class/cluster proposed in previous works.  
If a system could automatically generate a detailed and precise intent description for a search query, like human annotators did, that would indicate much better query understanding has been achieved.

In this paper, we thus introduce a novel Query-to-Intent-Description (Q2ID) task for query understanding.  
Given a query associated with a set of relevant and irrelevant documents, the Q2ID task aims to generate a natural language intent description, which interprets the information need of the query.    
The Q2ID task can be viewed as a reverse task of those existing ranking tasks. 
Specifically, the Q2ID task generates a description based on both relevant and irrelevant documents of a given query, while existing ranking tasks explicitly utilize the query and its description to compute the relevance of documents. 
Note the Q2ID task is quite different from traditional query-based multi-document summarization tasks which typically do not consider the irrelevant documents. 
To facilitate the study and evaluation of the Q2ID task, we build a  benchmark dataset\footnote{The dataset is available at https://github.com/daqingchong/Q2ID-benchmark-dataset.} based on several public IR collections, i.e., Dynamic Domain Track in TREC, Robust Track in TREC and Task 3 in SemEval.

To address this new task, we introduce a novel Contrastive Generation model, named CtrsGen for short, to generate the intent description by  contrasting the relevant documents with the irrelevant documents of a given query. 
The key idea is that a good intent description should be able to distinguish relevant documents from those irrelevant ones.  
Specifically, our CtrsGen model employs the Seq2Seq framework \cite{bahdanau2014neural}, which has achieved tremendous success in natural text generation.  
In the encoding phase, rather than treat each sentence in relevant documents equally, we introduce a \textit{query-aware encoder attention mechanism} to identify those important sentences that can reveal the essential topics of the query.  
In the decoding phase, we employ a \textit{contrast-based decoder attention mechanism} to adjust the importance of sentences in relevant documents with respect to their similarity to  irrelevant documents.  
In this way, the CtrsGen model could identify those most distinguishing topics based on contrast.

Empirical results on the Q2ID benchmark dataset demonstrate that intent description generation for query understanding is feasible and our proposed method can outperform all the baselines significantly.  
We provide detailed analysis on the proposed model, and conduct case studies to gain better understanding on the learned description.  
Moreover, we discuss the potential usage of such Q2ID technique through an example application. 

\section{Related Work}

To the best of our knowledge, the Query-to-Intent-Description task  is a new task in the IR community. 
In this section, we briefly review three lines of related work, i.e., query understanding, summarization and interpretability.   

\subsection{Query Understanding}

The most closely related query understanding tasks include query classification, query clustering, and query expansion.

\subsubsection{Query Classification}

Query classification has long been studied for understanding the search intent behind the queries which aims to classify search queries into pre-defined categories.
Query classification is quite different from traditional text classification since queries are usually short (e.g., 2-3 terms) and ambiguous \cite{hu2009understanding}. 
Initial studies focuses on the type of the queries based on the information needed by the user and various taxonomies of search intent have been proposed. 
\cite{broder2002taxonomy} firstly classified queries according to their intent into 3 types, i.e., Navigational, Informational and Transactional. 
This trichotomy is the most widely adopted one in automatic query intent classification work probably due to its simplicity and essence.   
Later, many taxonomies based on \cite{broder2002taxonomy} are established \cite{rose2004understanding,baeza2006intention}. 
Since the query is often incomplete and indirect, and the intent is highly subjective, many works are proposed to consider the topic taxonomy for queries. 
Typical topic taxonomies used in literature include that proposed in the KDD Cup 2005 \cite{li2005kdd}, and a manully collected one from AOL \cite{beitzel2005improving}. 
Different methods have been leveraged for this task. 
Some works are focusing to tackle the training data sparsity by introducing an intermediate taxonomy for mapping \cite{shen2005q,kardkovacs2005ferrety}, while some mainly considers the difficulty in representing the short and ambiguous query \cite{beitzel2005improving}.   
Beyond the previous major classification tasks on search queries, there has been research work paying attention to other ``dimensions'', e.g., information type \cite{pasca2001high}, time requirement \cite{jones2007temporal} and geographical location \cite{gravano2003categorizing}.

\subsubsection{Query Clustering}

Query clustering aims to group similar queries together to understand the user's search intent. 
The very early query clustering technique comes from information retrieval studies. 
The key issue is that how to measure the similarity of queries. 
Similarity functions such as cosine similarity or Jaccard similarity were used to measure the distance between two queries. 
Since queries are quite short and ambiguous, these measures suffer from the sparse nature of queries. 
To address this problem, click-through query logs have been mined to yield similar queries.
\cite{beeferman2000agglomerative} first introduced the agglomerative clustering method to discover similar queries using clicked URLs and query logs. 
\cite{wen2002query} analyzed both query contents and click-through bipartite graph, and then applied DBSCAN algorithm to group similar queries. 
\cite{zhang2006mining} presented a method that groups similar queries by analyzing users' sequential search behavior.  
\cite{Radlinski2010InferringQI} combined click and reformulation information to find intent clusters. 
\cite{Duan2012ClickPA} proposed the use of click patterns through a hierarchical clustering algorithm. 
\cite{Qian2013DynamicQI} proposed to dynamically mine query intents from search query logs. 
\cite{meng2013new} calculated query similarity using feature terms extracted from user clicked documents with the help of WordNet. 
\cite{hong2016accurate} quantified query similarity based on the queries' top ranked search results. 
Recently, \cite{kolluru2016query} used word2vec to obtain query representations and then applies Divide Merge Clustering on top of it. 
\cite{rani2019efficient} proposed a new document clustering prototype, where the information is periodically updated to cater to the distributed environment.

\subsubsection{Query Expansion}
Query expansion aims to reformulate the initial query by adding similar terms that help in retrieving more relevant results. 
It was first applied by \cite{maron1960relevance} as a technique for literature indexing and searching in a mechanized library system. 
Early works \cite{voorhees1994query,fang2008re} expanded the initial query terms by analyzing the expansion features, e.g., lexical, semantic and syntactic relationships, from large knowledge resources, e.g., WordNet and ConceptNet. 
Later, some works recognized the expansion features over the whole corpus \cite{zhang2016learning,singh2015context,singh2015co} to create co-relations between terms, while many works utilized user's search logs for expanding original query \cite{cui2002probabilistic,dang2010query}. 
Besides, several works used both positive and negative relevance feedback for query expansion \cite{Peng2009Approximating,Karimzadehgan2011Improving}. 
Recently, natural language generation models are used to automatically expand queries. 
\cite{otsuka2018query} designed a query expansion model on the basis of a neural encoder-decoder model for question answering.  
\cite{lee2018rare} proposed to introduce the conditional generative adversarial network framework to directly generate the related keywords from a given query. 

\subsection{Summarization}

The most closely related summarization tasks include multi-document summarization and query-based multi-document summarization.  

\subsubsection{Multi-Document Summarization}
Multi-document summarization aims to produce summaries from document clusters on the same topic. 
Traditional multi-document summarization models studied in the past are extractive in nature, which try to extract the most important sentences in the document and rearranging them into a new summary \cite{carbonell1998use,erkan2004lexrank,mihalcea2004textrank,steinberger2004text}. 
Recently, with the emergence of neural network models for text generation, a vast majority of the literature on summarization is dedicated to abstractive methods which are largely in the single-document setting. 
Most methods for abstractive text summarization \cite{rush2015neural,chopra2016abstractive,hsu2018unified,song2018structure} are based on the neural encoder-decoder architecture \cite{sutskever2014sequence,bahdanau2014neural}.  
Many recent studies have attempted to adapt encoder-decoder models trained on single-document summarization datasets to multi-document summarization \cite{zhang2018adapting,lebanoff2018adapting}. 
Recently, for assessing sentence redundancy, \cite{cho2019improving} introduced the improved similarity measure inspired by capsule networks and \cite{fabbri2019multi} incorporated MMR into a pointer-generator network for multi-document summarization. 

\subsubsection{Query-based Multi-document Summarization}
Query-based multi-document summarization is the process of automatically generating natural summaries of text documents in the context of a given query.  
An early work for extractive query-based multi-document summarization is presented by \cite{goldstein1999summarizing}, which ranked sentences using a weighted combination of statistical and linguistic features. 
\cite{daume2006bayesian} presented to extract sentences based on the language model, Bayesian model, and graphical model. 
\cite{mohamed2006improving} introduced the graph information to look for relevant sentences.  
\cite{schilder2008fastsum} used the multi-modality manifold-ranking algorithm to extract topic-focused summary from multiple documents. 
Recently, some works employ the encoder-decoder framework to produce the query-based summaries. 
\cite{hasselqvist2017query} trained a pointer-generator model, and \cite{baumel2018query} incorporated relevance into a neural seq2seq models for query-based abstractive summarization.  
\cite{nema2017diversity} introduced a new diversity based attention mechanism to alleviate the problem of repeating phrases. 

\subsection{Interpretability}

Interpretability, in machine learning has been studied under the ill-defined notion of ``the ability to explain or to present in understandable terms to a human'' \cite{Doshi2017Towards}. 
Recently there have been some works on explaining results of a ranking model in IR. 
\cite{Singh2018EXS} utilized a posthoc model agnostic interpretability approach for generating explanations, which are used to answer interpretability questions specific to ranking. 
\cite{Manisha2019LIRME}  explored several sampling methods in generating local explanations for a document scored with respect to a query by an IR model. 
\cite{MAI2020} proposed a model-agnostic approach, which attempts to locally approximate a complex ranker by using a simple ranking model in the term space.
Also, researchers have explored various approaches \cite{Xu2016Learning,Costa2017Automatic,NLE2019SIGIR} towards explainable recommendation systems, which can not only provide users with the recommendation lists, but also intuitive explanations about why these items are recommended.

The Q2ID task introduced in our work is quite different from the above existing tasks.  
Firstly, query classification and query clustering focused on the coarse-grained understanding of queries at the intent class/cluster level, while our Q2ID task provides fine-grained understanding of queries by generating a detailed intent description. 
Secondly, query expansion adds additional similar terms into the initial query, while our Q2ID task generates new sentences as the intent description of a given query. 
Then, the Q2ID task generates the intent description based on the   relevant and irrelevant documents of a given query, while there is no such consideration of irrelevant documents in multi-document summarization or query-based multi-document  summarization. 
Finally, most previous explainable search/recommender systems aim to explain why a single document/product was considered relevant, while our Q2ID task aims to generate an intent description for a given query based on both the relevant and irrelevant documents.


\section{Problem Formalization}

In this section, we introduce the Q2ID task, and describe the benchmark dataset in detail. 

\subsection{Task Description}

Given a query associated with a set of relevant documents and irrelevant documents, the Q2ID task aims to generate a natural language intent description, which precisely interprets the search intent that can help distinguish the relevant documents from the irrelevant documents.

Formally, given a query $q=\{ w_1^q,\dots,w_C^q\}$ with a sequence of $C$ words, a set of $M$ relevant documents $\mathcal{R} = \{D^r_1, \dots, D^r_M\}$ where each relevant document $D^r_m$ is composed of $T_m$ sentences, and a set of $N$ irrelevant documents $\mathcal{I} =\{D^i_1, \dots, D^i_N\}$ where each irrelevant document $D^i_n$ is composed of  $T_n$ sentences, the Q2ID task is to learn a mapping function $g(\cdot)$ to produce an intent description $y=\{w_1^y,\dots,w_Z^y\}$ which contains a sequence of $Z$ words, i.e., 
\begin{equation}
	g(q, \mathcal{R}, \mathcal{I}) = y,
\end{equation}
where $M \ge 1$ and $N \ge 0$. Specifically, the collection of irrelevant documents could be empty while the relevant documents are necessary for intent description generation.

\subsection{Data Construction}

In order to study and evaluate the Q2ID task, we build a benchmark dataset based on the public TREC and SemEval collections.

\begin{itemize}[leftmargin=*]
\item \textbf{TREC} is an ongoing series of workshops focusing on a list of different IR research areas. We utilize the TREC 2015, 2016 and 2017 Dynamic Domain Track\footnote{https://trec.nist.gov/data/dd.html} and TREC 2004 Robust Track\footnote{https://trec.nist.gov/data/robust.html}. 
\item \textbf{SemEval} is an ongoing series of evaluations of computational semantic analysis systems. We utilize the SemEval–2015 Task 3\footnote{http://alt.qcri.org/semeval2015/task3/} and SemEval–2016 Task 3\footnote{http://alt.qcri.org/semeval2016/task3/} for English\footnote{We do not consider the Arabic language and SemEval–2017 \cite{nakov2017semeval} task which reran the four subtasks from SemEval-2016.}. 

\end{itemize}

\begin{table}[t]
\small
\renewcommand{\arraystretch}{1}
 \setlength\tabcolsep{14pt}
  \centering
  \caption{Data statistics: \#s denotes the number of sentences, \#w denotes the number of words, \#r denotes the number of relevant documents, and \#i denotes the number of  irrelevant documents.}
  \begin{tabular}{l c } \hline \hline
   Query & 5358  \\ 
   Query: avg \#w & 4.7 \\
   Query: avg \#r & 10.8 \\
   Query: avg \#i &  65.5 \\ 
   \hline
   Relevant documents &  \num{62916}\\
   Irrelevant documents & \num{196787}\\
   Relevant documents: avg \#s &  20.7 \\ 
   Irrelevant documents: avg \#s &  23.2 \\  \hline
   Intent Description: avg \#w & 31.0 \\ 
    \hline \hline
  \end{tabular} 
  \label{stat}
\end{table}

In these IR collections, queries are associated with a human-written detailed description, and documents are annotated with multi-graded relevance labels indicating a varying degree of match with the query intent/information.   
As a primary study on the intent description generation, here, we convert multi-graded relevance labels into binary relevance labels, indicating a document is relevant or irrelevant to a query. 
We leave the Q2ID task over multi-graded relevance labels as our future work. 

With regard to different tasks, we define the binary relevance labels for documents as follows:

\begin{itemize}[leftmargin=*] 
	\item For Dynamic Domain Track, each passage is graded at a scale of $0 \sim 4$ according to the relevance to a query (i.e., subtopic). We treat passages with rating of \textit{1} or \textit{higher} as relevant documents, and passages with rating of \textit{0} as irrelevant documents. 
	\item For Robust Track, the judgment of each document is on a   three-way scale of relevance to a query (i.e., topic). We treat documents annotated as \textit{highly relevant} and \textit{relevant} as relevant documents, and documents annotated as \textit{not relevant} as irrelevant documents.
	\item For Task 3 in both SemEval-2015 and SemEval-2016, each answer is classified as \textit{good}, \textit{bad}, or \textit{potential} according to the relevance to a query (i.e., question). We treat answers annotated as \textit{good} as relevant documents, and the rest as irrelevant documents. 
\end{itemize}

In this way, we obtain the $<$query, relevant documents, irrelevant documents, intent description$>$ quadruples, as ground-truth data for training/validation/testing.   
Queries with no relevant documents are removed. 
Table \ref{stat} shows the overall statistics of our Q2ID benchmark dataset. 
Note there are $743$ queries without corresponding irrelevant documents in our dataset.

\begin{figure*}[t]
	\centering
		\includegraphics[scale=0.28]{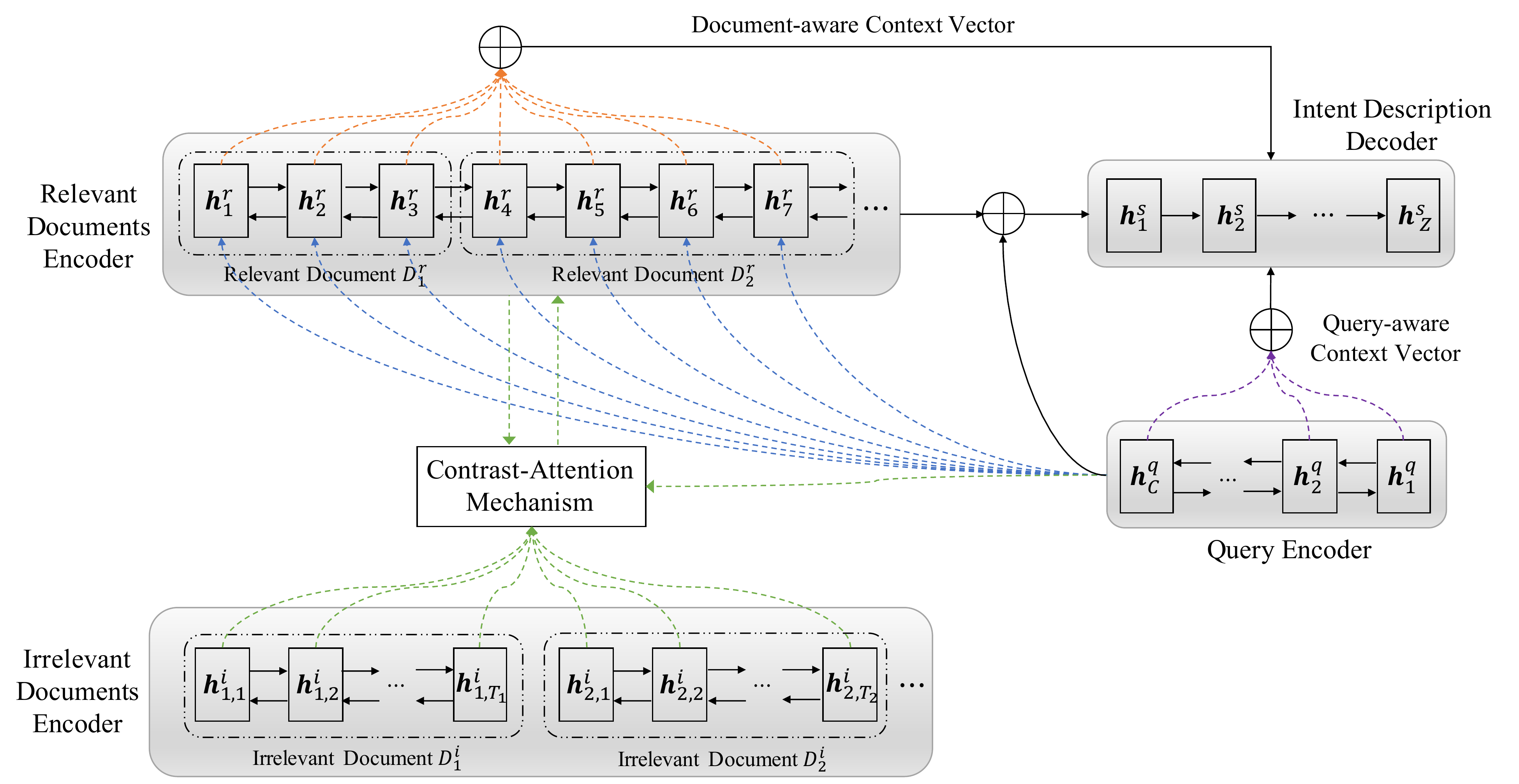}
		\caption{The overall architecture of contrastive generation model (CtrsGen).}
\label{fig:CtrsGen}  
\end{figure*}

\section{Our Approach}

In this section, we introduce our proposed approach for the Q2ID task in detail. We first give an overview of the model architecture, and then describe each component of our model as well as the learning procedure.

\subsection{Overview}

Without loss of generality, the Q2ID task needs to distill the salient information from the relevant documents and remove   unrelated information from the irrelevant documents with respect to a query.   
For example, as shown in Table \ref{example}, given the keyword query ``Amazon rain forest'', there might be different underlying users' search intents, like ``location of Amazon rain forest'', ``age of Amazon rain forest'' or ``protection of Amazon rain forest''. 
From the relevant document 1 and 2, we can find several topics, e.g., ``the measures to preserve Amazon rain forest'' and ``the effect of tree burning''. 
From the irrelevant document 3, it mainly talks about the topic of ``the effect of tree burning''.  
By contrasting the relevant documents with the irrelevant  documents, we find that the query ``Amazon rain forest''  aims to search for ``the measures to preserve Amazon rain forest''  instead of ``the effect of tree burning''. 
Therefore, in this work, we formulate the Q2ID task as a novel contrastive generation problem and introduce a CtrsGen model to solve it.

Basically, our CtrsGen model contains the following four components: 
1) Query Encoder, to obtain the representation of a query;    
2) Relevant Documents Encoder, to obtain the representations of  relevant documents by finding common salient topics through the consideration of the semantic interaction between relevant documents;   
3) Irrelevant Documents Encoder, to obtain the representations of irrelevant documents by modeling each irrelevant document separately;   
4) Intent Description Decoder, to generate the intent description by contrasting the relevant documents with the irrelevant documents given a query.  
The overall architecture is depicted in Figure \ref{fig:CtrsGen} and we will detail our model as follows.

\subsection{Query Encoder}

The goal of the query encoder is to map the input query to a vector representation. 
Specifically, we use a bi-directional GRU \cite{cho2014learning} as the query encoder.   
Each word $w_c^q$ in the query $q$ is firstly represented by its semantic representation $\textbf{e}_c^q$ as the input of the encoder. 
Then, the query encoder represents the query $q$ as a series of hidden vectors $\{\textbf{h}_c^q\}_{c=1}^C$ modeling the sequence from both forward and backward directions.
Finally, we use the concatenated forward and backward hidden state as the query representation $\textbf{x}^q$. 

\subsection{Relevant Documents Encoder}

Generally, the relevant documents encoder takes in the input relevant documents, and encodes them into a series of hidden representations. 
Relevant documents are assumed to share similar underlying topics  since they are all related to the specific information need behind a query to different extent. 
To achieve this purpose, we propose to encode the relevant documents together by concatenating the source $M$ relevant documents into a single relevant mega-document $D_r=\{s_1^r, \dots, s_U^r\}$, with $U$ (i.e., $U=\sum_{m=1}^M T_m$) sentences where each sentence $s_u^r$ contains $L_u^r$ words. 

We adopt a hierarchical encoder framework, where a word encoder  encodes the words of a sentence $s_u^r$, and a sentence encoder encodes the sentences of a relevant mega-document $D_r$. 
We use a bi-directional GRU as both the word and sentence encoder. 
Firstly, we obtain the hidden state $\textbf{h}_{u,l}^r$ for a given word $w_{u,l}^r$ in each sentence $s_u^r$ by concatenating the forward and backward hidden states of the word encoder. 
Then, we concatenate the last hidden states of the forward and backward passes as the embedding representation $\textbf{e}_u^r$ of the sentence $s_u^r$.
A sentence encoder is used to sequentially receive the embeddings of sentences $\{\textbf{e}_u^r\}_{u=1}^{U}$ and the hidden representation $\textbf{h}_u^r$ of each sentence $s_u^r$ is given by concatenating the forward and backward hidden states of the sentence encoder.

Different from previous simple methods \cite{hasselqvist2017query,zhang2018adapting} which directly concatenate or aggregate the hidden states of sentences to obtain the relevant mega-document representation, we further employ a \textit{query-aware encoder attention mechanism} as follows, to aggregate the sentence representations according to their importance to obtain a good relevant mega-document representation.

\textbf{Query-aware Encoder Attention Mechanism} The key idea of the query-aware encoder attention mechanism is to identify those important sentences in the relevant documents that can reveal the essential topics of the query rather than treat each sentence equally. 
Different from pre-defined relevance scores using unigram overlap between query and sentences \cite{baumel2018query}, we leverage  the query representation $\textbf{x}^q$ to estimate the importance of each sentence through attention.  

Specifically, the importance score (or weight) $\gamma_u^r$ of each sentence $s_u^r$ is given by ${\gamma}_u^r = softmax(\textbf{x}^q \cdot \textbf{Q} \cdot \textbf{h}_u^r)$, where $\textbf{Q}$ is a parameter matrix to be learned and the softmax function ensures all the weights sum up to 1. Finally, we obtain the representation of the relevant mega-document $\textbf{x}^r$ by using the weighted sums of hidden states $\{\textbf{h}_1^r,\cdots,\textbf{h}_U^r \}$, i.e., $\textbf{x}^r = \sum_{u=1}^U {\gamma}_u^r \textbf{h}_u^r.$

\subsection{Irrelevant Documents Encoder}

Different from the relevant documents encoder which encodes relevant documents together, the irrelevant documents encoder processes each irrelevant document separately.   
The key idea is that while relevant documents could be similar to each other, there might be quite different ways for documents to be irrelevant to a query. Therefore, it is unreasonable to take the semantic interaction between irrelevant documents into consideration.

For each irrelevant document $D_n^i=\{s_{n,1}^{i}, \dots, s_{n,T_n}^{i}\}$ where each sentence $s_{n,t}^{i}$ contains $K_{n,t}^{i}$ words, we adopt a hierarchical encoder framework similar with that in relevant documents encoder. 
Thus, we can obtain the hidden representation of each word $w_{n,t,k}^i$ in each sentence $s_{n,t}^{i}$, i.e., $\{\textbf{h}_{n,t,k}^i\}_{k=1}^{K_{n,t}^{i}}$, 
the embedding representation of each sentence $s_{n,t}^{i}$ in each irrelevant document $D_n^i$, i.e., $\{\textbf{e}_{n,t}^{i}\}_{t=1}^{T_n}$, 
and the hidden representation of each sentence $s_{n,t}^{i}$, i.e., $\{\textbf{h}_{n,t}^{i}\}_{t=1}^{T_n}$.  
Finally, we  obtain all the sentence hidden representations in $N$ irrelevant documents, i.e., $\{\textbf{h}_{n,t}^{i}\}_{n=1,t=1}^{N,T_n}$.

\subsection{Intent Description Decoder}

The intent description decoder is responsible for producing the  intent description given the representations of the query, relevant documents and irrelevant documents.  
To generate the intent description $y$, we employ 1) a \textit{query-aware decoder attention mechanism}, which maintains a query-aware context vector to make sure more important content in the query is attended, and 2) a \textit{contrast-based decoder attention mechanism}, which maintains a document-aware context vector for description generation to distinguish relevant documents from those irrelevant ones with respect to a query.

Specifically, the query-aware context vector $\textbf{c}_{z}^q$ and the document-aware context vector $\textbf{c}_{z}^d$ are provided as extra inputs to derive the hidden state $\textbf{h}_{z}^{s}$ of the $z$-th word $w_z^y$ in an intent description and later the probability distribution for choosing the word $w_z^y$. 

Concretely, $\textbf{h}_{z}^{s}$ is defined as,  
\begin{equation}
\label{h_ts}
\textbf{h}_{z}^{s} = f_s(w_{z-1}^y,\textbf{h}_{z-1}^{s},\textbf{c}_{z}^q,\textbf{c}_{z}^d),
\end{equation}
where $f_s$ is a GRU unit, $w_{z-1}^y$ is the predicted word from vocabulary at $z-1$-th step when decoding the intent description $y$. 

The initial hidden state of the decoder is defined as the weighted sums of query and relevant mega-document representations, 
\begin{equation}
\textbf{h}_{0}^{s} = \textbf{W}_q \textbf{x}^q + \textbf{W}_r \textbf{x}^r,
\end{equation}
where $\textbf{W}_q$ and $\textbf{W}_r$ are learned parameters.

The probability for choosing the word $w_{z}^y$ is defined as,  
\begin{equation}
\label{word_choose}
p(w_{z}^y|w_{<z}^y,q,\mathcal{R}, \mathcal{I}) = f_g(w_{z-1}^y, \textbf{h}_{z}^{s},\textbf{c}_{z}^q, \textbf{c}_{z}^d),
\end{equation}
where $f_g$ is a nonlinear function that computes the probability vector for all legal output words at each output time. We now describe the specific mechanism in the follows.

\subsubsection{Query-aware Decoder Attention Mechanism}

The key idea of the query-aware decoder attention mechanism is to make the generation of a description focusing on the query. 
The description should contain as much information relevant to the query as possible.  
We maintain a query-aware context vector $\textbf{c}_{z}^q$ for generating the $z$-th word $w_z^y$ in the description. 
Specifically, $\textbf{c}_{z}^q$ is a weighted sum of the hidden representations of all the words in the query $q$,   
\begin{equation}
\label{equ:section-aware}
\textbf{c}_{z}^q = \sum_{c=1}^{C}\alpha_{z,c}^q \textbf{h}_{c}^q,
\end{equation}
where $\alpha_{z,c}^q$ indicates how much the $c$-th word $w_c^q$ from the query $q$ contributes to generating the $z$-th word in the intent description $y$, and is usually computed as, 
\begin{equation}
\label{query-att}
\alpha_{z,c}^q = softmax(\textbf{h}_{c}^q \cdot \textbf{W}_1 \cdot \textbf{h}_{z-1}^s),
\end{equation}
where $\textbf{W}_1$ is a learned parameter.

\subsubsection{Contrast-based Decoder Attention Mechanism}

The intent description should cover topics in relevant documents and eliminate topics in irrelevant documents of a given query.  
To achieve this purpose, we introduce a contrast-based decoder attention mechanism between relevant documents and irrelevant documents with respect to a query.  

Specifically, the contrast-based decoder attention mechanism contains two steps: 1) to compute the sentence-level attention weights in the relevant documents when decoding the intent description, 2) to compute the the contrast scores to adjust the sentence-level attention weights in the relevant documents.
\begin{itemize}[leftmargin=*]

\item \textbf{Sentence-level Attention Weight} Firstly, we compute the attention weight $\alpha_{z,u}^r$ of each sentence $s_u^r$ in relevant mega-document $D_r$, which indicates how much each   sentence $s_u^r$ contributes to generating the $z$-th word $w_z^y$ in the intent description $y$, 
\begin{equation}
\label{relevant-att}
	\alpha_{z,u}^r = softmax(v^T \tanh (\textbf{W}_2 \textbf{h}_{u}^r + \textbf{W}_3 \textbf{c}_{z}^q + \textbf{W}_4 \textbf{h}_{z-1}^s)), 
\end{equation}
where $\textbf{W}_2$, $\textbf{W}_3$ and $\textbf{W}_4$ are learned parameters. Specifically, the query-aware context vector $\textbf{c}_{z}^q$ is used to incorporate query relevance into the focus on the relevant documents. 

\item \textbf{Contrast Score} Then, we compute the contrast score for each sentence $s_u^r \in D_r$, to omit the information similar with the irrelevant documents. 
The contrast score $\beta_{z,u}^r$ for generating the $z$-th word in the description is defined as, 
\begin{equation}
\label{equ:saliency}
	\hat{\beta}_{z,u}^r = \lambda Sim(s_u^r,y_{<z}) - (1- \lambda ) \max_{s_{n,t}^i \in \mathcal{I}} Sim(s_u^r, s_{n,t}^i),
\end{equation}
\begin{equation}
\beta_{z,u}^r = softmax(\hat{\beta}_{z,u}^r),
\end{equation}
where $\lambda$ is a balancing factor. The two similarity functions are defined as,

\begin{itemize}
\item The similarity function $Sim(s_u^r, y_{<z})$ between each sentence $s_u^r$ in the relevant mega-document and current generated description $y_{< z}$ is defined as, 
\begin{equation}
Sim(s_u^r,y_{<z}) = \textbf{h}_u^r \cdot \textbf{W}_5 \cdot \textbf{h}_{z-1}^s,	
\end{equation}
where $\textbf{W}_5$ is a learned parameter.

\item The similarity function $Sim(s_u^r, s_{n,t}^i)$ between each sentence $s_u^r$ in the relevant mega-document and each sentence  $s_{n,t}^i$ in the irrelevant documents is defined as, 
\begin{equation}
	Sim(s_u^r, s_{n,t}^i) = softmax(\tanh (\textbf{h}_{u}^r \cdot  \textbf{W}_6 \cdot \textbf{h}_{n,t}^{i}),
\end{equation}
where $\textbf{W}_6$ is a learned parameter. 	
\end{itemize}

Finally, we maintain a document-aware context vector $\textbf{c}_{z}^d$ for generating the $z$-th word $w_z^y$ in the description $y$, 
\begin{equation}
\label{equ:section-aware}
\textbf{c}_{z}^d = \sum_{u=1}^{U} \beta_{z,u}^r  \alpha_{z,u}^r \textbf{h}_{u}^r.
\end{equation}

\end{itemize}

\subsection{Model Learning}

We employ maximum likelihood estimation (MLE) to learn our CtrsGen model in an end-to-end way. Specifically, the training objective is a probability over the training corpus $\mathcal{D}$, 
\begin{equation}
\arg \max_{\theta}  \sum_{(q,\mathcal{R}, \mathcal{I},y) \in \mathcal{D}} \log p(y|q,\mathcal{R}, \mathcal{I};\theta).
\end{equation}

We apply stochastic gradient decent method Adam \cite{kingma2014adam} to learn the model parameters $\theta$.

\section{Experiments}

In this section, we conduct experiments to verify the effectiveness of our proposed model.

\subsection{Experimental Settings}

To evaluate the performance of our model, we conducted experiments on our Q2ID benchmark dataset. 
In preprocessing, all the words in queries, documents and descriptions are white-space tokenized and lower-cased, and pure digit words and non-English characters are removed. 
We randomly divide the $5358$ queries into training(\num{5000})/validation(100)/test(258) query sets.

  

We keep the \num{120000} most frequently occurring words in our experiments.  
All the other words outside the vocabularies are replaced by a special token $<$UNK$>$ symbol. 
We implement our model in Tensorflow.   
Specifically, we use one layer of bi-directional GRU for each encoder and uni-directional GRU for decoder, with the GRU hidden unit size set as 256 in both the encoder and decoder.  
The dimension of word embeddings is 300. We use pretrained word2vec vectors trained on the same corpus to initialize the word embeddings, and the word embeddings will be further fine-tuned during training. 
The learning rate of Adam algorithm is set as 0.0005. 
The learnable parameters (e.g., the parameters $\textbf{W}_q$ and $\textbf{W}_1$) are uniformly initialized in the range $[-0.1, 0.1]$. 
The mini-batch size for the update is set as 16. We clip the gradient when its norm exceeds 5. 
We set the $\lambda = 0.5$ to calculate the contrast score in Equ.  \ref{equ:saliency}. 
All the hyper-parameters are tuned on the validation set. 

\subsection{Baselines}

\subsubsection{Model Variants}

Here, we firstly employ some variants of our  model by removing some components and adopting different model architectures.
\begin{itemize}[leftmargin=*]
\item \textbf{CtrsGen$_\text{-I}$} removes the input irrelevant documents, and only considers the effect of the queries and relevant documents. 
\item \textbf{CtrsGen$_\text{-I+Con}$} is similar with CtrsGen$_\text{-I}$, but considers the decoder attention over the concatenation of the hidden states of the queries and relevant documents. 
\item \textbf{CtrsGen$_\text{-Q}$} removes the query, and only considers the effect of the relevant and irrelevant documents.    
\item \textbf{CtrsGen$_\text{-Q-I}$} removes the query and irrelevant documents, and only considers the effect of the relevant documents.  
\item \textbf{CtrsGen$_\text{IrCon}$} concatenates the  irrelevant documents and encodes them as a single input. 
\item \textbf{CtrsGen$_\text{ReIrCon}$} concatenates the relevant and irrelevant documents into a single input as the relevant mega-document.

\end{itemize}

\subsubsection{Extractive Models}
We also apply extractive summarization models to extract a sentence from the relevant documents as the intent description.

\begin{itemize}[leftmargin=*]
\item \textbf{LSA} \cite{steinberger2004text} applys Singular Value Decomposition (SVD) to pick a representative sentence.
\item \textbf{TextRank} \cite{mihalcea2004textrank} is a graph-based method inspired by the PageRank algorithm.
\item \textbf{LexRank} \cite{erkan2004lexrank} is also a graph-based method inspired by the PageRank algorithm. The difference with TextRank is to use different methods to calculate the similarity between two sentences. 
\end{itemize}

\subsubsection{Abstractive Models}

Additionally, we consider neural abstractive models, including query-independent and query-dependent abstractive summarization models, to illustrate how well these systems perform on the Q2ID task.  
\begin{itemize}[leftmargin=*]

\item \textbf{Query-independent abstractive summarization models} based on the relevant documents include,  

\begin{itemize}[leftmargin=*]
\item \textbf{ABS} \cite{rush2015neural} is the attention bag-of-words encoder based sentence summarization model.  
\item \textbf{Extract+Rewrite} \cite{song2018structure} firstly scores sentences using LexRank and then generates a title-like summary. 
\end{itemize}

\item \textbf{Query-dependent abstractive summarization models} based on the query and relevant documents include, 

\begin{itemize}[leftmargin=*]
\item  \textbf{PG} \cite{hasselqvist2017query} employs a pointer-generator model for query-based summarization. 
\item \textbf{RSA} \cite{baumel2018query} incorporates query relevance into the seq2seq framework for query-based summarization. 
\item  \textbf{SD$_2$} \cite{nema2017diversity} introduces a diversification mechanism in query-based summarization for handling the duplication problem.
\end{itemize}

\end{itemize}

\subsection{Evaluation Methodologies}
\label{eva}

We use both automatic and human evaluation to measure the quality of intent descriptions generated by our model and the baselines.

For automatic evaluation, following the previous studies \cite{fabbri2019multi,lebanoff2018adapting,rush2015neural}, we adopt the widely used automatic metric Rouge \cite{lin2004rouge} to evaluate n-grams of the generated intent descriptions with gold-standard descriptions as references. We report recall results on Rouge-1, Rouge-2 and Rouge-L. 

For human evaluation, we consider two evaluation metrics: 1) Naturalness, which indicates whether the intent description is grammatically correct and fluent; and 2) Reasonableness, which measures the semantic similarity between generated intent descriptions and the golden baseline descriptions. We asked three professional native speakers to rate the 258 test quadruples in terms of the metrics mentioned above on a 1$\sim$5 scale (5 for the best).

\begin{table}[t]\centering
 \renewcommand{\arraystretch}{0.95}
 \setlength\tabcolsep{10pt}
  \caption{Ablation analysis of our CtrsGen model with its variants under the automatic evaluation (\%). Two-tailed t-tests demonstrate the improvements of CtrsGen to the variants are statistically significant ($^{\ddag}$ indicates $\text{p-value} < 0.01$).}
  \begin{tabular}{l |c c c}\hline  \hline
     Model & Rouge-1 & Rouge-2  & Rouge-L  \\\hline
     CtrsGen$_\text{-I}$ & 23.31 & 4.05 & 19.03 \\
     CtrsGen$_\text{-I+Con}$  & 23.26  & 4.03 & 19.02 \\
     CtrsGen$_\text{-Q}$  & 23.07 & 3.94 & 18.61 \\
     CtrsGen$_\text{-Q-I}$  & 22.55 & 3.59 & 17.25 \\
     CtrsGen$_\text{IrCon}$ &  24.19 & 4.51  &  19.43 \\
     CtrsGen$_\text{ReIrCon}$ & 22.62 & 3.60 & 17.14 \\\hline
     CtrsGen  &  \textbf{24.76}$^{\ddag}$ & \textbf{4.62} &  \textbf{20.21}$^{\ddag}$ \\ 	
    \hline \hline
    \end{tabular}
   
  \label{tab:abl}
\end{table}

\begin{table}[t]\centering
 \renewcommand{\arraystretch}{0.95}
 \setlength\tabcolsep{9pt}
  \caption{Comparisons between our CtrsGen and the baselines under the automatic evaluation (\%).}
  \begin{tabular}{l |c c c}\hline  \hline
     Model & Rouge-1 & Rouge-2  & Rouge-L  \\\hline      
      LexRank & 13.73 & 1.74 & 10.92 \\ 
      LSA & 18.49 & 2.05 & 14.50 \\ 
      TextRank & 20.15 & 2.95 & 16.43 \\\hline 
      ABS & 17.21 & 1.23 & 12.83  \\
      Extract+Rewrite & 18.85  & 2.49 &  15.45 \\
      PG & 20.81 & 3.04 & 17.25 \\ 
      RSA & 19.65 & 2.39 & 16.17 \\ 
      SD$_2$ & 21.37 & 3.25 & 18.49\\       \hline
    CtrsGen & \textbf{24.76} & \textbf{4.62} &  \textbf{20.21} \\   	
     \hline \hline
    \end{tabular}
   
  \label{tab:Com}
\end{table}

\subsection{Ablation Analysis}

We conduct ablation analysis to investigate the effect of proposed mechanisms in our CtrsGen model. 
As shown in Table \ref{tab:abl}, we can find that: 
(1) By removing the irrelevant documents, the performance of \textit{CtrsGen$_\text{-I}$} and \textit{CtrsGen$_\text{-I+Con}$} has a significant drop as compared with \textit{CtrsGen}. The results indicate that contrasting the relevant documents with the irrelevant documents does help generate the intent description better. 
(2) \textit{CtrsGen$_\text{-Q}$} performs worse than \textit{CtrsGen$_\text{-I}$}, showing that the information in the query has much bigger impact than that in irrelevant documents for extracting salient information for intent description generation.  
(3) \textit{CtrsGen$_\text{IrCon}$} performs worse than \textit{CtrsGen}. 
The reason might be that it tends to bring noisy information when considering the interaction between irrelevant documents. 
(4) The performance of \textit{CtrsGen$_\text{ReIrCon}$} is relatively poor, indicating that considering all judged documents as relevant tends to bring noisy information that may hurt the intent description generation. 
(5) \textit{CtrsGen$_\text{-Q-I}$} gives the worst performance, indicating that traditional multi-document summarization model without considering the query and irrelevant documents is not suitable for the Q2ID task. 
(6) By including all the mechanisms, \textit{CtrsGen} achieves the best performance in terms of all the evaluation metrics.

\subsection{Baseline Comparison}

The performance comparisons between our model and the baselines are shown in Table \ref{tab:Com}. We have the following observations: 
(1) The abstractive methods generally outperform the extractive methods, since those extractive methods are unsupervised in nature.  
(2) The query-independent abstractive summarization models (i.e., \textit{ABS} and \textit{Extract+Rewrite}) perform worse than the query-dependent abstractive summarization models (i.e., \textit{PG}, \textit{RSA} and \textit{SD$_2$}), showing that it is necessary to generate the intent description with the guidance of the query. 
(3) By introducing a diversification mechanism for solving the duplication problem, \textit{SD$_2$} improves Rouge scores when compared to \textit{PG} and \textit{RSA}. 
(4) As compared with the best-performing baseline \textit{SD$_2$}, the relative improvement of \textit{CtrsGen} over \textit{SD$_2$} is about 42.1\% in terms of Rouge-2. 
(5) Our \textit{CtrsGen} model can outperform all the baselines significantly (p-value $<$ 0.01), demonstrating the effectiveness of the contrastive generation idea.

Table \ref{table:human} shows the results of the human evaluation. We can see that our \textit{CtrsGen} outperforms the best performing baseline \textit{SD$_2$} in all evaluation metrics. 
The results imply that our model can generate fluent and grammatically correct intent descriptions (i.e., Naturalness) which better interpret the search intent behind the queries (i.e., Reasonableness) than the baseline \textit{SD$_2$}.

\subsection{Breakdown Analysis}

Beyond above overall performance analysis, we also take some breakdown analysis for the Q2ID task. 

\begin{table}[t]
\renewcommand{\arraystretch}{0.95}
 \setlength\tabcolsep{6pt}
  \centering
   \caption{Results on the human evaluation. Best\% is the ratio of the best score in the two metrics.}
  \begin{tabular}{l c c c} \hline \hline
      & Naturalness & Reasonableness & Best\% \\ \hline
    $\text{SD}_2$& 3.14 & 1.73 & 17.46  \\
    $CtrsGen$& \textbf{3.46} & \textbf{2.86} & \textbf{34.63} \\\hline
    Human & 4.72 & 4.21 & 70.42  \\
    \hline \hline
  \end{tabular}
  \label{table:human}
\end{table}

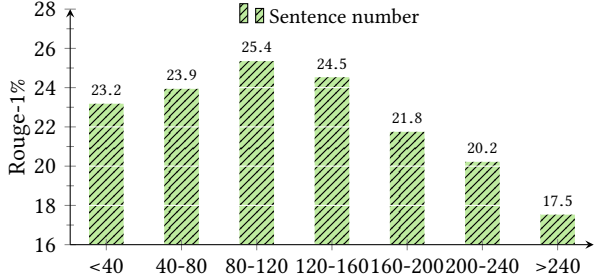
\begin{figure}[t]
\centering 
\begin{tikzpicture}
 \begin{axis}[
  ybar,
  axis on top,
        height=.265\textwidth,
        width=0.48\textwidth,
        bar width=0.46cm,
        ytick={16,18,20,22,24,26,28},
        ymajorgrids, tick align=inside,
        major grid style={draw=white},
        minor y tick num={1},
        enlarge y limits={value=.1,upper},
        ymin=16, ymax=28,
        axis lines=left,
        enlarge x limits=0.08,
        legend style={
            at={(0.5,1.05)},
            font=\small,
            anchor=north,
            draw=none,
            legend columns=-0.5,
            /tikz/every even column/.append style={column sep=0.2cm}
        },
        ylabel={Rouge-1\%},
        ylabel style={
            anchor=south,
            at={(ticklabel* cs:0.5)},
            yshift=-23pt
        },
        symbolic x coords={
           <40,40-80,80-120,120-160,160-200,200-240,>240},
       xtick=data,
       nodes near coords={
        \pgfmathprintnumber[fixed zerofill,precision=1]{\pgfplotspointmeta}
       },
       every node near coord/.append style={anchor=south, font=\fontsize{6pt}{4pt}\selectfont},
 ]
 \addplot [draw=none, fill=c2, postaction={pattern=north east lines}] coordinates {
      (<40,23.18)
      (40-80, 23.94)
      (80-120, 25.35)
      (120-160, 24.53)
      (160-200, 21.75)
      (200-240, 20.22)
      (>240,17.53)
      };
      \legend{Sentence number}
 \end{axis}

\end{tikzpicture}
\caption{Performance comparison of CtrsGen for different sentence numbers of relevant documents.}
\label{fig:length}
\end{figure}

\subsubsection{Analysis on Query Types}

There exist two query types in our Q2ID dataset, i.e., natural language questions from SemEval and keyword-based queries from TREC. 
There are 103 keyword queries and 155 natural language questions respectively in the test dataset. 
Here, we analyze the generated intent description for different query types from our \textit{CtrsGen} model. 
As shown in Table \ref{tab:type}, we can see that \textit{CtrsGen} model for questions perform better than that for keywords.  
The major reason might be that the natural language questions are usually long and thus bring more key information needs, which could help capture the search intent better.


\subsubsection{Analysis on Relevant Document Length}

We also analyze the effect of relevant document length for intent description generation.  
Since the relevant documents are concatenated into a single relevant  mega-document in our \textit{CtrsGen} model, we depict the histogram of Rouge-1 results over different sentence numbers of relevant mega-documents. 
The results are shown in Figure \ref{fig:length}.   
For the test dataset, the average sentence number of relevant mega-documents is 203. 
From the results, we can observe that by introducing less than 80  or more than 120 relevant documents, our model tends to bring insufficient information or noisy information that may hurt the intent description generation.  

\begin{table}[t]\centering
 \renewcommand{\arraystretch}{0.95}
 \setlength\tabcolsep{10pt}
  \caption{Performance comparison of our CtrsGen model for different query types (\%).}
  \begin{tabular}{l |c c c}\hline  \hline
     Query Type & Rouge-1 & Rouge-2  & Rouge-L  \\\hline      
      Question & 24.99 & 4.99 & 21.35 \\ 
      Keyword & 23.12 & 4.15 & 19.56 \\   	
     \hline \hline
    \end{tabular}
   
  \label{tab:type}
\end{table}

\begin{table}[t]\centering
 \renewcommand{\arraystretch}{0.95}
 \setlength\tabcolsep{8pt}
  \caption{Performance comparison of our CtrsGen model for queries with and without irrelevant documents (\%).}
  \begin{tabular}{l |c c c}\hline  \hline
     Irrelevant Documents & Rouge-1 & Rouge-2  & Rouge-L  \\\hline      
      With &  25.21 & 4.73  & 20.44 \\ 
      Without & 24.03 & 4.14 & 19.12 \\   	
     \hline \hline
    \end{tabular}
   
  \label{tab:exitst}
\end{table}

\begin{table}[t]
\small
   \setlength\tabcolsep{2pt}
   \renewcommand{\arraystretch}{0.9}
  \centering
   \caption{An example from the test Q2ID data.}
  \begin{tabular}{p{8.2cm}}
  \hline \hline
  \textbf{Query}: Community Care Centers \\\hline
  \textbf{Relevant Documents}: ($D_1^p$) Another issue the county faced was Ebola's spread in community care centers. Community care centers are part of the strategy to combat the ongoing Ebola epidemic ... ($D_2^p$) Centre provides an alternative to Ebola Treatment Units where residents can seek ... $D_{38}^p$  \\\midrule
   \textbf{Irrelevant Documents}: ($D_1^n$) Liberia is actively considering the concept of Community Care Centers (CCCs) smaller 10-20 bed units located in ``hot spot'' ... ($D_2^n$) Save the Children is constructing and operating Ebola Community Care Centers to provide ``close to the community'' care ...  \\  
 \midrule 
   \textbf{Intent Description}: \\
   \textbf{Ground Truth}: Describe the new initiative, and also the controversy, surrounding the opening of Community Care Centers throughout West Africa and how they will aid in combating the spread of Ebola. \\
    \textbf{SD$_2$}: Community Care Centers is constructed to provide services and care for Ebola cases.  \\
   \textbf{CtrsGen}: World Health Organization establishes Community Care Centers in Africa to combat Ebola epidemic. What measures are taken to prevent the Ebola's spread. \\ 
 \hline  \hline
  \end{tabular}
   \label{Tab:case}
\end{table}

\begin{figure*}[t]
	\centering
		\includegraphics[scale=0.33]{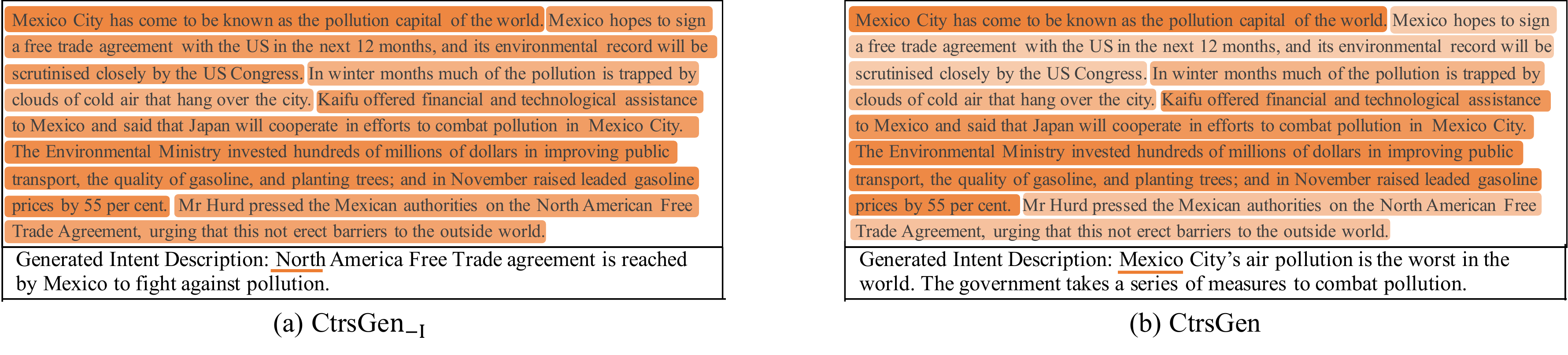}
		\caption{(a) and (b) is the heatmap of the sentence-level decoder attention weights in relevant documents for generating the first word in the description, given by \textit{CtrsGen$_\text{-I}$} and \textit{CtrsGen} respectively. Deeper shading denotes higher value.}
\label{fig:heatmap}  
\end{figure*}

\subsubsection{Analysis on Irrelevant Documents Existence}

To further analyze the effect of irrelevant documents, we conduct a comparison of the generated intent descriptions from our \textit{CtrsGen} model for queries with and without irrelevant documents. 
There are 223 and 35 queries with and without irrelevant documents respectively in the test dataset,  
As shown in Table \ref{tab:exitst}, we can find that the performance of \textit{CtrsGen} for queries with irrelevant documents is better than that for queries without irrelevant documents.  
These results again demonstrate the effectiveness of  contrasting  the relevant documents with the irrelevant documents of a given query.

\subsection{Case Study}

To better understand how different models perform, we show the generated intent description from our \textit{CtrsGen} model as well as that from the best baseline model \textit{SD$_2$}. 
We take one query ``Community Care Centers'' from the test data as an example. 
Due to the limited space, we only show some key sentences. 
As shown in Table \ref{Tab:case}, we can see that without the consideration of irrelevant documents, \textit{SD$_2$} focuses on the ``services for Ebola cases'', instead of ``combat for  Ebola spread''.  
On the contrary, by leveraging the irrelevant documents, our model can better distill the essential information, and then generate a much more accurate intent description which is more consistent with the ground-truth.

Furthermore, we analyze the effect of irrelevant documents in our \textit{CtrsGen} model.  
As shown in Figure \ref{fig:heatmap}, we visualize the sentence-level decoder attention weights $\alpha_{z,u}^r$ (Eq. (\ref{relevant-att})) over the relevant documents from our model variant \textit{CtrsGen$_\text{-I}$}, and the adjusted weights $\beta_{z,u}^r \alpha_{z,u}^r$ (Eq. (\ref{equ:section-aware})) over the relevant documents from our \textit{CtrsGen} model.    
From the test data, we select a new query ``Mexican Air Pollution'' with ground-truth intent description ``Mexico City has the worst air pollution in the world. Pertinent Documents would contain the specific steps Mexican authorities have taken to combat this deplorable situation''. 
Due to space limitation, we only visualize sampled $6$ sentences in the relevant documents for generating the first word in the description.  
As we can see, $\textit{CtrsGen$_\text{-I}$}$ pays too much attention on the 2-$th$ and 6-$th$ sentences which confuse the model to  generate a description mainly about the ``Free Trade Agreement''.    
Specifically, the attention weight of the 2-$th$ and 6-$th$ sentence computed by \textit{CtrsGen} has a significantly drop as compared with \textit{CtrsGen$_\text{-I}$} due to the high similarity with irrelevant documents. This in turn guides the decoder to pay attention to those informative sentences and generate a much better intent description.

\subsection{Potential Application}

Here, we discuss the potential usage of such Q2ID technique. 
An interesting application would be to facilitate the exploratory search, i.e., interpreting the search results in exploratory search. 

\begin{figure}[t]
	\centering
		\includegraphics[scale=0.295]{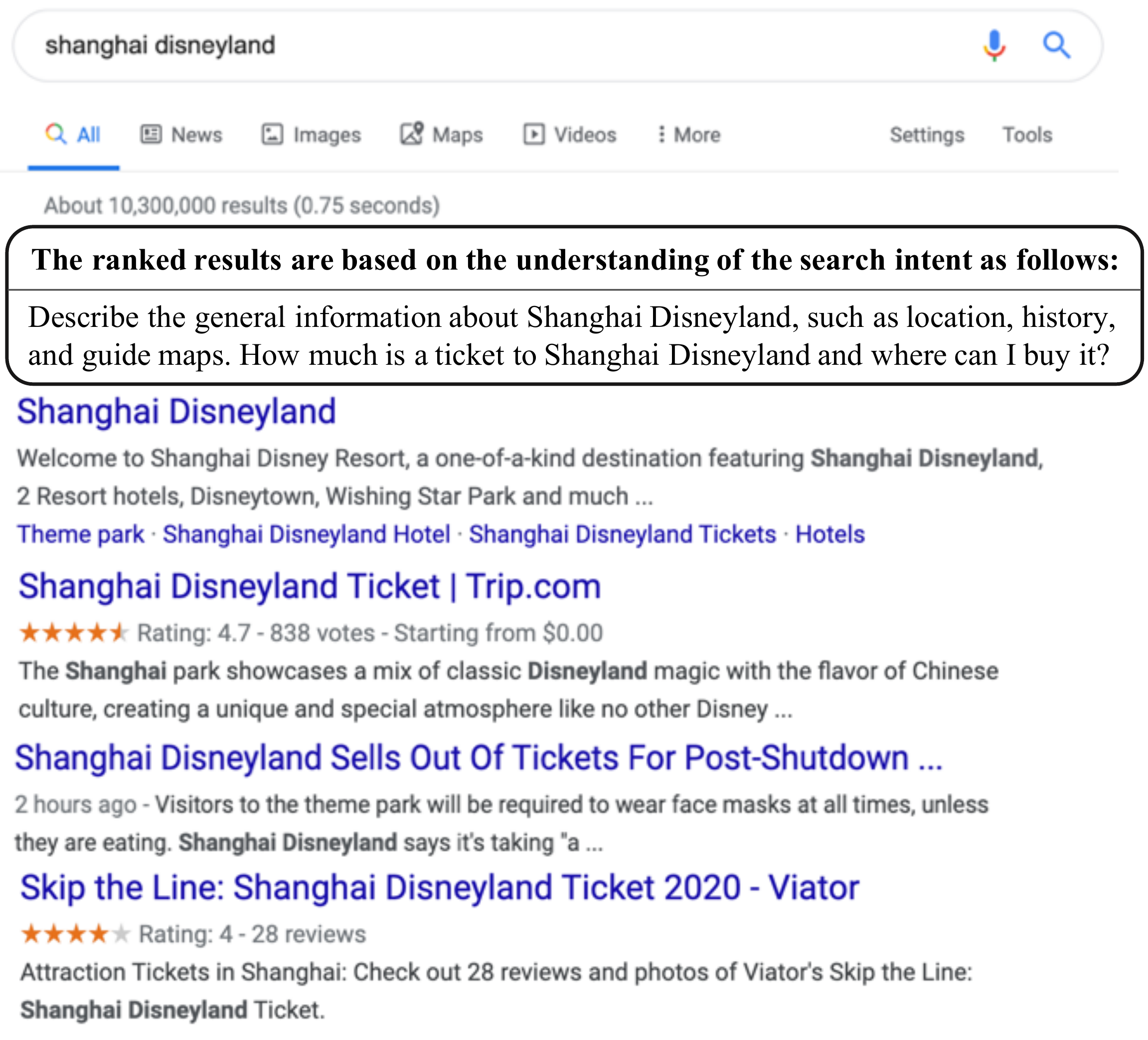}
		\caption{An example application in exploratory search using our Q2ID technique.}
\label{fig:rank}  
\end{figure}

In exploratory search, users are often unclear with their information needs initially and thus queries may be broad and vague. 
Based on the pseudo relevant feedback idea \cite{Tao2006Regularized}, we can treat the top $k$ ranked documents as relevant documents and others as irrelevant documents, and then leverage the Q2ID technique to generate the intent description. 
Such description could be viewed as an explanation of how the search engine understands the query and why those documents are displayed at the top ranks. 
With such interpretation, users may better understand the search results and find a direction to refine their query. 
For instance, as shown in Figure \ref{fig:rank}, the query is ``Shanghai Disneyland'', and an intent description of this ambiguous query ``Describe the general information about Shanghai Disneyland, such as location, history, and guide maps. How much is a ticket to Shanghai Disneyland and where can I buy it?'' is generated based on the understanding of the search intent by the search engine. 
If the user's intent is to find ``What are the must-see attractions at Shanghai Disneyland?'', he/she will easily find that the search engine has not captured that aspect and a refinement (e.g., Shanghai Disneyland Attractions) is necessary in the following. 
This can provide the user with a better overall search experience, in turn, boost the retrieval performance and the confidence of a search engine.

\section{Conclusion}
In this paper, we introduced a challenging Q2ID task for query understanding via generating a natural language intent description based on relevant and irrelevant documents of a given query.  
To tackle this problem, we developed a novel Contrastive Generation model to contrast the relevant documents with the irrelevant documents given a query.    
Empirical results over our constructed Q2ID dataset showed that our model can well understand the query through a detailed and precise intent description. 

In the future work, we would like to consider the multi-graded relevance labels of documents with respect to the query intent, and realize the potential usage of such Q2ID technique in the above discussed application. 
Also, it is valuable to do experiments on transferring intent descriptions learned on one corpus to another due to the lack of labeled data of this kind.

\section{Acknowledgments}

 This work was supported by Beijing Academy of Artificial Intelligence (BAAI) under Grants No. BAAI2019ZD0306 and BAAI2020ZJ0303, and funded by the National Natural Science Foundation of China (NSFC) under Grants No. 61722211, 61773362, 61872338, and 61902381, the Youth Innovation Promotion Association CAS under Grants No. 20144310, and 2016102, the National Key RD Program of China under Grants No. 2016QY02D0405, the Lenovo-CAS Joint Lab Youth Scientist Project, the K.C.Wong Education Foundation, and the Foundation and Frontier Research Key Program of Chongqing Science and Technology Commission (No. cstc2017jcyjBX0059).

\bibliographystyle{ACM-Reference-Format}
\bibliography{sample-bibliography}


\begin{thebibliography}{40}


\ifx \showCODEN    \undefined \def \showCODEN     #1{\unskip}     \fi
\ifx \showDOI      \undefined \def \showDOI       #1{#1}\fi
\ifx \showISBNx    \undefined \def \showISBNx     #1{\unskip}     \fi
\ifx \showISBNxiii \undefined \def \showISBNxiii  #1{\unskip}     \fi
\ifx \showISSN     \undefined \def \showISSN      #1{\unskip}     \fi
\ifx \showLCCN     \undefined \def \showLCCN      #1{\unskip}     \fi
\ifx \shownote     \undefined \def \shownote      #1{#1}          \fi
\ifx \showarticletitle \undefined \def \showarticletitle #1{#1}   \fi
\ifx \showURL      \undefined \def \showURL       {\relax}        \fi
\providecommand\bibfield[2]{#2}
\providecommand\bibinfo[2]{#2}
\providecommand\natexlab[1]{#1}
\providecommand\showeprint[2][]{arXiv:#2}

\bibitem[\protect\citeauthoryear{Bahdanau, Cho, and Bengio}{Bahdanau
  et~al\mbox{.}}{2015}]%
        {bahdanau2014neural}
\bibfield{author}{\bibinfo{person}{Dzmitry Bahdanau},
  \bibinfo{person}{Kyunghyun Cho}, {and} \bibinfo{person}{Yoshua Bengio}.}
  \bibinfo{year}{2015}\natexlab{}.
\newblock \showarticletitle{Neural machine translation by jointly learning to
  align and translate}. In \bibinfo{booktitle}{{\em ICLR}}.
\newblock


\bibitem[\protect\citeauthoryear{Banko, Mittal, and Witbrock}{Banko
  et~al\mbox{.}}{2000}]%
        {banko2000headline}
\bibfield{author}{\bibinfo{person}{Michele Banko}, \bibinfo{person}{Vibhu~O
  Mittal}, {and} \bibinfo{person}{Michael~J Witbrock}.}
  \bibinfo{year}{2000}\natexlab{}.
\newblock \showarticletitle{Headline generation based on statistical
  translation}. In \bibinfo{booktitle}{{\em Proceedings of the 38th Annual
  Meeting on Association for Computational Linguistics}}. Association for
  Computational Linguistics, \bibinfo{pages}{318--325}.
\newblock


\bibitem[\protect\citeauthoryear{Blei, Ng, and Jordan}{Blei
  et~al\mbox{.}}{2003}]%
        {blei2003latent}
\bibfield{author}{\bibinfo{person}{David~M Blei}, \bibinfo{person}{Andrew~Y
  Ng}, {and} \bibinfo{person}{Michael~I Jordan}.}
  \bibinfo{year}{2003}\natexlab{}.
\newblock \showarticletitle{Latent dirichlet allocation}.
\newblock \bibinfo{journal}{{\em Journal of machine Learning research\/}}
  \bibinfo{volume}{3}, \bibinfo{number}{Jan} (\bibinfo{year}{2003}),
  \bibinfo{pages}{993--1022}.
\newblock


\bibitem[\protect\citeauthoryear{Bougouin, Boudin, and Daille}{Bougouin
  et~al\mbox{.}}{2013}]%
        {bougouin2013topicrank}
\bibfield{author}{\bibinfo{person}{Adrien Bougouin}, \bibinfo{person}{Florian
  Boudin}, {and} \bibinfo{person}{B{\'e}atrice Daille}.}
  \bibinfo{year}{2013}\natexlab{}.
\newblock \showarticletitle{Topicrank: Graph-based topic ranking for keyphrase
  extraction}. In \bibinfo{booktitle}{{\em International Joint Conference on
  Natural Language Processing (IJCNLP)}}. \bibinfo{pages}{543--551}.
\newblock


\bibitem[\protect\citeauthoryear{Chen, Zhang, Wu, Yan, and Li}{Chen
  et~al\mbox{.}}{2018}]%
        {chen2018keyphrase}
\bibfield{author}{\bibinfo{person}{Jun Chen}, \bibinfo{person}{Xiaoming Zhang},
  \bibinfo{person}{Yu Wu}, \bibinfo{person}{Zhao Yan}, {and}
  \bibinfo{person}{Zhoujun Li}.} \bibinfo{year}{2018}\natexlab{}.
\newblock \showarticletitle{Keyphrase generation with correlation constraints}.
\newblock \bibinfo{journal}{{\em arXiv preprint arXiv:1808.07185\/}}
  (\bibinfo{year}{2018}).
\newblock


\bibitem[\protect\citeauthoryear{Cho, Van~Merri{\"e}nboer, Gulcehre, Bahdanau,
  Bougares, Schwenk, and Bengio}{Cho et~al\mbox{.}}{2014}]%
        {cho2014learning}
\bibfield{author}{\bibinfo{person}{Kyunghyun Cho}, \bibinfo{person}{Bart
  Van~Merri{\"e}nboer}, \bibinfo{person}{Caglar Gulcehre},
  \bibinfo{person}{Dzmitry Bahdanau}, \bibinfo{person}{Fethi Bougares},
  \bibinfo{person}{Holger Schwenk}, {and} \bibinfo{person}{Yoshua Bengio}.}
  \bibinfo{year}{2014}\natexlab{}.
\newblock \showarticletitle{Learning phrase representations using RNN
  encoder-decoder for statistical machine translation}. In
  \bibinfo{booktitle}{{\em Proceedings of the conference on empirical methods
  in natural language processing}}.
\newblock


\bibitem[\protect\citeauthoryear{Chopra, Auli, Rush, and Harvard}{Chopra
  et~al\mbox{.}}{2016}]%
        {chopra2016abstractive}
\bibfield{author}{\bibinfo{person}{Sumit Chopra}, \bibinfo{person}{Michael
  Auli}, \bibinfo{person}{Alexander~M Rush}, {and} \bibinfo{person}{SEAS
  Harvard}.} \bibinfo{year}{2016}\natexlab{}.
\newblock \showarticletitle{Abstractive Sentence Summarization with Attentive
  Recurrent Neural Networks}. In \bibinfo{booktitle}{{\em HLT-NAACL}}.
  \bibinfo{pages}{93--98}.
\newblock


\bibitem[\protect\citeauthoryear{Dorr, Zajic, and Schwartz}{Dorr
  et~al\mbox{.}}{2003}]%
        {dorr2003hedge}
\bibfield{author}{\bibinfo{person}{Bonnie Dorr}, \bibinfo{person}{David Zajic},
  {and} \bibinfo{person}{Richard Schwartz}.} \bibinfo{year}{2003}\natexlab{}.
\newblock \showarticletitle{Hedge trimmer: A parse-and-trim approach to
  headline generation}. In \bibinfo{booktitle}{{\em HLT-NAACL}}. Association
  for Computational Linguistics, \bibinfo{pages}{1--8}.
\newblock


\bibitem[\protect\citeauthoryear{Edmundson}{Edmundson}{1964}]%
        {edmundson1964problems}
\bibfield{author}{\bibinfo{person}{HP Edmundson}.}
  \bibinfo{year}{1964}\natexlab{}.
\newblock \showarticletitle{Problems in automatic abstracting}.
\newblock \bibinfo{journal}{{\it Commun. ACM}} \bibinfo{volume}{7},
  \bibinfo{number}{4} (\bibinfo{year}{1964}), \bibinfo{pages}{259--263}.
\newblock


\bibitem[\protect\citeauthoryear{Frank, Paynter, Witten, Gutwin, and
  Nevill-Manning}{Frank et~al\mbox{.}}{1999}]%
        {frank1999domain}
\bibfield{author}{\bibinfo{person}{Eibe Frank}, \bibinfo{person}{Gordon~W
  Paynter}, \bibinfo{person}{Ian~H Witten}, \bibinfo{person}{Carl Gutwin},
  {and} \bibinfo{person}{Craig~G Nevill-Manning}.}
  \bibinfo{year}{1999}\natexlab{}.
\newblock \showarticletitle{Domain-specific keyphrase extraction}. In
  \bibinfo{booktitle}{{\em 16th International joint conference on artificial
  intelligence (IJCAI 99)}}, Vol.~\bibinfo{volume}{2}. Morgan Kaufmann
  Publishers Inc., San Francisco, CA, USA, \bibinfo{pages}{668--673}.
\newblock


\bibitem[\protect\citeauthoryear{Gilks, Richardson, and Spiegelhalter}{Gilks
  et~al\mbox{.}}{1995}]%
        {gilks1995markov}
\bibfield{author}{\bibinfo{person}{Walter~R Gilks}, \bibinfo{person}{Sylvia
  Richardson}, {and} \bibinfo{person}{David Spiegelhalter}.}
  \bibinfo{year}{1995}\natexlab{}.
\newblock \bibinfo{booktitle}{{\em Markov chain Monte Carlo in practice}}.
\newblock \bibinfo{publisher}{Chapman and Hall/CRC}.
\newblock


\bibitem[\protect\citeauthoryear{Gollapalli and Caragea}{Gollapalli and
  Caragea}{2014}]%
        {gollapalli2014extracting}
\bibfield{author}{\bibinfo{person}{Sujatha~Das Gollapalli} {and}
  \bibinfo{person}{Cornelia Caragea}.} \bibinfo{year}{2014}\natexlab{}.
\newblock \showarticletitle{Extracting Keyphrases from Research Papers Using
  Citation Networks.}. In \bibinfo{booktitle}{{\em AAAI}}.
  \bibinfo{pages}{1629--1635}.
\newblock


\bibitem[\protect\citeauthoryear{Gu, Lu, Li, and Li}{Gu et~al\mbox{.}}{2016}]%
        {gu2016incorporating}
\bibfield{author}{\bibinfo{person}{Jiatao Gu}, \bibinfo{person}{Zhengdong Lu},
  \bibinfo{person}{Hang Li}, {and} \bibinfo{person}{Victor~OK Li}.}
  \bibinfo{year}{2016}\natexlab{}.
\newblock \showarticletitle{Incorporating copying mechanism in
  sequence-to-sequence learning}.
\newblock \bibinfo{journal}{{\em arXiv preprint arXiv:1603.06393\/}}
  (\bibinfo{year}{2016}).
\newblock


\bibitem[\protect\citeauthoryear{Hofmann}{Hofmann}{1999}]%
        {hofmann1999probabilistic}
\bibfield{author}{\bibinfo{person}{Thomas Hofmann}.}
  \bibinfo{year}{1999}\natexlab{}.
\newblock \showarticletitle{Probabilistic latent semantic indexing}. In
  \bibinfo{booktitle}{{\em SIGIR}}. ACM, \bibinfo{pages}{50--57}.
\newblock


\bibitem[\protect\citeauthoryear{Hulth}{Hulth}{2003}]%
        {hulth2003improved}
\bibfield{author}{\bibinfo{person}{Anette Hulth}.}
  \bibinfo{year}{2003}\natexlab{}.
\newblock \showarticletitle{Improved automatic keyword extraction given more
  linguistic knowledge}. In \bibinfo{booktitle}{{\em Proceedings of the 2003
  conference on Empirical methods in natural language processing}}. Association
  for Computational Linguistics, \bibinfo{pages}{216--223}.
\newblock


\bibitem[\protect\citeauthoryear{Kingma and Ba}{Kingma and Ba}{2015}]%
        {kingma2014adam}
\bibfield{author}{\bibinfo{person}{Diederik Kingma} {and}
  \bibinfo{person}{Jimmy Ba}.} \bibinfo{year}{2015}\natexlab{}.
\newblock \showarticletitle{Adam: A method for stochastic optimization}. In
  \bibinfo{booktitle}{{\em ICLR}}.
\newblock


\bibitem[\protect\citeauthoryear{Langer, L{\"u}ngen, and Bayerl}{Langer
  et~al\mbox{.}}{2004}]%
        {langer2004text}
\bibfield{author}{\bibinfo{person}{Hagen Langer}, \bibinfo{person}{Harald
  L{\"u}ngen}, {and} \bibinfo{person}{Petra~Saskia Bayerl}.}
  \bibinfo{year}{2004}\natexlab{}.
\newblock \showarticletitle{Text type structure and logical document
  structure}. In \bibinfo{booktitle}{{\em Proceedings of the 2004 ACL Workshop
  on Discourse Annotation}}. Association for Computational Linguistics,
  \bibinfo{pages}{49--56}.
\newblock


\bibitem[\protect\citeauthoryear{Li, Luong, and Jurafsky}{Li
  et~al\mbox{.}}{2015}]%
        {li2015hierarchical}
\bibfield{author}{\bibinfo{person}{Jiwei Li}, \bibinfo{person}{Minh-Thang
  Luong}, {and} \bibinfo{person}{Dan Jurafsky}.}
  \bibinfo{year}{2015}\natexlab{}.
\newblock \showarticletitle{A hierarchical neural autoencoder for paragraphs
  and documents}.
\newblock \bibinfo{journal}{{\em arXiv preprint arXiv:1506.01057\/}}
  (\bibinfo{year}{2015}).
\newblock


\bibitem[\protect\citeauthoryear{Lin}{Lin}{2004}]%
        {lin2004rouge}
\bibfield{author}{\bibinfo{person}{Chin-Yew Lin}.}
  \bibinfo{year}{2004}\natexlab{}.
\newblock \showarticletitle{Rouge: A package for automatic evaluation of
  summaries}. In \bibinfo{booktitle}{{\em Text summarization branches out:
  Proceedings of the ACL-04 workshop}}, Vol.~\bibinfo{volume}{8}. Barcelona,
  Spain.
\newblock


\bibitem[\protect\citeauthoryear{Liu, Pennell, Liu, and Liu}{Liu
  et~al\mbox{.}}{2009}]%
        {liu2009unsupervised}
\bibfield{author}{\bibinfo{person}{Feifan Liu}, \bibinfo{person}{Deana
  Pennell}, \bibinfo{person}{Fei Liu}, {and} \bibinfo{person}{Yang Liu}.}
  \bibinfo{year}{2009}\natexlab{}.
\newblock \showarticletitle{Unsupervised approaches for automatic keyword
  extraction using meeting transcripts}. In \bibinfo{booktitle}{{\em
  Proceedings of human language technologies: The 2009 annual conference of the
  North American chapter of the association for computational linguistics}}.
  Association for Computational Linguistics, \bibinfo{pages}{620--628}.
\newblock


\bibitem[\protect\citeauthoryear{Lopyrev}{Lopyrev}{2015}]%
        {lopyrev2015generating}
\bibfield{author}{\bibinfo{person}{Konstantin Lopyrev}.}
  \bibinfo{year}{2015}\natexlab{}.
\newblock \showarticletitle{Generating news headlines with recurrent neural
  networks}.
\newblock \bibinfo{journal}{{\em arXiv preprint arXiv:1512.01712\/}}
  (\bibinfo{year}{2015}).
\newblock


\bibitem[\protect\citeauthoryear{Luhn}{Luhn}{1958}]%
        {luhn1958automatic}
\bibfield{author}{\bibinfo{person}{Hans~Peter Luhn}.}
  \bibinfo{year}{1958}\natexlab{}.
\newblock \showarticletitle{The automatic creation of literature abstracts}.
\newblock \bibinfo{journal}{{\em IBM Journal of research and development\/}}
  \bibinfo{volume}{2}, \bibinfo{number}{2} (\bibinfo{year}{1958}),
  \bibinfo{pages}{159--165}.
\newblock


\bibitem[\protect\citeauthoryear{Medelyan, Frank, and Witten}{Medelyan
  et~al\mbox{.}}{2009}]%
        {medelyan2009human}
\bibfield{author}{\bibinfo{person}{Olena Medelyan}, \bibinfo{person}{Eibe
  Frank}, {and} \bibinfo{person}{Ian~H Witten}.}
  \bibinfo{year}{2009}\natexlab{}.
\newblock \showarticletitle{Human-competitive tagging using automatic keyphrase
  extraction}. In \bibinfo{booktitle}{{\em Proceedings of the 2009 Conference
  on Empirical Methods in Natural Language Processing: Volume 3-Volume 3}}.
  Association for Computational Linguistics, \bibinfo{pages}{1318--1327}.
\newblock


\bibitem[\protect\citeauthoryear{Meng, Zhao, Han, He, Brusilovsky, and
  Chi}{Meng et~al\mbox{.}}{2017}]%
        {meng2017deep}
\bibfield{author}{\bibinfo{person}{Rui Meng}, \bibinfo{person}{Sanqiang Zhao},
  \bibinfo{person}{Shuguang Han}, \bibinfo{person}{Daqing He},
  \bibinfo{person}{Peter Brusilovsky}, {and} \bibinfo{person}{Yu Chi}.}
  \bibinfo{year}{2017}\natexlab{}.
\newblock \showarticletitle{Deep keyphrase generation}.
\newblock \bibinfo{journal}{{\em arXiv preprint arXiv:1704.06879\/}}
  (\bibinfo{year}{2017}).
\newblock


\bibitem[\protect\citeauthoryear{Mihalcea and Tarau}{Mihalcea and
  Tarau}{2004}]%
        {mihalcea2004textrank}
\bibfield{author}{\bibinfo{person}{Rada Mihalcea} {and} \bibinfo{person}{Paul
  Tarau}.} \bibinfo{year}{2004}\natexlab{}.
\newblock \showarticletitle{Textrank: Bringing order into text}. In
  \bibinfo{booktitle}{{\em Proceedings of the 2004 conference on empirical
  methods in natural language processing}}.
\newblock


\bibitem[\protect\citeauthoryear{Nallapati, Zhou, Gulcehre, Xiang,
  et~al\mbox{.}}{Nallapati et~al\mbox{.}}{2016}]%
        {nallapati2016abstractive}
\bibfield{author}{\bibinfo{person}{Ramesh Nallapati}, \bibinfo{person}{Bowen
  Zhou}, \bibinfo{person}{Caglar Gulcehre}, \bibinfo{person}{Bing Xiang},
  {et~al\mbox{.}}} \bibinfo{year}{2016}\natexlab{}.
\newblock \showarticletitle{Abstractive text summarization using
  sequence-to-sequence rnns and beyond}.
\newblock \bibinfo{journal}{{\em arXiv preprint arXiv:1602.06023\/}}
  (\bibinfo{year}{2016}).
\newblock


\bibitem[\protect\citeauthoryear{Nenkova and Vanderwende}{Nenkova and
  Vanderwende}{[n. d.]}]%
        {nenkova2005impact}
\bibfield{author}{\bibinfo{person}{Ani Nenkova} {and} \bibinfo{person}{Lucy
  Vanderwende}.} \bibinfo{year}{[n. d.]}\natexlab{}.
\newblock \showarticletitle{The impact of frequency on summarization}.
\newblock  (\bibinfo{year}{[n. d.]}).
\newblock


\bibitem[\protect\citeauthoryear{Radev and McKeown}{Radev and McKeown}{1998}]%
        {radev1998generating}
\bibfield{author}{\bibinfo{person}{Dragomir~R Radev} {and}
  \bibinfo{person}{Kathleen~R McKeown}.} \bibinfo{year}{1998}\natexlab{}.
\newblock \showarticletitle{Generating natural language summaries from multiple
  on-line sources}.
\newblock \bibinfo{journal}{{\em Computational Linguistics\/}}
  \bibinfo{volume}{24}, \bibinfo{number}{3} (\bibinfo{year}{1998}),
  \bibinfo{pages}{470--500}.
\newblock


\bibitem[\protect\citeauthoryear{Robertson and Walker}{Robertson and
  Walker}{1994}]%
        {robertson1994some}
\bibfield{author}{\bibinfo{person}{Stephen~E Robertson} {and}
  \bibinfo{person}{Steve Walker}.} \bibinfo{year}{1994}\natexlab{}.
\newblock \showarticletitle{Some simple effective approximations to the
  2-poisson model for probabilistic weighted retrieval}. In
  \bibinfo{booktitle}{{\em SIGIR}}. Springer-Verlag New York, Inc.,
  \bibinfo{pages}{232--241}.
\newblock


\bibitem[\protect\citeauthoryear{Rush, Chopra, and Weston}{Rush
  et~al\mbox{.}}{2015}]%
        {rush2015neural}
\bibfield{author}{\bibinfo{person}{Alexander~M Rush}, \bibinfo{person}{Sumit
  Chopra}, {and} \bibinfo{person}{Jason Weston}.}
  \bibinfo{year}{2015}\natexlab{}.
\newblock \showarticletitle{A neural attention model for abstractive sentence
  summarization}. In \bibinfo{booktitle}{{\em EMNLP}}.
\newblock


\bibitem[\protect\citeauthoryear{See, Liu, and Manning}{See
  et~al\mbox{.}}{2017}]%
        {see2017get}
\bibfield{author}{\bibinfo{person}{Abigail See}, \bibinfo{person}{Peter~J Liu},
  {and} \bibinfo{person}{Christopher~D Manning}.}
  \bibinfo{year}{2017}\natexlab{}.
\newblock \showarticletitle{Get to the point: Summarization with
  pointer-generator networks}.
\newblock \bibinfo{journal}{{\em arXiv preprint arXiv:1704.04368\/}}
  (\bibinfo{year}{2017}).
\newblock


\bibitem[\protect\citeauthoryear{Shang, Liu, Jiang, Ren, Voss, and Han}{Shang
  et~al\mbox{.}}{2018}]%
        {shang2018automated}
\bibfield{author}{\bibinfo{person}{Jingbo Shang}, \bibinfo{person}{Jialu Liu},
  \bibinfo{person}{Meng Jiang}, \bibinfo{person}{Xiang Ren},
  \bibinfo{person}{Clare~R Voss}, {and} \bibinfo{person}{Jiawei Han}.}
  \bibinfo{year}{2018}\natexlab{}.
\newblock \showarticletitle{Automated phrase mining from massive text corpora}.
\newblock \bibinfo{journal}{{\em IEEE Transactions on Knowledge and Data
  Engineering\/}} \bibinfo{volume}{30}, \bibinfo{number}{10}
  (\bibinfo{year}{2018}), \bibinfo{pages}{1825--1837}.
\newblock


\bibitem[\protect\citeauthoryear{Stede and Suriyawongkul}{Stede and
  Suriyawongkul}{2010}]%
        {stede2010identifying}
\bibfield{author}{\bibinfo{person}{Manfred Stede} {and} \bibinfo{person}{Arthit
  Suriyawongkul}.} \bibinfo{year}{2010}\natexlab{}.
\newblock \showarticletitle{Identifying logical structure and content structure
  in loosely-structured documents}.
\newblock In \bibinfo{booktitle}{{\em Linguistic Modeling of Information and
  Markup Languages}}. \bibinfo{publisher}{Springer}, \bibinfo{pages}{81--96}.
\newblock


\bibitem[\protect\citeauthoryear{Sutskever, Vinyals, and Le}{Sutskever
  et~al\mbox{.}}{2014}]%
        {sutskever2014sequence}
\bibfield{author}{\bibinfo{person}{Ilya Sutskever}, \bibinfo{person}{Oriol
  Vinyals}, {and} \bibinfo{person}{Quoc~V Le}.}
  \bibinfo{year}{2014}\natexlab{}.
\newblock \showarticletitle{Sequence to sequence learning with neural
  networks}. In \bibinfo{booktitle}{{\em Advances in neural information
  processing systems}}. \bibinfo{pages}{3104--3112}.
\newblock


\bibitem[\protect\citeauthoryear{Tan, Wan, and Xiao}{Tan et~al\mbox{.}}{2017}]%
        {tanneural}
\bibfield{author}{\bibinfo{person}{Jiwei Tan}, \bibinfo{person}{Xiaojun Wan},
  {and} \bibinfo{person}{Jianguo Xiao}.} \bibinfo{year}{2017}\natexlab{}.
\newblock \showarticletitle{From Neural Sentence Summarization to Headline
  Generation: A Coarse-to-Fine Approach}. In \bibinfo{booktitle}{{\em IJCAI}}.
\newblock


\bibitem[\protect\citeauthoryear{Tixier, Malliaros, and Vazirgiannis}{Tixier
  et~al\mbox{.}}{2016}]%
        {tixier2016graph}
\bibfield{author}{\bibinfo{person}{Antoine Tixier}, \bibinfo{person}{Fragkiskos
  Malliaros}, {and} \bibinfo{person}{Michalis Vazirgiannis}.}
  \bibinfo{year}{2016}\natexlab{}.
\newblock \showarticletitle{A graph degeneracy-based approach to keyword
  extraction}. In \bibinfo{booktitle}{{\em Proceedings of the 2016 Conference
  on Empirical Methods in Natural Language Processing}}.
  \bibinfo{pages}{1860--1870}.
\newblock


\bibitem[\protect\citeauthoryear{Turney}{Turney}{2002}]%
        {turney2002learning}
\bibfield{author}{\bibinfo{person}{Peter~D Turney}.}
  \bibinfo{year}{2002}\natexlab{}.
\newblock \showarticletitle{Learning to extract keyphrases from text}.
\newblock \bibinfo{journal}{{\em arXiv preprint cs/0212013\/}}
  (\bibinfo{year}{2002}).
\newblock


\bibitem[\protect\citeauthoryear{Wang, Liu, and Wang}{Wang
  et~al\mbox{.}}{2007}]%
        {wang2007keyword}
\bibfield{author}{\bibinfo{person}{Jinghua Wang}, \bibinfo{person}{Jianyi Liu},
  {and} \bibinfo{person}{Cong Wang}.} \bibinfo{year}{2007}\natexlab{}.
\newblock \showarticletitle{Keyword extraction based on pagerank}. In
  \bibinfo{booktitle}{{\em Pacific-Asia Conference on Knowledge Discovery and
  Data Mining}}. Springer, \bibinfo{pages}{857--864}.
\newblock


\bibitem[\protect\citeauthoryear{Woodsend, Feng, and Lapata}{Woodsend
  et~al\mbox{.}}{2010}]%
        {woodsend2010generation}
\bibfield{author}{\bibinfo{person}{Kristian Woodsend}, \bibinfo{person}{Yansong
  Feng}, {and} \bibinfo{person}{Mirella Lapata}.}
  \bibinfo{year}{2010}\natexlab{}.
\newblock \showarticletitle{Generation with quasi-synchronous grammar}. In
  \bibinfo{booktitle}{{\em EMNLP}}. Association for Computational Linguistics,
  \bibinfo{pages}{513--523}.
\newblock


\bibitem[\protect\citeauthoryear{Xu, Yang, and Lau}{Xu et~al\mbox{.}}{2010}]%
        {xu2010keyword}
\bibfield{author}{\bibinfo{person}{Songhua Xu}, \bibinfo{person}{Shaohui Yang},
  {and} \bibinfo{person}{Francis Chi-Moon Lau}.}
  \bibinfo{year}{2010}\natexlab{}.
\newblock \showarticletitle{Keyword Extraction and Headline Generation Using
  Novel Word Features.}. In \bibinfo{booktitle}{{\em AAAI}}.
  \bibinfo{pages}{1461--1466}.
\newblock


\end{thebibliography}



\begin{thebibliography}{63}


\ifx \showCODEN    \undefined \def \showCODEN     #1{\unskip}     \fi
\ifx \showDOI      \undefined \def \showDOI       #1{#1}\fi
\ifx \showISBNx    \undefined \def \showISBNx     #1{\unskip}     \fi
\ifx \showISBNxiii \undefined \def \showISBNxiii  #1{\unskip}     \fi
\ifx \showISSN     \undefined \def \showISSN      #1{\unskip}     \fi
\ifx \showLCCN     \undefined \def \showLCCN      #1{\unskip}     \fi
\ifx \shownote     \undefined \def \shownote      #1{#1}          \fi
\ifx \showarticletitle \undefined \def \showarticletitle #1{#1}   \fi
\ifx \showURL      \undefined \def \showURL       {\relax}        \fi
\providecommand\bibfield[2]{#2}
\providecommand\bibinfo[2]{#2}
\providecommand\natexlab[1]{#1}
\providecommand\showeprint[2][]{arXiv:#2}

\bibitem[\protect\citeauthoryear{AlessandroMoschitti, AbedAlhakimFreihat,
  Glass, and Randeree}{AlessandroMoschitti et~al\mbox{.}}{2016}]%
        {alessandromoschitti2016semeval}
\bibfield{author}{\bibinfo{person}{Llu{\i}sMarquez AlessandroMoschitti,
  PreslavNakov}, \bibinfo{person}{WalidMagdy~HamdyMubarak AbedAlhakimFreihat},
  \bibinfo{person}{James Glass}, {and} \bibinfo{person}{Bilal Randeree}.}
  \bibinfo{year}{2016}\natexlab{}.
\newblock \showarticletitle{Semeval-2016 task 3: Community question answering}.
  In \bibinfo{booktitle}{{\em SemEval}}.
\newblock


\bibitem[\protect\citeauthoryear{Bahdanau, Cho, and Bengio}{Bahdanau
  et~al\mbox{.}}{2015}]%
        {bahdanau2014neural}
\bibfield{author}{\bibinfo{person}{Dzmitry Bahdanau},
  \bibinfo{person}{Kyunghyun Cho}, {and} \bibinfo{person}{Yoshua Bengio}.}
  \bibinfo{year}{2015}\natexlab{}.
\newblock \showarticletitle{Neural machine translation by jointly learning to
  align and translate}. In \bibinfo{booktitle}{{\em ICLR}}.
\newblock


\bibitem[\protect\citeauthoryear{Baumel, Eyal, and Elhadad}{Baumel
  et~al\mbox{.}}{2018}]%
        {baumel2018query}
\bibfield{author}{\bibinfo{person}{Tal Baumel}, \bibinfo{person}{Matan Eyal},
  {and} \bibinfo{person}{Michael Elhadad}.} \bibinfo{year}{2018}\natexlab{}.
\newblock \showarticletitle{Query focused abstractive summarization:
  Incorporating query relevance, multi-document coverage, and summary length
  constraints into seq2seq models}.
\newblock \bibinfo{journal}{{\em arXiv preprint arXiv:1801.07704\/}}
  (\bibinfo{year}{2018}).
\newblock


\bibitem[\protect\citeauthoryear{Beeferman and Berger}{Beeferman and
  Berger}{2000}]%
        {beeferman2000agglomerative}
\bibfield{author}{\bibinfo{person}{Doug Beeferman} {and} \bibinfo{person}{Adam
  Berger}.} \bibinfo{year}{2000}\natexlab{}.
\newblock \showarticletitle{Agglomerative clustering of a search engine query
  log}.
\newblock


\bibitem[\protect\citeauthoryear{Beitzel, Jensen, Frieder, Grossman, Lewis,
  Chowdhury, and Kolcz}{Beitzel et~al\mbox{.}}{2005a}]%
        {beitzel2005automatic}
\bibfield{author}{\bibinfo{person}{Steven~M Beitzel}, \bibinfo{person}{Eric~C
  Jensen}, \bibinfo{person}{Ophir Frieder}, \bibinfo{person}{David Grossman},
  \bibinfo{person}{David~D Lewis}, \bibinfo{person}{Abdur Chowdhury}, {and}
  \bibinfo{person}{Aleksandr Kolcz}.} \bibinfo{year}{2005}\natexlab{a}.
\newblock \showarticletitle{Automatic web query classification using labeled
  and unlabeled training data}. In \bibinfo{booktitle}{{\em SIGIR}}. ACM,
  \bibinfo{pages}{581--582}.
\newblock


\bibitem[\protect\citeauthoryear{Beitzel, Jensen, Frieder, Lewis, Chowdhury,
  and Kolcz}{Beitzel et~al\mbox{.}}{2005b}]%
        {beitzel2005improving}
\bibfield{author}{\bibinfo{person}{Steven~M Beitzel}, \bibinfo{person}{Eric~C
  Jensen}, \bibinfo{person}{Ophir Frieder}, \bibinfo{person}{David~D Lewis},
  \bibinfo{person}{Abdur Chowdhury}, {and} \bibinfo{person}{Aleksander Kolcz}.}
  \bibinfo{year}{2005}\natexlab{b}.
\newblock \showarticletitle{Improving automatic query classification via
  semi-supervised learning}. In \bibinfo{booktitle}{{\em ICDM}}.
\newblock


\bibitem[\protect\citeauthoryear{Bosma}{Bosma}{2005}]%
        {bosma2005query}
\bibfield{author}{\bibinfo{person}{Wauter Bosma}.}
  \bibinfo{year}{2005}\natexlab{}.
\newblock \showarticletitle{Query-based summarization using rhetorical
  structure theory}.
\newblock \bibinfo{journal}{{\em LOT Occasional Series\/}}
  (\bibinfo{year}{2005}).
\newblock


\bibitem[\protect\citeauthoryear{Broder}{Broder}{2002}]%
        {broder2002taxonomy}
\bibfield{author}{\bibinfo{person}{Andrei Broder}.}
  \bibinfo{year}{2002}\natexlab{}.
\newblock \showarticletitle{A taxonomy of web search}. In
  \bibinfo{booktitle}{{\em ACM Sigir forum}}.
\newblock


\bibitem[\protect\citeauthoryear{Broder, Fontoura, Gabrilovich, Joshi,
  Josifovski, and Zhang}{Broder et~al\mbox{.}}{2007}]%
        {broder2007robust}
\bibfield{author}{\bibinfo{person}{Andrei~Z Broder}, \bibinfo{person}{Marcus
  Fontoura}, \bibinfo{person}{Evgeniy Gabrilovich}, \bibinfo{person}{Amruta
  Joshi}, \bibinfo{person}{Vanja Josifovski}, {and} \bibinfo{person}{Tong
  Zhang}.} \bibinfo{year}{2007}\natexlab{}.
\newblock \showarticletitle{Robust classification of rare queries using web
  knowledge}. In \bibinfo{booktitle}{{\em SIGIR}}.
\newblock


\bibitem[\protect\citeauthoryear{Cao, Nie, Gao, and Robertson}{Cao
  et~al\mbox{.}}{2008}]%
        {cao2008selecting}
\bibfield{author}{\bibinfo{person}{Guihong Cao}, \bibinfo{person}{Jian-Yun
  Nie}, \bibinfo{person}{Jianfeng Gao}, {and} \bibinfo{person}{Stephen
  Robertson}.} \bibinfo{year}{2008}\natexlab{}.
\newblock \showarticletitle{Selecting good expansion terms for pseudo-relevance
  feedback}. In \bibinfo{booktitle}{{\em SIGIR}}.
\newblock


\bibitem[\protect\citeauthoryear{Cao, Hu, Shen, Jiang, Sun, Chen, and Yang}{Cao
  et~al\mbox{.}}{2009}]%
        {cao2009context}
\bibfield{author}{\bibinfo{person}{Huanhuan Cao}, \bibinfo{person}{Derek~Hao
  Hu}, \bibinfo{person}{Dou Shen}, \bibinfo{person}{Daxin Jiang},
  \bibinfo{person}{Jian-Tao Sun}, \bibinfo{person}{Enhong Chen}, {and}
  \bibinfo{person}{Qiang Yang}.} \bibinfo{year}{2009}\natexlab{}.
\newblock \showarticletitle{Context-aware query classification}. In
  \bibinfo{booktitle}{{\em SIGIR}}.
\newblock


\bibitem[\protect\citeauthoryear{Carbonell and Goldstein}{Carbonell and
  Goldstein}{1998}]%
        {carbonell1998use}
\bibfield{author}{\bibinfo{person}{Jaime~G Carbonell} {and}
  \bibinfo{person}{Jade Goldstein}.} \bibinfo{year}{1998}\natexlab{}.
\newblock \showarticletitle{The use of MMR, diversity-based reranking for
  reordering documents and producing summaries.}. In \bibinfo{booktitle}{{\em
  SIGIR}}.
\newblock


\bibitem[\protect\citeauthoryear{Chen and Bansal}{Chen and Bansal}{2018}]%
        {chen2018fast}
\bibfield{author}{\bibinfo{person}{Yen-Chun Chen} {and} \bibinfo{person}{Mohit
  Bansal}.} \bibinfo{year}{2018}\natexlab{}.
\newblock \showarticletitle{Fast abstractive summarization with
  reinforce-selected sentence rewriting}. In \bibinfo{booktitle}{{\em ACL}}.
\newblock


\bibitem[\protect\citeauthoryear{Cho, Van~Merri{\"e}nboer, Gulcehre, Bahdanau,
  Bougares, Schwenk, and Bengio}{Cho et~al\mbox{.}}{2014}]%
        {cho2014learning}
\bibfield{author}{\bibinfo{person}{Kyunghyun Cho}, \bibinfo{person}{Bart
  Van~Merri{\"e}nboer}, \bibinfo{person}{Caglar Gulcehre},
  \bibinfo{person}{Dzmitry Bahdanau}, \bibinfo{person}{Fethi Bougares},
  \bibinfo{person}{Holger Schwenk}, {and} \bibinfo{person}{Yoshua Bengio}.}
  \bibinfo{year}{2014}\natexlab{}.
\newblock \showarticletitle{Learning phrase representations using RNN
  encoder-decoder for statistical machine translation}. In
  \bibinfo{booktitle}{{\em EMNLP}}.
\newblock


\bibitem[\protect\citeauthoryear{Cho, Lebanoff, Foroosh, and Liu}{Cho
  et~al\mbox{.}}{2019}]%
        {cho2019improving}
\bibfield{author}{\bibinfo{person}{Sangwoo Cho}, \bibinfo{person}{Logan
  Lebanoff}, \bibinfo{person}{Hassan Foroosh}, {and} \bibinfo{person}{Fei
  Liu}.} \bibinfo{year}{2019}\natexlab{}.
\newblock \showarticletitle{Improving the Similarity Measure of Determinantal
  Point Processes for Extractive Multi-Document Summarization}. In
  \bibinfo{booktitle}{{\em ACL}}.
\newblock


\bibitem[\protect\citeauthoryear{Chopra, Auli, and Rush}{Chopra
  et~al\mbox{.}}{2016}]%
        {chopra2016abstractive}
\bibfield{author}{\bibinfo{person}{Sumit Chopra}, \bibinfo{person}{Michael
  Auli}, {and} \bibinfo{person}{Alexander~M Rush}.}
  \bibinfo{year}{2016}\natexlab{}.
\newblock \showarticletitle{Abstractive sentence summarization with attentive
  recurrent neural networks}. In \bibinfo{booktitle}{{\em NAACL}}.
\newblock


\bibitem[\protect\citeauthoryear{Cronen-Townsend, Zhou, and
  Croft}{Cronen-Townsend et~al\mbox{.}}{2002}]%
        {cronen2002predicting}
\bibfield{author}{\bibinfo{person}{Steve Cronen-Townsend}, \bibinfo{person}{Yun
  Zhou}, {and} \bibinfo{person}{W~Bruce Croft}.}
  \bibinfo{year}{2002}\natexlab{}.
\newblock \showarticletitle{Predicting query performance}. In
  \bibinfo{booktitle}{{\em SIGIR}}.
\newblock


\bibitem[\protect\citeauthoryear{Cui, Wen, Nie, and Ma}{Cui
  et~al\mbox{.}}{2002}]%
        {cui2002probabilistic}
\bibfield{author}{\bibinfo{person}{Hang Cui}, \bibinfo{person}{Ji-Rong Wen},
  \bibinfo{person}{Jian-Yun Nie}, {and} \bibinfo{person}{Wei-Ying Ma}.}
  \bibinfo{year}{2002}\natexlab{}.
\newblock \showarticletitle{Probabilistic query expansion using query logs}. In
  \bibinfo{booktitle}{{\em WWW}}.
\newblock


\bibitem[\protect\citeauthoryear{Dang and Croft}{Dang and Croft}{2010}]%
        {dang2010query}
\bibfield{author}{\bibinfo{person}{Van Dang} {and} \bibinfo{person}{Bruce~W
  Croft}.} \bibinfo{year}{2010}\natexlab{}.
\newblock \showarticletitle{Query reformulation using anchor text}. In
  \bibinfo{booktitle}{{\em WSDM}}.
\newblock


\bibitem[\protect\citeauthoryear{Daum{\'e}~III and Marcu}{Daum{\'e}~III and
  Marcu}{2006}]%
        {daume2006bayesian}
\bibfield{author}{\bibinfo{person}{Hal Daum{\'e}~III} {and}
  \bibinfo{person}{Daniel Marcu}.} \bibinfo{year}{2006}\natexlab{}.
\newblock \showarticletitle{Bayesian query-focused summarization}. In
  \bibinfo{booktitle}{{\em ACL}}.
\newblock


\bibitem[\protect\citeauthoryear{Erkan and Radev}{Erkan and Radev}{2004}]%
        {erkan2004lexrank}
\bibfield{author}{\bibinfo{person}{G{\"u}nes Erkan} {and}
  \bibinfo{person}{Dragomir~R Radev}.} \bibinfo{year}{2004}\natexlab{}.
\newblock \showarticletitle{Lexrank: Graph-based lexical centrality as salience
  in text summarization}.
\newblock \bibinfo{journal}{{\em Journal of artificial intelligence
  research\/}} (\bibinfo{year}{2004}).
\newblock


\bibitem[\protect\citeauthoryear{Fang}{Fang}{2008}]%
        {fang2008re}
\bibfield{author}{\bibinfo{person}{Hui Fang}.} \bibinfo{year}{2008}\natexlab{}.
\newblock \showarticletitle{A re-examination of query expansion using lexical
  resources}. In \bibinfo{booktitle}{{\em ACL}}.
\newblock


\bibitem[\protect\citeauthoryear{Goldstein, Kantrowitz, Mittal, and
  Carbonell}{Goldstein et~al\mbox{.}}{1999}]%
        {goldstein1999summarizing}
\bibfield{author}{\bibinfo{person}{Jade Goldstein}, \bibinfo{person}{Mark
  Kantrowitz}, \bibinfo{person}{Vibhu Mittal}, {and} \bibinfo{person}{Jaime
  Carbonell}.} \bibinfo{year}{1999}\natexlab{}.
\newblock \showarticletitle{Summarizing text documents: sentence selection and
  evaluation metrics}. In \bibinfo{booktitle}{{\em SIGIR}}.
\newblock


\bibitem[\protect\citeauthoryear{Hanauer, Wu, Yang, Mei, Murkowski-Steffy,
  Vydiswaran, and Zheng}{Hanauer et~al\mbox{.}}{2017}]%
        {hanauer2017development}
\bibfield{author}{\bibinfo{person}{David~A Hanauer}, \bibinfo{person}{Danny~TY
  Wu}, \bibinfo{person}{Lei Yang}, \bibinfo{person}{Qiaozhu Mei},
  \bibinfo{person}{Katherine~B Murkowski-Steffy}, \bibinfo{person}{VG~Vinod
  Vydiswaran}, {and} \bibinfo{person}{Kai Zheng}.}
  \bibinfo{year}{2017}\natexlab{}.
\newblock \showarticletitle{Development and empirical user-centered evaluation
  of semantically-based query recommendation for an electronic health record
  search engine}.
\newblock \bibinfo{journal}{{\em Journal of biomedical informatics\/}}
  (\bibinfo{year}{2017}).
\newblock


\bibitem[\protect\citeauthoryear{Hasselqvist, Helmertz, and
  K{\aa}geb{\"a}ck}{Hasselqvist et~al\mbox{.}}{2017}]%
        {hasselqvist2017query}
\bibfield{author}{\bibinfo{person}{Johan Hasselqvist}, \bibinfo{person}{Niklas
  Helmertz}, {and} \bibinfo{person}{Mikael K{\aa}geb{\"a}ck}.}
  \bibinfo{year}{2017}\natexlab{}.
\newblock \showarticletitle{Query-based abstractive summarization using neural
  networks}.
\newblock \bibinfo{journal}{{\em arXiv preprint arXiv:1712.06100\/}}
  (\bibinfo{year}{2017}).
\newblock


\bibitem[\protect\citeauthoryear{He and Ounis}{He and Ounis}{2006}]%
        {he2006query}
\bibfield{author}{\bibinfo{person}{Ben He} {and} \bibinfo{person}{Iadh Ounis}.}
  \bibinfo{year}{2006}\natexlab{}.
\newblock \showarticletitle{Query performance prediction}.
\newblock \bibinfo{journal}{{\em Information Systems\/}} \bibinfo{volume}{31},
  \bibinfo{number}{7} (\bibinfo{year}{2006}), \bibinfo{pages}{585--594}.
\newblock


\bibitem[\protect\citeauthoryear{Hermann, Kocisky, Grefenstette, Espeholt, Kay,
  Suleyman, and Blunsom}{Hermann et~al\mbox{.}}{2015}]%
        {hermann2015teaching}
\bibfield{author}{\bibinfo{person}{Karl~Moritz Hermann}, \bibinfo{person}{Tomas
  Kocisky}, \bibinfo{person}{Edward Grefenstette}, \bibinfo{person}{Lasse
  Espeholt}, \bibinfo{person}{Will Kay}, \bibinfo{person}{Mustafa Suleyman},
  {and} \bibinfo{person}{Phil Blunsom}.} \bibinfo{year}{2015}\natexlab{}.
\newblock \showarticletitle{Teaching machines to read and comprehend}. In
  \bibinfo{booktitle}{{\em NIPS}}.
\newblock


\bibitem[\protect\citeauthoryear{Hong, Vaidya, Lu, and Liu}{Hong
  et~al\mbox{.}}{2016}]%
        {hong2016accurate}
\bibfield{author}{\bibinfo{person}{Yuan Hong}, \bibinfo{person}{Jaideep
  Vaidya}, \bibinfo{person}{Haibing Lu}, {and} \bibinfo{person}{Wen~Ming Liu}.}
  \bibinfo{year}{2016}\natexlab{}.
\newblock \showarticletitle{Accurate and efficient query clustering via top
  ranked search results}. In \bibinfo{booktitle}{{\em Web Intelligence}}.
\newblock


\bibitem[\protect\citeauthoryear{Hsu, Lin, Lee, Min, Tang, and Sun}{Hsu
  et~al\mbox{.}}{2018}]%
        {hsu2018unified}
\bibfield{author}{\bibinfo{person}{Wan-Ting Hsu}, \bibinfo{person}{Chieh-Kai
  Lin}, \bibinfo{person}{Ming-Ying Lee}, \bibinfo{person}{Kerui Min},
  \bibinfo{person}{Jing Tang}, {and} \bibinfo{person}{Min Sun}.}
  \bibinfo{year}{2018}\natexlab{}.
\newblock \showarticletitle{A unified model for extractive and abstractive
  summarization using inconsistency loss}. In \bibinfo{booktitle}{{\em ACL}}.
\newblock


\bibitem[\protect\citeauthoryear{Hu, Wang, Lochovsky, Sun, and Chen}{Hu
  et~al\mbox{.}}{2009}]%
        {hu2009understanding}
\bibfield{author}{\bibinfo{person}{Jian Hu}, \bibinfo{person}{Gang Wang},
  \bibinfo{person}{Fred Lochovsky}, \bibinfo{person}{Jian-tao Sun}, {and}
  \bibinfo{person}{Zheng Chen}.} \bibinfo{year}{2009}\natexlab{}.
\newblock \showarticletitle{Understanding user's query intent with wikipedia}.
  In \bibinfo{booktitle}{{\em WWW}}.
\newblock


\bibitem[\protect\citeauthoryear{Kingma and Ba}{Kingma and Ba}{2015}]%
        {kingma2014adam}
\bibfield{author}{\bibinfo{person}{Diederik~P Kingma} {and}
  \bibinfo{person}{Jimmy Ba}.} \bibinfo{year}{2015}\natexlab{}.
\newblock \showarticletitle{Adam: A method for stochastic optimization}. In
  \bibinfo{booktitle}{{\em ICLR}}.
\newblock


\bibitem[\protect\citeauthoryear{Kolluru and Mukherjee}{Kolluru and
  Mukherjee}{2016}]%
        {kolluru2016query}
\bibfield{author}{\bibinfo{person}{SK Kolluru} {and} \bibinfo{person}{Prasenjit
  Mukherjee}.} \bibinfo{year}{2016}\natexlab{}.
\newblock \showarticletitle{Query Clustering using Segment Specific Context
  Embeddings}.
\newblock \bibinfo{journal}{{\em arXiv preprint arXiv:1608.01247\/}}
  (\bibinfo{year}{2016}).
\newblock


\bibitem[\protect\citeauthoryear{Lebanoff, Song, and Liu}{Lebanoff
  et~al\mbox{.}}{2018}]%
        {lebanoff2018adapting}
\bibfield{author}{\bibinfo{person}{Logan Lebanoff}, \bibinfo{person}{Kaiqiang
  Song}, {and} \bibinfo{person}{Fei Liu}.} \bibinfo{year}{2018}\natexlab{}.
\newblock \showarticletitle{Adapting the neural encoder-decoder framework from
  single to multi-document summarization}. In \bibinfo{booktitle}{{\em EMNLP}}.
\newblock


\bibitem[\protect\citeauthoryear{Lee, Gao, and Zhang}{Lee
  et~al\mbox{.}}{2018}]%
        {lee2018rare}
\bibfield{author}{\bibinfo{person}{Mu-Chu Lee}, \bibinfo{person}{Bin Gao},
  {and} \bibinfo{person}{Ruofei Zhang}.} \bibinfo{year}{2018}\natexlab{}.
\newblock \showarticletitle{Rare query expansion through generative adversarial
  networks in search advertising}. In \bibinfo{booktitle}{{\em SIGKDD}}.
\newblock


\bibitem[\protect\citeauthoryear{Lee, Liu, and Cho}{Lee et~al\mbox{.}}{2005}]%
        {lee2005automatic}
\bibfield{author}{\bibinfo{person}{Uichin Lee}, \bibinfo{person}{Zhenyu Liu},
  {and} \bibinfo{person}{Junghoo Cho}.} \bibinfo{year}{2005}\natexlab{}.
\newblock \showarticletitle{Automatic identification of user goals in web
  search}. In \bibinfo{booktitle}{{\em WWW}}.
\newblock


\bibitem[\protect\citeauthoryear{Li, Qian, and Liu}{Li et~al\mbox{.}}{2013}]%
        {li2013using}
\bibfield{author}{\bibinfo{person}{Chen Li}, \bibinfo{person}{Xian Qian}, {and}
  \bibinfo{person}{Yang Liu}.} \bibinfo{year}{2013}\natexlab{}.
\newblock \showarticletitle{Using supervised bigram-based ILP for extractive
  summarization}. In \bibinfo{booktitle}{{\em ACL}}.
\newblock


\bibitem[\protect\citeauthoryear{Lin}{Lin}{2004}]%
        {lin2004rouge}
\bibfield{author}{\bibinfo{person}{Chin-Yew Lin}.}
  \bibinfo{year}{2004}\natexlab{}.
\newblock \showarticletitle{Rouge: A package for automatic evaluation of
  summaries}. In \bibinfo{booktitle}{{\em Text summarization branches out}}.
\newblock


\bibitem[\protect\citeauthoryear{Maron and Kuhns}{Maron and Kuhns}{1960}]%
        {maron1960relevance}
\bibfield{author}{\bibinfo{person}{Melvin~Earl Maron} {and}
  \bibinfo{person}{John~Larry Kuhns}.} \bibinfo{year}{1960}\natexlab{}.
\newblock \showarticletitle{On relevance, probabilistic indexing and
  information retrieval}.
\newblock \bibinfo{journal}{{\em JACM\/}} (\bibinfo{year}{1960}).
\newblock


\bibitem[\protect\citeauthoryear{Meng, Huang, and Gu}{Meng
  et~al\mbox{.}}{2013}]%
        {meng2013new}
\bibfield{author}{\bibinfo{person}{Lingling Meng}, \bibinfo{person}{Runqing
  Huang}, {and} \bibinfo{person}{Junzhong Gu}.}
  \bibinfo{year}{2013}\natexlab{}.
\newblock \showarticletitle{A new algorithm of web queries clustering using
  user feedback}.
\newblock \bibinfo{journal}{{\em International Journal of Signal Processing,
  Image Processing and Pattern Recognition\/}} (\bibinfo{year}{2013}).
\newblock


\bibitem[\protect\citeauthoryear{Mihalcea and Tarau}{Mihalcea and
  Tarau}{2004}]%
        {mihalcea2004textrank}
\bibfield{author}{\bibinfo{person}{Rada Mihalcea} {and} \bibinfo{person}{Paul
  Tarau}.} \bibinfo{year}{2004}\natexlab{}.
\newblock \showarticletitle{Textrank: Bringing order into text}. In
  \bibinfo{booktitle}{{\em EMNLP}}.
\newblock


\bibitem[\protect\citeauthoryear{Mohamed and Rajasekaran}{Mohamed and
  Rajasekaran}{2006}]%
        {mohamed2006improving}
\bibfield{author}{\bibinfo{person}{Ahmed~A Mohamed} {and}
  \bibinfo{person}{Sanguthevar Rajasekaran}.} \bibinfo{year}{2006}\natexlab{}.
\newblock \showarticletitle{Improving query-based summarization using document
  graphs}. In \bibinfo{booktitle}{{\em ISSPIT}}.
\newblock


\bibitem[\protect\citeauthoryear{Nakov, Hoogeveen, M{\`a}rquez, Moschitti,
  Mubarak, Baldwin, and Verspoor}{Nakov et~al\mbox{.}}{2017}]%
        {nakov2017semeval}
\bibfield{author}{\bibinfo{person}{Preslav Nakov}, \bibinfo{person}{Doris
  Hoogeveen}, \bibinfo{person}{Llu{\'\i}s M{\`a}rquez},
  \bibinfo{person}{Alessandro Moschitti}, \bibinfo{person}{Hamdy Mubarak},
  \bibinfo{person}{Timothy Baldwin}, {and} \bibinfo{person}{Karin Verspoor}.}
  \bibinfo{year}{2017}\natexlab{}.
\newblock \showarticletitle{SemEval-2017 task 3: Community question answering}.
  In \bibinfo{booktitle}{{\em SemEval}}.
\newblock


\bibitem[\protect\citeauthoryear{Nakov, M{\`a}rquez, Magdy, Moschitti, Glass,
  and Randeree}{Nakov et~al\mbox{.}}{2015}]%
        {nakov-etal-2015-semeval}
\bibfield{author}{\bibinfo{person}{Preslav Nakov}, \bibinfo{person}{Llu{\'\i}s
  M{\`a}rquez}, \bibinfo{person}{Walid Magdy}, \bibinfo{person}{Alessandro
  Moschitti}, \bibinfo{person}{James Glass}, {and} \bibinfo{person}{Bilal
  Randeree}.} \bibinfo{year}{2015}\natexlab{}.
\newblock \showarticletitle{SemEval-2015 Task 3: Answer Selection in Community
  Question Answering}. In \bibinfo{booktitle}{{\em SemEval}}.
\newblock


\bibitem[\protect\citeauthoryear{Nema, Khapra, Laha, and Ravindran}{Nema
  et~al\mbox{.}}{2017}]%
        {nema2017diversity}
\bibfield{author}{\bibinfo{person}{Preksha Nema}, \bibinfo{person}{Mitesh
  Khapra}, \bibinfo{person}{Anirban Laha}, {and} \bibinfo{person}{Balaraman
  Ravindran}.} \bibinfo{year}{2017}\natexlab{}.
\newblock \showarticletitle{Diversity driven attention model for query-based
  abstractive summarization}. In \bibinfo{booktitle}{{\em ACL}}.
\newblock


\bibitem[\protect\citeauthoryear{Otsuka, Nishida, Bessho, Asano, and
  Tomita}{Otsuka et~al\mbox{.}}{2018}]%
        {otsuka2018query}
\bibfield{author}{\bibinfo{person}{Atsushi Otsuka}, \bibinfo{person}{Kyosuke
  Nishida}, \bibinfo{person}{Katsuji Bessho}, \bibinfo{person}{Hisako Asano},
  {and} \bibinfo{person}{Junji Tomita}.} \bibinfo{year}{2018}\natexlab{}.
\newblock \showarticletitle{Query expansion with neural question-to-answer
  translation for FAQ-based question answering}. In \bibinfo{booktitle}{{\em
  WebConf}}.
\newblock


\bibitem[\protect\citeauthoryear{Radev, Jing, Sty{\'s}, and Tam}{Radev
  et~al\mbox{.}}{2004}]%
        {radev2004centroid}
\bibfield{author}{\bibinfo{person}{Dragomir~R Radev}, \bibinfo{person}{Hongyan
  Jing}, \bibinfo{person}{Ma{\l}gorzata Sty{\'s}}, {and}
  \bibinfo{person}{Daniel Tam}.} \bibinfo{year}{2004}\natexlab{}.
\newblock \showarticletitle{Centroid-based summarization of multiple
  documents}.
\newblock \bibinfo{journal}{{\em Information Processing \& Management\/}}
  (\bibinfo{year}{2004}).
\newblock


\bibitem[\protect\citeauthoryear{Rose and Levinson}{Rose and Levinson}{2004}]%
        {rose2004understanding}
\bibfield{author}{\bibinfo{person}{Daniel~E Rose} {and} \bibinfo{person}{Danny
  Levinson}.} \bibinfo{year}{2004}\natexlab{}.
\newblock \showarticletitle{Understanding user goals in web search}. In
  \bibinfo{booktitle}{{\em WWW}}.
\newblock


\bibitem[\protect\citeauthoryear{Rush, Chopra, and Weston}{Rush
  et~al\mbox{.}}{2015}]%
        {rush2015neural}
\bibfield{author}{\bibinfo{person}{Alexander~M Rush}, \bibinfo{person}{Sumit
  Chopra}, {and} \bibinfo{person}{Jason Weston}.}
  \bibinfo{year}{2015}\natexlab{}.
\newblock \showarticletitle{A neural attention model for abstractive sentence
  summarization}. In \bibinfo{booktitle}{{\em ACL}}.
\newblock


\bibitem[\protect\citeauthoryear{Schilder and Kondadadi}{Schilder and
  Kondadadi}{2008}]%
        {schilder2008fastsum}
\bibfield{author}{\bibinfo{person}{Frank Schilder} {and}
  \bibinfo{person}{Ravikumar Kondadadi}.} \bibinfo{year}{2008}\natexlab{}.
\newblock \showarticletitle{FastSum: fast and accurate query-based
  multi-document summarization}. In \bibinfo{booktitle}{{\em ACL}}.
\newblock


\bibitem[\protect\citeauthoryear{Shen, Pan, Sun, Pan, Wu, Yin, and Yang}{Shen
  et~al\mbox{.}}{2006a}]%
        {shen2006query}
\bibfield{author}{\bibinfo{person}{Dou Shen}, \bibinfo{person}{Rong Pan},
  \bibinfo{person}{Jian-Tao Sun}, \bibinfo{person}{Jeffrey~Junfeng Pan},
  \bibinfo{person}{Kangheng Wu}, \bibinfo{person}{Jie Yin}, {and}
  \bibinfo{person}{Qiang Yang}.} \bibinfo{year}{2006}\natexlab{a}.
\newblock \showarticletitle{Query enrichment for web-query classification}.
\newblock \bibinfo{journal}{{\em TOIS\/}} \bibinfo{volume}{24},
  \bibinfo{number}{3} (\bibinfo{year}{2006}), \bibinfo{pages}{320--352}.
\newblock


\bibitem[\protect\citeauthoryear{Shen, Sun, Yang, and Chen}{Shen
  et~al\mbox{.}}{2006b}]%
        {shen2006building}
\bibfield{author}{\bibinfo{person}{Dou Shen}, \bibinfo{person}{Jian-Tao Sun},
  \bibinfo{person}{Qiang Yang}, {and} \bibinfo{person}{Zheng Chen}.}
  \bibinfo{year}{2006}\natexlab{b}.
\newblock \showarticletitle{Building bridges for web query classification}. In
  \bibinfo{booktitle}{{\em SIGIR}}.
\newblock


\bibitem[\protect\citeauthoryear{Shi, Yao, Tian, and Jiang}{Shi
  et~al\mbox{.}}{2016}]%
        {shi2016deep}
\bibfield{author}{\bibinfo{person}{Yangyang Shi}, \bibinfo{person}{Kaisheng
  Yao}, \bibinfo{person}{Le Tian}, {and} \bibinfo{person}{Daxin Jiang}.}
  \bibinfo{year}{2016}\natexlab{}.
\newblock \showarticletitle{Deep LSTM based feature mapping for query
  classification}. In \bibinfo{booktitle}{{\em NAACL}}.
\newblock


\bibitem[\protect\citeauthoryear{Singh and Sharan}{Singh and Sharan}{2015a}]%
        {singh2015co}
\bibfield{author}{\bibinfo{person}{Jagendra Singh} {and} \bibinfo{person}{Aditi
  Sharan}.} \bibinfo{year}{2015}\natexlab{a}.
\newblock \showarticletitle{Co-occurrence and semantic similarity based hybrid
  approach for improving automatic query expansion in information retrieval}.
  In \bibinfo{booktitle}{{\em International Conference on Distributed Computing
  and Internet Technology}}.
\newblock


\bibitem[\protect\citeauthoryear{Singh and Sharan}{Singh and Sharan}{2015b}]%
        {singh2015context}
\bibfield{author}{\bibinfo{person}{Jagendra Singh} {and} \bibinfo{person}{Aditi
  Sharan}.} \bibinfo{year}{2015}\natexlab{b}.
\newblock \showarticletitle{Context window based co-occurrence approach for
  improving feedback based query expansion in information retrieval}.
\newblock \bibinfo{journal}{{\em IJIRR\/}} (\bibinfo{year}{2015}).
\newblock


\bibitem[\protect\citeauthoryear{Sutskever, Vinyals, and Le}{Sutskever
  et~al\mbox{.}}{2014}]%
        {sutskever2014sequence}
\bibfield{author}{\bibinfo{person}{I Sutskever}, \bibinfo{person}{O Vinyals},
  {and} \bibinfo{person}{QV Le}.} \bibinfo{year}{2014}\natexlab{}.
\newblock \showarticletitle{Sequence to sequence learning with neural
  networks}.
\newblock \bibinfo{journal}{{\em NIPS\/}} (\bibinfo{year}{2014}).
\newblock


\bibitem[\protect\citeauthoryear{Tan, Wan, and Xiao}{Tan et~al\mbox{.}}{[n.
  d.]}]%
        {tan2017abstractive}
\bibfield{author}{\bibinfo{person}{Jiwei Tan}, \bibinfo{person}{Xiaojun Wan},
  {and} \bibinfo{person}{Jianguo Xiao}.} \bibinfo{year}{[n. d.]}\natexlab{}.
\newblock \showarticletitle{Abstractive document summarization with a
  graph-based attentional neural model}. In \bibinfo{booktitle}{{\em ACL}}.
\newblock


\bibitem[\protect\citeauthoryear{Tran, Cimiano, Rudolph, and Studer}{Tran
  et~al\mbox{.}}{2007}]%
        {tran2007ontology}
\bibfield{author}{\bibinfo{person}{Thanh Tran}, \bibinfo{person}{Philipp
  Cimiano}, \bibinfo{person}{Sebastian Rudolph}, {and} \bibinfo{person}{Rudi
  Studer}.} \bibinfo{year}{2007}\natexlab{}.
\newblock \showarticletitle{Ontology-based interpretation of keywords for
  semantic search}.
\newblock In \bibinfo{booktitle}{{\em The Semantic Web}}.
\newblock


\bibitem[\protect\citeauthoryear{Voorhees}{Voorhees}{1994}]%
        {voorhees1994query}
\bibfield{author}{\bibinfo{person}{Ellen~M Voorhees}.}
  \bibinfo{year}{1994}\natexlab{}.
\newblock \showarticletitle{Query expansion using lexical-semantic relations}.
  In \bibinfo{booktitle}{{\em SIGIR}}.
\newblock


\bibitem[\protect\citeauthoryear{Wen, Nie, and Zhang}{Wen
  et~al\mbox{.}}{2002}]%
        {wen2002query}
\bibfield{author}{\bibinfo{person}{Ji-Rong Wen}, \bibinfo{person}{Jian-Yun
  Nie}, {and} \bibinfo{person}{Hong-Jiang Zhang}.}
  \bibinfo{year}{2002}\natexlab{}.
\newblock \showarticletitle{Query clustering using user logs}.
\newblock \bibinfo{journal}{{\em TOIS\/}} (\bibinfo{year}{2002}).
\newblock


\bibitem[\protect\citeauthoryear{Yom-Tov, Fine, Carmel, and Darlow}{Yom-Tov
  et~al\mbox{.}}{2005}]%
        {yom2005learning}
\bibfield{author}{\bibinfo{person}{Elad Yom-Tov}, \bibinfo{person}{Shai Fine},
  \bibinfo{person}{David Carmel}, {and} \bibinfo{person}{Adam Darlow}.}
  \bibinfo{year}{2005}\natexlab{}.
\newblock \showarticletitle{Learning to estimate query difficulty: including
  applications to missing content detection and distributed information
  retrieval}. In \bibinfo{booktitle}{{\em SIGIR}}.
\newblock


\bibitem[\protect\citeauthoryear{Zhang, Deng, and Li}{Zhang
  et~al\mbox{.}}{2009}]%
        {zhang2009concept}
\bibfield{author}{\bibinfo{person}{Jiuling Zhang}, \bibinfo{person}{Beixing
  Deng}, {and} \bibinfo{person}{Xing Li}.} \bibinfo{year}{2009}\natexlab{}.
\newblock \showarticletitle{Concept based query expansion using wordnet}. In
  \bibinfo{booktitle}{{\em AST}}.
\newblock


\bibitem[\protect\citeauthoryear{Zhang, Tan, and Wan}{Zhang
  et~al\mbox{.}}{2018}]%
        {zhang2018adapting}
\bibfield{author}{\bibinfo{person}{Jianmin Zhang}, \bibinfo{person}{Jiwei Tan},
  {and} \bibinfo{person}{Xiaojun Wan}.} \bibinfo{year}{2018}\natexlab{}.
\newblock \showarticletitle{Adapting neural single-document summarization model
  for abstractive multi-document summarization: A pilot study}. In
  \bibinfo{booktitle}{{\em Proceedings of the 11th International Conference on
  Natural Language Generation}}.
\newblock


\bibitem[\protect\citeauthoryear{Zhang, Wang, Si, and Gao}{Zhang
  et~al\mbox{.}}{2016}]%
        {zhang2016learning}
\bibfield{author}{\bibinfo{person}{Zhiwei Zhang}, \bibinfo{person}{Qifan Wang},
  \bibinfo{person}{Luo Si}, {and} \bibinfo{person}{Jianfeng Gao}.}
  \bibinfo{year}{2016}\natexlab{}.
\newblock \showarticletitle{Learning for efficient supervised query expansion
  via two-stage feature selection}. In \bibinfo{booktitle}{{\em SIGIR}}.
\newblock


\end{thebibliography}



\begin{thebibliography}{68}


\ifx \showCODEN    \undefined \def \showCODEN     #1{\unskip}     \fi
\ifx \showDOI      \undefined \def \showDOI       #1{#1}\fi
\ifx \showISBNx    \undefined \def \showISBNx     #1{\unskip}     \fi
\ifx \showISBNxiii \undefined \def \showISBNxiii  #1{\unskip}     \fi
\ifx \showISSN     \undefined \def \showISSN      #1{\unskip}     \fi
\ifx \showLCCN     \undefined \def \showLCCN      #1{\unskip}     \fi
\ifx \shownote     \undefined \def \shownote      #1{#1}          \fi
\ifx \showarticletitle \undefined \def \showarticletitle #1{#1}   \fi
\ifx \showURL      \undefined \def \showURL       {\relax}        \fi
\providecommand\bibfield[2]{#2}
\providecommand\bibinfo[2]{#2}
\providecommand\natexlab[1]{#1}
\providecommand\showeprint[2][]{arXiv:#2}

\bibitem[\protect\citeauthoryear{Baeza-Yates, Calder{\'o}n-Benavides, and
  Gonz{\'a}lez-Caro}{Baeza-Yates et~al\mbox{.}}{2006}]%
        {baeza2006intention}
\bibfield{author}{\bibinfo{person}{Ricardo Baeza-Yates},
  \bibinfo{person}{Liliana Calder{\'o}n-Benavides}, {and}
  \bibinfo{person}{Cristina Gonz{\'a}lez-Caro}.}
  \bibinfo{year}{2006}\natexlab{}.
\newblock \showarticletitle{The intention behind web queries}. In
  \bibinfo{booktitle}{{\em SPIRE}}.
\newblock


\bibitem[\protect\citeauthoryear{Bahdanau, Cho, and Bengio}{Bahdanau
  et~al\mbox{.}}{2015}]%
        {bahdanau2014neural}
\bibfield{author}{\bibinfo{person}{Dzmitry Bahdanau},
  \bibinfo{person}{Kyunghyun Cho}, {and} \bibinfo{person}{Yoshua Bengio}.}
  \bibinfo{year}{2015}\natexlab{}.
\newblock \showarticletitle{Neural machine translation by jointly learning to
  align and translate}. In \bibinfo{booktitle}{{\em ICLR}}.
\newblock


\bibitem[\protect\citeauthoryear{Baumel, Eyal, and Elhadad}{Baumel
  et~al\mbox{.}}{2018}]%
        {baumel2018query}
\bibfield{author}{\bibinfo{person}{Tal Baumel}, \bibinfo{person}{Matan Eyal},
  {and} \bibinfo{person}{Michael Elhadad}.} \bibinfo{year}{2018}\natexlab{}.
\newblock \showarticletitle{Query focused abstractive summarization:
  Incorporating query relevance, multi-document coverage, and summary length
  constraints into seq2seq models}. In \bibinfo{booktitle}{{\em arXiv}}.
\newblock


\bibitem[\protect\citeauthoryear{Beeferman and Berger}{Beeferman and
  Berger}{2000}]%
        {beeferman2000agglomerative}
\bibfield{author}{\bibinfo{person}{Doug Beeferman} {and} \bibinfo{person}{Adam
  Berger}.} \bibinfo{year}{2000}\natexlab{}.
\newblock \showarticletitle{Agglomerative clustering of a search engine query
  log}. In \bibinfo{booktitle}{{\em KDD}}.
\newblock


\bibitem[\protect\citeauthoryear{Beitzel, Jensen, Frieder, Lewis, Chowdhury,
  and Kolcz}{Beitzel et~al\mbox{.}}{2005}]%
        {beitzel2005improving}
\bibfield{author}{\bibinfo{person}{Steven~M Beitzel}, \bibinfo{person}{Eric~C
  Jensen}, \bibinfo{person}{Ophir Frieder}, \bibinfo{person}{David~D Lewis},
  \bibinfo{person}{Abdur Chowdhury}, {and} \bibinfo{person}{Aleksander Kolcz}.}
  \bibinfo{year}{2005}\natexlab{}.
\newblock \showarticletitle{Improving automatic query classification via
  semi-supervised learning}. In \bibinfo{booktitle}{{\em ICDM}}.
\newblock


\bibitem[\protect\citeauthoryear{Broder}{Broder}{2002}]%
        {broder2002taxonomy}
\bibfield{author}{\bibinfo{person}{Andrei Broder}.}
  \bibinfo{year}{2002}\natexlab{}.
\newblock \showarticletitle{A taxonomy of web search}. In
  \bibinfo{booktitle}{{\em ACM Sigir forum}}.
\newblock


\bibitem[\protect\citeauthoryear{Cao, Hu, Shen, Jiang, Sun, Chen, and Yang}{Cao
  et~al\mbox{.}}{2009}]%
        {cao2009context}
\bibfield{author}{\bibinfo{person}{Huanhuan Cao}, \bibinfo{person}{Derek~Hao
  Hu}, \bibinfo{person}{Dou Shen}, \bibinfo{person}{Daxin Jiang},
  \bibinfo{person}{Jian-Tao Sun}, \bibinfo{person}{Enhong Chen}, {and}
  \bibinfo{person}{Qiang Yang}.} \bibinfo{year}{2009}\natexlab{}.
\newblock \showarticletitle{Context-aware query classification}. In
  \bibinfo{booktitle}{{\em SIGIR}}.
\newblock


\bibitem[\protect\citeauthoryear{Carbonell and Goldstein}{Carbonell and
  Goldstein}{1998}]%
        {carbonell1998use}
\bibfield{author}{\bibinfo{person}{Jaime~G Carbonell} {and}
  \bibinfo{person}{Jade Goldstein}.} \bibinfo{year}{1998}\natexlab{}.
\newblock \showarticletitle{The use of MMR, diversity-based reranking for
  reordering documents and producing summaries.}. In \bibinfo{booktitle}{{\em
  SIGIR}}.
\newblock


\bibitem[\protect\citeauthoryear{Chen, Xu, Shaoyun, and Yongfeng}{Chen
  et~al\mbox{.}}{2019}]%
        {NLE2019SIGIR}
\bibfield{author}{\bibinfo{person}{Hanxiong Chen}, \bibinfo{person}{Chen Xu},
  \bibinfo{person}{Shi Shaoyun}, {and} \bibinfo{person}{Zhang Yongfeng}.}
  \bibinfo{year}{2019}\natexlab{}.
\newblock \showarticletitle{Generate Natural Language Explanations for
  Recommendation}. In \bibinfo{booktitle}{{\em SIGIR}}.
\newblock


\bibitem[\protect\citeauthoryear{Cho, Van~Merri{\"e}nboer, Gulcehre, Bahdanau,
  Bougares, Schwenk, and Bengio}{Cho et~al\mbox{.}}{2014}]%
        {cho2014learning}
\bibfield{author}{\bibinfo{person}{Kyunghyun Cho}, \bibinfo{person}{Bart
  Van~Merri{\"e}nboer}, \bibinfo{person}{Caglar Gulcehre},
  \bibinfo{person}{Dzmitry Bahdanau}, \bibinfo{person}{Fethi Bougares},
  \bibinfo{person}{Holger Schwenk}, {and} \bibinfo{person}{Yoshua Bengio}.}
  \bibinfo{year}{2014}\natexlab{}.
\newblock \showarticletitle{Learning phrase representations using RNN
  encoder-decoder for statistical machine translation}. In
  \bibinfo{booktitle}{{\em EMNLP}}.
\newblock


\bibitem[\protect\citeauthoryear{Cho, Lebanoff, Foroosh, and Liu}{Cho
  et~al\mbox{.}}{2019}]%
        {cho2019improving}
\bibfield{author}{\bibinfo{person}{Sangwoo Cho}, \bibinfo{person}{Logan
  Lebanoff}, \bibinfo{person}{Hassan Foroosh}, {and} \bibinfo{person}{Fei
  Liu}.} \bibinfo{year}{2019}\natexlab{}.
\newblock \showarticletitle{Improving the Similarity Measure of Determinantal
  Point Processes for Extractive Multi-Document Summarization}. In
  \bibinfo{booktitle}{{\em ACL}}.
\newblock


\bibitem[\protect\citeauthoryear{Chopra, Auli, and Rush}{Chopra
  et~al\mbox{.}}{2016}]%
        {chopra2016abstractive}
\bibfield{author}{\bibinfo{person}{Sumit Chopra}, \bibinfo{person}{Michael
  Auli}, {and} \bibinfo{person}{Alexander~M Rush}.}
  \bibinfo{year}{2016}\natexlab{}.
\newblock \showarticletitle{Abstractive sentence summarization with attentive
  recurrent neural networks}. In \bibinfo{booktitle}{{\em NAACL}}.
\newblock


\bibitem[\protect\citeauthoryear{Costa, Ouyang, Dolog, and Lawlor}{Costa
  et~al\mbox{.}}{2017}]%
        {Costa2017Automatic}
\bibfield{author}{\bibinfo{person}{Felipe Costa}, \bibinfo{person}{Sixun
  Ouyang}, \bibinfo{person}{Peter Dolog}, {and} \bibinfo{person}{Aonghus
  Lawlor}.} \bibinfo{year}{2017}\natexlab{}.
\newblock \showarticletitle{Automatic Generation of Natural Language
  Explanations}.
\newblock  (\bibinfo{year}{2017}).
\newblock


\bibitem[\protect\citeauthoryear{Cui, Wen, Nie, and Ma}{Cui
  et~al\mbox{.}}{2002}]%
        {cui2002probabilistic}
\bibfield{author}{\bibinfo{person}{Hang Cui}, \bibinfo{person}{Ji-Rong Wen},
  \bibinfo{person}{Jian-Yun Nie}, {and} \bibinfo{person}{Wei-Ying Ma}.}
  \bibinfo{year}{2002}\natexlab{}.
\newblock \showarticletitle{Probabilistic query expansion using query logs}. In
  \bibinfo{booktitle}{{\em WWW}}.
\newblock


\bibitem[\protect\citeauthoryear{Dang and Croft}{Dang and Croft}{2010}]%
        {dang2010query}
\bibfield{author}{\bibinfo{person}{Van Dang} {and} \bibinfo{person}{Bruce~W
  Croft}.} \bibinfo{year}{2010}\natexlab{}.
\newblock \showarticletitle{Query reformulation using anchor text}. In
  \bibinfo{booktitle}{{\em WSDM}}.
\newblock


\bibitem[\protect\citeauthoryear{Daum{\'e}~III and Marcu}{Daum{\'e}~III and
  Marcu}{2006}]%
        {daume2006bayesian}
\bibfield{author}{\bibinfo{person}{Hal Daum{\'e}~III} {and}
  \bibinfo{person}{Daniel Marcu}.} \bibinfo{year}{2006}\natexlab{}.
\newblock \showarticletitle{Bayesian query-focused summarization}. In
  \bibinfo{booktitle}{{\em ACL}}. \bibinfo{pages}{305--312}.
\newblock


\bibitem[\protect\citeauthoryear{Doshi-Velez and Kim}{Doshi-Velez and
  Kim}{2017}]%
        {Doshi2017Towards}
\bibfield{author}{\bibinfo{person}{Finale Doshi-Velez} {and}
  \bibinfo{person}{Been Kim}.} \bibinfo{year}{2017}\natexlab{}.
\newblock \showarticletitle{Towards A Rigorous Science of Interpretable Machine
  Learning}.
\newblock  (\bibinfo{year}{2017}).
\newblock


\bibitem[\protect\citeauthoryear{Duan, Kiciman, and Zhai}{Duan
  et~al\mbox{.}}{2012}]%
        {Duan2012ClickPA}
\bibfield{author}{\bibinfo{person}{Huizhong Duan}, \bibinfo{person}{Emre
  Kiciman}, {and} \bibinfo{person}{ChengXiang Zhai}.}
  \bibinfo{year}{2012}\natexlab{}.
\newblock \showarticletitle{Click patterns: an empirical representation of
  complex query intents}. In \bibinfo{booktitle}{{\em CIKM}}.
\newblock


\bibitem[\protect\citeauthoryear{Erkan and Radev}{Erkan and Radev}{2004}]%
        {erkan2004lexrank}
\bibfield{author}{\bibinfo{person}{G{\"u}nes Erkan} {and}
  \bibinfo{person}{Dragomir~R Radev}.} \bibinfo{year}{2004}\natexlab{}.
\newblock \showarticletitle{Lexrank: Graph-based lexical centrality as salience
  in text summarization}.
\newblock \bibinfo{journal}{{\em Journal of artificial intelligence
  research\/}} (\bibinfo{year}{2004}).
\newblock


\bibitem[\protect\citeauthoryear{Fabbri, Li, She, Li, and Radev}{Fabbri
  et~al\mbox{.}}{2019}]%
        {fabbri2019multi}
\bibfield{author}{\bibinfo{person}{Alexander~R Fabbri}, \bibinfo{person}{Irene
  Li}, \bibinfo{person}{Tianwei She}, \bibinfo{person}{Suyi Li}, {and}
  \bibinfo{person}{Dragomir~R Radev}.} \bibinfo{year}{2019}\natexlab{}.
\newblock \showarticletitle{Multi-News: a Large-Scale Multi-Document
  Summarization Dataset and Abstractive Hierarchical Model}. In
  \bibinfo{booktitle}{{\em ACL}}.
\newblock


\bibitem[\protect\citeauthoryear{Fang}{Fang}{2008}]%
        {fang2008re}
\bibfield{author}{\bibinfo{person}{Hui Fang}.} \bibinfo{year}{2008}\natexlab{}.
\newblock \showarticletitle{A re-examination of query expansion using lexical
  resources}. In \bibinfo{booktitle}{{\em ACL}}.
\newblock


\bibitem[\protect\citeauthoryear{Goldstein, Kantrowitz, Mittal, and
  Carbonell}{Goldstein et~al\mbox{.}}{1999}]%
        {goldstein1999summarizing}
\bibfield{author}{\bibinfo{person}{Jade Goldstein}, \bibinfo{person}{Mark
  Kantrowitz}, \bibinfo{person}{Vibhu Mittal}, {and} \bibinfo{person}{Jaime
  Carbonell}.} \bibinfo{year}{1999}\natexlab{}.
\newblock \showarticletitle{Summarizing text documents: sentence selection and
  evaluation metrics}. In \bibinfo{booktitle}{{\em SIGIR}}.
\newblock


\bibitem[\protect\citeauthoryear{Gravano, Hatzivassiloglou, and
  Lichtenstein}{Gravano et~al\mbox{.}}{2003}]%
        {gravano2003categorizing}
\bibfield{author}{\bibinfo{person}{Luis Gravano}, \bibinfo{person}{Vasileios
  Hatzivassiloglou}, {and} \bibinfo{person}{Richard Lichtenstein}.}
  \bibinfo{year}{2003}\natexlab{}.
\newblock \showarticletitle{Categorizing web queries according to geographical
  locality}. In \bibinfo{booktitle}{{\em CIKM}}.
\newblock


\bibitem[\protect\citeauthoryear{Hasselqvist, Helmertz, and
  K{\aa}geb{\"a}ck}{Hasselqvist et~al\mbox{.}}{2017}]%
        {hasselqvist2017query}
\bibfield{author}{\bibinfo{person}{Johan Hasselqvist}, \bibinfo{person}{Niklas
  Helmertz}, {and} \bibinfo{person}{Mikael K{\aa}geb{\"a}ck}.}
  \bibinfo{year}{2017}\natexlab{}.
\newblock \showarticletitle{Query-based abstractive summarization using neural
  networks}. In \bibinfo{booktitle}{{\em arXiv}}.
\newblock


\bibitem[\protect\citeauthoryear{Hong, Vaidya, Lu, and Liu}{Hong
  et~al\mbox{.}}{2016}]%
        {hong2016accurate}
\bibfield{author}{\bibinfo{person}{Yuan Hong}, \bibinfo{person}{Jaideep
  Vaidya}, \bibinfo{person}{Haibing Lu}, {and} \bibinfo{person}{Wen~Ming Liu}.}
  \bibinfo{year}{2016}\natexlab{}.
\newblock \showarticletitle{Accurate and efficient query clustering via top
  ranked search results}. In \bibinfo{booktitle}{{\em Web Intelligence}}.
\newblock


\bibitem[\protect\citeauthoryear{Hsu, Lin, Lee, Min, Tang, and Sun}{Hsu
  et~al\mbox{.}}{2018}]%
        {hsu2018unified}
\bibfield{author}{\bibinfo{person}{Wan-Ting Hsu}, \bibinfo{person}{Chieh-Kai
  Lin}, \bibinfo{person}{Ming-Ying Lee}, \bibinfo{person}{Kerui Min},
  \bibinfo{person}{Jing Tang}, {and} \bibinfo{person}{Min Sun}.}
  \bibinfo{year}{2018}\natexlab{}.
\newblock \showarticletitle{A unified model for extractive and abstractive
  summarization using inconsistency loss}. In \bibinfo{booktitle}{{\em ACL}}.
\newblock


\bibitem[\protect\citeauthoryear{Hu, Wang, Lochovsky, Sun, and Chen}{Hu
  et~al\mbox{.}}{2009}]%
        {hu2009understanding}
\bibfield{author}{\bibinfo{person}{Jian Hu}, \bibinfo{person}{Gang Wang},
  \bibinfo{person}{Fred Lochovsky}, \bibinfo{person}{Jian-tao Sun}, {and}
  \bibinfo{person}{Zheng Chen}.} \bibinfo{year}{2009}\natexlab{}.
\newblock \showarticletitle{Understanding user's query intent with wikipedia}.
  In \bibinfo{booktitle}{{\em WWW}}.
\newblock


\bibitem[\protect\citeauthoryear{Jones and Diaz}{Jones and Diaz}{2007}]%
        {jones2007temporal}
\bibfield{author}{\bibinfo{person}{Rosie Jones} {and} \bibinfo{person}{Fernando
  Diaz}.} \bibinfo{year}{2007}\natexlab{}.
\newblock \showarticletitle{Temporal profiles of queries}.
\newblock \bibinfo{journal}{{\em TOIS\/}} (\bibinfo{year}{2007}).
\newblock


\bibitem[\protect\citeauthoryear{Kardkov{\'a}cs, Tikk, and
  B{\'a}ns{\'a}ghi}{Kardkov{\'a}cs et~al\mbox{.}}{2005}]%
        {kardkovacs2005ferrety}
\bibfield{author}{\bibinfo{person}{Zsolt~T Kardkov{\'a}cs},
  \bibinfo{person}{Domonkos Tikk}, {and} \bibinfo{person}{Zolt{\'a}n
  B{\'a}ns{\'a}ghi}.} \bibinfo{year}{2005}\natexlab{}.
\newblock \showarticletitle{The ferrety algorithm for the KDD Cup 2005
  problem}.
\newblock \bibinfo{journal}{{\em SIGKDD\/}} (\bibinfo{year}{2005}).
\newblock


\bibitem[\protect\citeauthoryear{Karimzadehgan and Zhai}{Karimzadehgan and
  Zhai}{2011}]%
        {Karimzadehgan2011Improving}
\bibfield{author}{\bibinfo{person}{Maryam Karimzadehgan} {and}
  \bibinfo{person}{Cheng~Xiang Zhai}.} \bibinfo{year}{2011}\natexlab{}.
\newblock \showarticletitle{Improving retrieval accuracy of difficult queries
  through generalizing negative document language models}. In
  \bibinfo{booktitle}{{\em CIKM}}.
\newblock


\bibitem[\protect\citeauthoryear{Kingma and Ba}{Kingma and Ba}{2015}]%
        {kingma2014adam}
\bibfield{author}{\bibinfo{person}{Diederik~P Kingma} {and}
  \bibinfo{person}{Jimmy Ba}.} \bibinfo{year}{2015}\natexlab{}.
\newblock \showarticletitle{Adam: A method for stochastic optimization}. In
  \bibinfo{booktitle}{{\em ICLR}}.
\newblock


\bibitem[\protect\citeauthoryear{Kolluru and Mukherjee}{Kolluru and
  Mukherjee}{2016}]%
        {kolluru2016query}
\bibfield{author}{\bibinfo{person}{SK Kolluru} {and} \bibinfo{person}{Prasenjit
  Mukherjee}.} \bibinfo{year}{2016}\natexlab{}.
\newblock \showarticletitle{Query Clustering using Segment Specific Context
  Embeddings}. In \bibinfo{booktitle}{{\em arXiv}}.
\newblock


\bibitem[\protect\citeauthoryear{Lebanoff, Song, and Liu}{Lebanoff
  et~al\mbox{.}}{2018}]%
        {lebanoff2018adapting}
\bibfield{author}{\bibinfo{person}{Logan Lebanoff}, \bibinfo{person}{Kaiqiang
  Song}, {and} \bibinfo{person}{Fei Liu}.} \bibinfo{year}{2018}\natexlab{}.
\newblock \showarticletitle{Adapting the neural encoder-decoder framework from
  single to multi-document summarization}. In \bibinfo{booktitle}{{\em EMNLP}}.
\newblock


\bibitem[\protect\citeauthoryear{Lee, Gao, and Zhang}{Lee
  et~al\mbox{.}}{2018}]%
        {lee2018rare}
\bibfield{author}{\bibinfo{person}{Mu-Chu Lee}, \bibinfo{person}{Bin Gao},
  {and} \bibinfo{person}{Ruofei Zhang}.} \bibinfo{year}{2018}\natexlab{}.
\newblock \showarticletitle{Rare query expansion through generative adversarial
  networks in search advertising}. In \bibinfo{booktitle}{{\em SIGKDD}}.
\newblock


\bibitem[\protect\citeauthoryear{Li, Zheng, and Dai}{Li et~al\mbox{.}}{2005}]%
        {li2005kdd}
\bibfield{author}{\bibinfo{person}{Ying Li}, \bibinfo{person}{Zijian Zheng},
  {and} \bibinfo{person}{Honghua~Kathy Dai}.} \bibinfo{year}{2005}\natexlab{}.
\newblock \showarticletitle{KDD CUP-2005 report: Facing a great challenge}.
\newblock \bibinfo{journal}{{\em SIGKDD\/}} (\bibinfo{year}{2005}).
\newblock


\bibitem[\protect\citeauthoryear{Lin}{Lin}{2004}]%
        {lin2004rouge}
\bibfield{author}{\bibinfo{person}{Chin-Yew Lin}.}
  \bibinfo{year}{2004}\natexlab{}.
\newblock \showarticletitle{Rouge: A package for automatic evaluation of
  summaries}. In \bibinfo{booktitle}{{\em Text summarization branches out}}.
\newblock


\bibitem[\protect\citeauthoryear{Maron and Kuhns}{Maron and Kuhns}{1960}]%
        {maron1960relevance}
\bibfield{author}{\bibinfo{person}{Melvin~Earl Maron} {and}
  \bibinfo{person}{John~Larry Kuhns}.} \bibinfo{year}{1960}\natexlab{}.
\newblock \showarticletitle{On relevance, probabilistic indexing and
  information retrieval}.
\newblock \bibinfo{journal}{{\em JACM\/}} (\bibinfo{year}{1960}).
\newblock


\bibitem[\protect\citeauthoryear{Meng, Huang, and Gu}{Meng
  et~al\mbox{.}}{2013}]%
        {meng2013new}
\bibfield{author}{\bibinfo{person}{Lingling Meng}, \bibinfo{person}{Runqing
  Huang}, {and} \bibinfo{person}{Junzhong Gu}.}
  \bibinfo{year}{2013}\natexlab{}.
\newblock \showarticletitle{A new algorithm of web queries clustering using
  user feedback}.
\newblock \bibinfo{journal}{{\em IJSIP\/}} (\bibinfo{year}{2013}).
\newblock


\bibitem[\protect\citeauthoryear{Mihalcea and Tarau}{Mihalcea and
  Tarau}{2004}]%
        {mihalcea2004textrank}
\bibfield{author}{\bibinfo{person}{Rada Mihalcea} {and} \bibinfo{person}{Paul
  Tarau}.} \bibinfo{year}{2004}\natexlab{}.
\newblock \showarticletitle{Textrank: Bringing order into text}. In
  \bibinfo{booktitle}{{\em EMNLP}}.
\newblock


\bibitem[\protect\citeauthoryear{Mohamed and Rajasekaran}{Mohamed and
  Rajasekaran}{2006}]%
        {mohamed2006improving}
\bibfield{author}{\bibinfo{person}{Ahmed~A Mohamed} {and}
  \bibinfo{person}{Sanguthevar Rajasekaran}.} \bibinfo{year}{2006}\natexlab{}.
\newblock \showarticletitle{Improving query-based summarization using document
  graphs}. In \bibinfo{booktitle}{{\em ISSPIT}}.
\newblock


\bibitem[\protect\citeauthoryear{Nakov, Hoogeveen, M{\`a}rquez, Moschitti,
  Mubarak, Baldwin, and Verspoor}{Nakov et~al\mbox{.}}{2017}]%
        {nakov2017semeval}
\bibfield{author}{\bibinfo{person}{Preslav Nakov}, \bibinfo{person}{Doris
  Hoogeveen}, \bibinfo{person}{Llu{\'\i}s M{\`a}rquez},
  \bibinfo{person}{Alessandro Moschitti}, \bibinfo{person}{Hamdy Mubarak},
  \bibinfo{person}{Timothy Baldwin}, {and} \bibinfo{person}{Karin Verspoor}.}
  \bibinfo{year}{2017}\natexlab{}.
\newblock \showarticletitle{SemEval-2017 task 3: Community question answering}.
  In \bibinfo{booktitle}{{\em {S}em{E}val}}.
\newblock


\bibitem[\protect\citeauthoryear{nan Qian, Sakai, Ye, Zheng, and Li}{nan Qian
  et~al\mbox{.}}{2013}]%
        {Qian2013DynamicQI}
\bibfield{author}{\bibinfo{person}{Ya nan Qian}, \bibinfo{person}{Tetsuya
  Sakai}, \bibinfo{person}{Junting Ye}, \bibinfo{person}{Qinghua Zheng}, {and}
  \bibinfo{person}{Cong Li}.} \bibinfo{year}{2013}\natexlab{}.
\newblock \showarticletitle{Dynamic query intent mining from a search log
  stream}. In \bibinfo{booktitle}{{\em CIKM}}.
\newblock


\bibitem[\protect\citeauthoryear{Nema, Khapra, Laha, and Ravindran}{Nema
  et~al\mbox{.}}{2017}]%
        {nema2017diversity}
\bibfield{author}{\bibinfo{person}{Preksha Nema}, \bibinfo{person}{Mitesh
  Khapra}, \bibinfo{person}{Anirban Laha}, {and} \bibinfo{person}{Balaraman
  Ravindran}.} \bibinfo{year}{2017}\natexlab{}.
\newblock \showarticletitle{Diversity driven attention model for query-based
  abstractive summarization}. In \bibinfo{booktitle}{{\em ACL}}.
\newblock


\bibitem[\protect\citeauthoryear{Otsuka, Nishida, Bessho, Asano, and
  Tomita}{Otsuka et~al\mbox{.}}{2018}]%
        {otsuka2018query}
\bibfield{author}{\bibinfo{person}{Atsushi Otsuka}, \bibinfo{person}{Kyosuke
  Nishida}, \bibinfo{person}{Katsuji Bessho}, \bibinfo{person}{Hisako Asano},
  {and} \bibinfo{person}{Junji Tomita}.} \bibinfo{year}{2018}\natexlab{}.
\newblock \showarticletitle{Query expansion with neural question-to-answer
  translation for FAQ-based question answering}. In \bibinfo{booktitle}{{\em
  WebConf}}.
\newblock


\bibitem[\protect\citeauthoryear{Pasca and Harabagiu}{Pasca and
  Harabagiu}{2001}]%
        {pasca2001high}
\bibfield{author}{\bibinfo{person}{Marius~A Pasca} {and}
  \bibinfo{person}{Sandra~M Harabagiu}.} \bibinfo{year}{2001}\natexlab{}.
\newblock \showarticletitle{High performance question/answering}. In
  \bibinfo{booktitle}{{\em SIGIR}}.
\newblock


\bibitem[\protect\citeauthoryear{Peng, Hou, and Song}{Peng
  et~al\mbox{.}}{2009}]%
        {Peng2009Approximating}
\bibfield{author}{\bibinfo{person}{Zhang Peng}, \bibinfo{person}{Yuexian Hou},
  {and} \bibinfo{person}{Dawei Song}.} \bibinfo{year}{2009}\natexlab{}.
\newblock \showarticletitle{Approximating True Relevance Distribution from a
  Mixture Model based on Irrelevance Data}. In \bibinfo{booktitle}{{\em
  SIGIR}}.
\newblock


\bibitem[\protect\citeauthoryear{Radlinski, Szummer, and Craswell}{Radlinski
  et~al\mbox{.}}{2010}]%
        {Radlinski2010InferringQI}
\bibfield{author}{\bibinfo{person}{Filip Radlinski}, \bibinfo{person}{Martin
  Szummer}, {and} \bibinfo{person}{Nick Craswell}.}
  \bibinfo{year}{2010}\natexlab{}.
\newblock \showarticletitle{Inferring query intent from reformulations and
  clicks}. In \bibinfo{booktitle}{{\em WWW}}.
\newblock


\bibitem[\protect\citeauthoryear{Rani and Babu}{Rani and Babu}{2019}]%
        {rani2019efficient}
\bibfield{author}{\bibinfo{person}{Manukonda~Sumathi Rani} {and}
  \bibinfo{person}{Geddati~China Babu}.} \bibinfo{year}{2019}\natexlab{}.
\newblock \showarticletitle{Efficient Query Clustering Technique and Context
  Well-Informed Document Clustering}.
\newblock In \bibinfo{booktitle}{{\em Soft Computing and Signal Processing}}.
\newblock


\bibitem[\protect\citeauthoryear{Rose and Levinson}{Rose and Levinson}{2004}]%
        {rose2004understanding}
\bibfield{author}{\bibinfo{person}{Daniel~E Rose} {and} \bibinfo{person}{Danny
  Levinson}.} \bibinfo{year}{2004}\natexlab{}.
\newblock \showarticletitle{Understanding user goals in web search}. In
  \bibinfo{booktitle}{{\em WWW}}.
\newblock


\bibitem[\protect\citeauthoryear{Rush, Chopra, and Weston}{Rush
  et~al\mbox{.}}{2015}]%
        {rush2015neural}
\bibfield{author}{\bibinfo{person}{Alexander~M Rush}, \bibinfo{person}{Sumit
  Chopra}, {and} \bibinfo{person}{Jason Weston}.}
  \bibinfo{year}{2015}\natexlab{}.
\newblock \showarticletitle{A neural attention model for abstractive sentence
  summarization}. In \bibinfo{booktitle}{{\em ACL}}.
\newblock


\bibitem[\protect\citeauthoryear{Schilder and Kondadadi}{Schilder and
  Kondadadi}{2008}]%
        {schilder2008fastsum}
\bibfield{author}{\bibinfo{person}{Frank Schilder} {and}
  \bibinfo{person}{Ravikumar Kondadadi}.} \bibinfo{year}{2008}\natexlab{}.
\newblock \showarticletitle{FastSum: fast and accurate query-based
  multi-document summarization}. In \bibinfo{booktitle}{{\em ACL}}.
\newblock


\bibitem[\protect\citeauthoryear{Shen, Pan, Sun, Pan, Wu, Yin, and Yang}{Shen
  et~al\mbox{.}}{2005}]%
        {shen2005q}
\bibfield{author}{\bibinfo{person}{Dou Shen}, \bibinfo{person}{Rong Pan},
  \bibinfo{person}{Jian-Tao Sun}, \bibinfo{person}{Jeffrey~Junfeng Pan},
  \bibinfo{person}{Kangheng Wu}, \bibinfo{person}{Jie Yin}, {and}
  \bibinfo{person}{Qiang Yang}.} \bibinfo{year}{2005}\natexlab{}.
\newblock \showarticletitle{Q 2 C@ UST: our winning solution to query
  classification in KDDCUP 2005}.
\newblock \bibinfo{journal}{{\em SIGKDD\/}} (\bibinfo{year}{2005}).
\newblock


\bibitem[\protect\citeauthoryear{Shneiderman, Byrd, and Croft}{Shneiderman
  et~al\mbox{.}}{1997}]%
        {shneiderman1997clarifying}
\bibfield{author}{\bibinfo{person}{Ben Shneiderman}, \bibinfo{person}{Don
  Byrd}, {and} \bibinfo{person}{W~Bruce Croft}.}
  \bibinfo{year}{1997}\natexlab{}.
\newblock \showarticletitle{Clarifying search: A user-interface framework for
  text searches}.
\newblock \bibinfo{journal}{{\em D-lib magazine\/}} (\bibinfo{year}{1997}).
\newblock


\bibitem[\protect\citeauthoryear{Singh and Anand}{Singh and Anand}{2019}]%
        {Singh2018EXS}
\bibfield{author}{\bibinfo{person}{Jaspreet Singh} {and}
  \bibinfo{person}{Avishek Anand}.} \bibinfo{year}{2019}\natexlab{}.
\newblock \showarticletitle{EXS: Explainable Search Using Local Model Agnostic
  Interpretability}. In \bibinfo{booktitle}{{\em WSDM}}.
\newblock


\bibitem[\protect\citeauthoryear{Singh and Anand}{Singh and Anand}{2020}]%
        {MAI2020}
\bibfield{author}{\bibinfo{person}{Jaspreet Singh} {and}
  \bibinfo{person}{Avishek Anand}.} \bibinfo{year}{2020}\natexlab{}.
\newblock \showarticletitle{Model Agnostic Interpretability of Rankers via
  Intent Modelling}. In \bibinfo{booktitle}{{\em Proceedings of the 2020
  Conference on Fairness, Accountability, and Transparency}}.
\newblock


\bibitem[\protect\citeauthoryear{Singh and Sharan}{Singh and Sharan}{2015a}]%
        {singh2015co}
\bibfield{author}{\bibinfo{person}{Jagendra Singh} {and} \bibinfo{person}{Aditi
  Sharan}.} \bibinfo{year}{2015}\natexlab{a}.
\newblock \showarticletitle{Co-occurrence and semantic similarity based hybrid
  approach for improving automatic query expansion in information retrieval}.
  In \bibinfo{booktitle}{{\em ICDCIT}}.
\newblock


\bibitem[\protect\citeauthoryear{Singh and Sharan}{Singh and Sharan}{2015b}]%
        {singh2015context}
\bibfield{author}{\bibinfo{person}{Jagendra Singh} {and} \bibinfo{person}{Aditi
  Sharan}.} \bibinfo{year}{2015}\natexlab{b}.
\newblock \showarticletitle{Context window based co-occurrence approach for
  improving feedback based query expansion in information retrieval}. In
  \bibinfo{booktitle}{{\em IJIRR}}.
\newblock


\bibitem[\protect\citeauthoryear{Song, Zhao, and Liu}{Song
  et~al\mbox{.}}{2018}]%
        {song2018structure}
\bibfield{author}{\bibinfo{person}{Kaiqiang Song}, \bibinfo{person}{Lin Zhao},
  {and} \bibinfo{person}{Fei Liu}.} \bibinfo{year}{2018}\natexlab{}.
\newblock \bibinfo{title}{Structure-infused copy mechanisms for abstractive
  summarization}.
\newblock   (\bibinfo{year}{2018}).
\newblock


\bibitem[\protect\citeauthoryear{Steinberger and Je{\v{z}}ek}{Steinberger and
  Je{\v{z}}ek}{2004}]%
        {steinberger2004text}
\bibfield{author}{\bibinfo{person}{Josef Steinberger} {and}
  \bibinfo{person}{Karel Je{\v{z}}ek}.} \bibinfo{year}{2004}\natexlab{}.
\newblock \showarticletitle{Text summarization and singular value
  decomposition}. In \bibinfo{booktitle}{{\em International Conference on
  Advances in Information Systems}}. Springer, \bibinfo{pages}{245--254}.
\newblock


\bibitem[\protect\citeauthoryear{Sutskever, Vinyals, and Le}{Sutskever
  et~al\mbox{.}}{2014}]%
        {sutskever2014sequence}
\bibfield{author}{\bibinfo{person}{I Sutskever}, \bibinfo{person}{O Vinyals},
  {and} \bibinfo{person}{QV Le}.} \bibinfo{year}{2014}\natexlab{}.
\newblock \showarticletitle{Sequence to sequence learning with neural
  networks}. In \bibinfo{booktitle}{{\em NIPS}}.
\newblock


\bibitem[\protect\citeauthoryear{Tao and Zhai}{Tao and Zhai}{2006}]%
        {Tao2006Regularized}
\bibfield{author}{\bibinfo{person}{Tao Tao} {and} \bibinfo{person}{Cheng~Xiang
  Zhai}.} \bibinfo{year}{2006}\natexlab{}.
\newblock \showarticletitle{Regularized estimation of mixture models for robust
  pseudo-relevance feedback}. In \bibinfo{booktitle}{{\em SIGIR}}.
\newblock


\bibitem[\protect\citeauthoryear{Verma and Ganguly}{Verma and Ganguly}{2019}]%
        {Manisha2019LIRME}
\bibfield{author}{\bibinfo{person}{Manisha Verma} {and}
  \bibinfo{person}{Debasis Ganguly}.} \bibinfo{year}{2019}\natexlab{}.
\newblock \showarticletitle{LIRME: Locally Interpretable Ranking Model
  Explanation}. In \bibinfo{booktitle}{{\em SIGIR}}.
\newblock


\bibitem[\protect\citeauthoryear{Voorhees}{Voorhees}{1994}]%
        {voorhees1994query}
\bibfield{author}{\bibinfo{person}{Ellen~M Voorhees}.}
  \bibinfo{year}{1994}\natexlab{}.
\newblock \showarticletitle{Query expansion using lexical-semantic relations}.
  In \bibinfo{booktitle}{{\em SIGIR}}.
\newblock


\bibitem[\protect\citeauthoryear{Wen, Nie, and Zhang}{Wen
  et~al\mbox{.}}{2002}]%
        {wen2002query}
\bibfield{author}{\bibinfo{person}{Ji-Rong Wen}, \bibinfo{person}{Jian-Yun
  Nie}, {and} \bibinfo{person}{Hong-Jiang Zhang}.}
  \bibinfo{year}{2002}\natexlab{}.
\newblock \showarticletitle{Query clustering using user logs}.
\newblock \bibinfo{journal}{{\em TOIS\/}} (\bibinfo{year}{2002}).
\newblock


\bibitem[\protect\citeauthoryear{Xu, Zheng, Zhang, and Tao}{Xu
  et~al\mbox{.}}{2016}]%
        {Xu2016Learning}
\bibfield{author}{\bibinfo{person}{Chen Xu}, \bibinfo{person}{Qin Zheng},
  \bibinfo{person}{Yongfeng Zhang}, {and} \bibinfo{person}{Xu Tao}.}
  \bibinfo{year}{2016}\natexlab{}.
\newblock \showarticletitle{Learning to Rank Features for Recommendation over
  Multiple Categories}.
\newblock  (\bibinfo{year}{2016}).
\newblock


\bibitem[\protect\citeauthoryear{Zhang, Tan, and Wan}{Zhang
  et~al\mbox{.}}{2018}]%
        {zhang2018adapting}
\bibfield{author}{\bibinfo{person}{Jianmin Zhang}, \bibinfo{person}{Jiwei Tan},
  {and} \bibinfo{person}{Xiaojun Wan}.} \bibinfo{year}{2018}\natexlab{}.
\newblock \showarticletitle{Adapting neural single-document summarization model
  for abstractive multi-document summarization: A pilot study}. In
  \bibinfo{booktitle}{{\em INLG}}.
\newblock


\bibitem[\protect\citeauthoryear{Zhang and Nasraoui}{Zhang and
  Nasraoui}{2006}]%
        {zhang2006mining}
\bibfield{author}{\bibinfo{person}{Zhiyong Zhang} {and} \bibinfo{person}{Olfa
  Nasraoui}.} \bibinfo{year}{2006}\natexlab{}.
\newblock \showarticletitle{Mining search engine query logs for query
  recommendation}. In \bibinfo{booktitle}{{\em WWW}}.
\newblock


\bibitem[\protect\citeauthoryear{Zhang, Wang, Si, and Gao}{Zhang
  et~al\mbox{.}}{2016}]%
        {zhang2016learning}
\bibfield{author}{\bibinfo{person}{Zhiwei Zhang}, \bibinfo{person}{Qifan Wang},
  \bibinfo{person}{Luo Si}, {and} \bibinfo{person}{Jianfeng Gao}.}
  \bibinfo{year}{2016}\natexlab{}.
\newblock \showarticletitle{Learning for efficient supervised query expansion
  via two-stage feature selection}. In \bibinfo{booktitle}{{\em SIGIR}}.
\newblock


\end{thebibliography}



\begin{thebibliography}{54}


\ifx \showCODEN    \undefined \def \showCODEN     #1{\unskip}     \fi
\ifx \showDOI      \undefined \def \showDOI       #1{#1}\fi
\ifx \showISBNx    \undefined \def \showISBNx     #1{\unskip}     \fi
\ifx \showISBNxiii \undefined \def \showISBNxiii  #1{\unskip}     \fi
\ifx \showISSN     \undefined \def \showISSN      #1{\unskip}     \fi
\ifx \showLCCN     \undefined \def \showLCCN      #1{\unskip}     \fi
\ifx \shownote     \undefined \def \shownote      #1{#1}          \fi
\ifx \showarticletitle \undefined \def \showarticletitle #1{#1}   \fi
\ifx \showURL      \undefined \def \showURL       {\relax}        \fi
\providecommand\bibfield[2]{#2}
\providecommand\bibinfo[2]{#2}
\providecommand\natexlab[1]{#1}
\providecommand\showeprint[2][]{arXiv:#2}

\bibitem[\protect\citeauthoryear{Alfonseca, Pighin, and Garrido}{Alfonseca
  et~al\mbox{.}}{2013}]%
        {alfonseca2013heady}
\bibfield{author}{\bibinfo{person}{Enrique Alfonseca}, \bibinfo{person}{Daniele
  Pighin}, {and} \bibinfo{person}{Guillermo Garrido}.}
  \bibinfo{year}{2013}\natexlab{}.
\newblock \showarticletitle{HEADY: News headline abstraction through event
  pattern clustering.}. In \bibinfo{booktitle}{{\em ACL}}.
  \bibinfo{pages}{1243--1253}.
\newblock


\bibitem[\protect\citeauthoryear{Alley, Schreiber, Ramsdell, and Muffo}{Alley
  et~al\mbox{.}}{2006}]%
        {alley2006design}
\bibfield{author}{\bibinfo{person}{Michael Alley}, \bibinfo{person}{Madeline
  Schreiber}, \bibinfo{person}{Katrina Ramsdell}, {and} \bibinfo{person}{John
  Muffo}.} \bibinfo{year}{2006}\natexlab{}.
\newblock \showarticletitle{How the design of headlines in presentation slides
  affects audience retention}.
\newblock \bibinfo{journal}{{\em Technical communication\/}}
  \bibinfo{volume}{53}, \bibinfo{number}{2} (\bibinfo{year}{2006}),
  \bibinfo{pages}{225--234}.
\newblock


\bibitem[\protect\citeauthoryear{Ayana, Liu, and Sun}{Ayana
  et~al\mbox{.}}{2016}]%
        {ayana2016neural}
\bibfield{author}{\bibinfo{person}{Shiqi~Shen Ayana}, \bibinfo{person}{Zhiyuan
  Liu}, {and} \bibinfo{person}{Maosong Sun}.} \bibinfo{year}{2016}\natexlab{}.
\newblock \showarticletitle{Neural headline generation with minimum risk
  training}.
\newblock \bibinfo{journal}{{\em arXiv preprint arXiv:1604.01904\/}}
  (\bibinfo{year}{2016}).
\newblock


\bibitem[\protect\citeauthoryear{Bahdanau, Cho, and Bengio}{Bahdanau
  et~al\mbox{.}}{2015}]%
        {bahdanau2014neural}
\bibfield{author}{\bibinfo{person}{Dzmitry Bahdanau},
  \bibinfo{person}{Kyunghyun Cho}, {and} \bibinfo{person}{Yoshua Bengio}.}
  \bibinfo{year}{2015}\natexlab{}.
\newblock \showarticletitle{Neural machine translation by jointly learning to
  align and translate}. In \bibinfo{booktitle}{{\em ICLR}}.
\newblock


\bibitem[\protect\citeauthoryear{Banko, Mittal, and Witbrock}{Banko
  et~al\mbox{.}}{2000}]%
        {banko2000headline}
\bibfield{author}{\bibinfo{person}{Michele Banko}, \bibinfo{person}{Vibhu~O
  Mittal}, {and} \bibinfo{person}{Michael~J Witbrock}.}
  \bibinfo{year}{2000}\natexlab{}.
\newblock \showarticletitle{Headline generation based on statistical
  translation}. In \bibinfo{booktitle}{{\em Proceedings of the 38th Annual
  Meeting on Association for Computational Linguistics}}. Association for
  Computational Linguistics, \bibinfo{pages}{318--325}.
\newblock


\bibitem[\protect\citeauthoryear{Carbonell and Goldstein}{Carbonell and
  Goldstein}{1998}]%
        {carbonell1998use}
\bibfield{author}{\bibinfo{person}{Jaime Carbonell} {and} \bibinfo{person}{Jade
  Goldstein}.} \bibinfo{year}{1998}\natexlab{}.
\newblock \showarticletitle{The use of MMR, diversity-based reranking for
  reordering documents and producing summaries}. In \bibinfo{booktitle}{{\em
  SIGIR}}. ACM, \bibinfo{pages}{335--336}.
\newblock


\bibitem[\protect\citeauthoryear{Chakraborty, Paranjape, Kakarla, and
  Ganguly}{Chakraborty et~al\mbox{.}}{2016}]%
        {chakraborty2016stop}
\bibfield{author}{\bibinfo{person}{Abhijnan Chakraborty},
  \bibinfo{person}{Bhargavi Paranjape}, \bibinfo{person}{Sourya Kakarla}, {and}
  \bibinfo{person}{Niloy Ganguly}.} \bibinfo{year}{2016}\natexlab{}.
\newblock \showarticletitle{Stop clickbait: Detecting and preventing clickbaits
  in online news media}. In \bibinfo{booktitle}{{\em Advances in Social
  Networks Analysis and Mining (ASONAM), 2016 IEEE/ACM International Conference
  on}}. IEEE, \bibinfo{pages}{9--16}.
\newblock


\bibitem[\protect\citeauthoryear{Chali and Golestanirad}{Chali and
  Golestanirad}{2016}]%
        {chali2016ranking}
\bibfield{author}{\bibinfo{person}{Yllias Chali} {and} \bibinfo{person}{Sina
  Golestanirad}.} \bibinfo{year}{2016}\natexlab{}.
\newblock \showarticletitle{Ranking Automatically Generated Questions Using
  Common Human Queries.}. In \bibinfo{booktitle}{{\em INLG}}.
  \bibinfo{pages}{217--221}.
\newblock


\bibitem[\protect\citeauthoryear{Chen, Zhu, Ling, Wei, and Jiang}{Chen
  et~al\mbox{.}}{2016}]%
        {chen2016distraction}
\bibfield{author}{\bibinfo{person}{Qian Chen}, \bibinfo{person}{Xiaodan Zhu},
  \bibinfo{person}{Zhenhua Ling}, \bibinfo{person}{Si Wei}, {and}
  \bibinfo{person}{Hui Jiang}.} \bibinfo{year}{2016}\natexlab{}.
\newblock \showarticletitle{Distraction-Based Neural Networks for Document
  Summarization}. In \bibinfo{booktitle}{{\em IJCAI}}.
\newblock


\bibitem[\protect\citeauthoryear{Cho, Van~Merri{\"e}nboer, Gulcehre, Bahdanau,
  Bougares, Schwenk, and Bengio}{Cho et~al\mbox{.}}{2014}]%
        {cho2014learning}
\bibfield{author}{\bibinfo{person}{Kyunghyun Cho}, \bibinfo{person}{Bart
  Van~Merri{\"e}nboer}, \bibinfo{person}{Caglar Gulcehre},
  \bibinfo{person}{Dzmitry Bahdanau}, \bibinfo{person}{Fethi Bougares},
  \bibinfo{person}{Holger Schwenk}, {and} \bibinfo{person}{Yoshua Bengio}.}
  \bibinfo{year}{2014}\natexlab{}.
\newblock \showarticletitle{Learning phrase representations using RNN
  encoder-decoder for statistical machine translation}. In
  \bibinfo{booktitle}{{\em EMNLP}}.
\newblock


\bibitem[\protect\citeauthoryear{Chopra, Auli, Rush, and Harvard}{Chopra
  et~al\mbox{.}}{2016}]%
        {chopra2016abstractive}
\bibfield{author}{\bibinfo{person}{Sumit Chopra}, \bibinfo{person}{Michael
  Auli}, \bibinfo{person}{Alexander~M Rush}, {and} \bibinfo{person}{SEAS
  Harvard}.} \bibinfo{year}{2016}\natexlab{}.
\newblock \showarticletitle{Abstractive Sentence Summarization with Attentive
  Recurrent Neural Networks}. In \bibinfo{booktitle}{{\em HLT-NAACL}}.
  \bibinfo{pages}{93--98}.
\newblock


\bibitem[\protect\citeauthoryear{Colmenares, Litvak, Mantrach, and
  Silvestri}{Colmenares et~al\mbox{.}}{2015}]%
        {colmenares2015heads}
\bibfield{author}{\bibinfo{person}{Carlos~A Colmenares},
  \bibinfo{person}{Marina Litvak}, \bibinfo{person}{Amin Mantrach}, {and}
  \bibinfo{person}{Fabrizio Silvestri}.} \bibinfo{year}{2015}\natexlab{}.
\newblock \showarticletitle{HEADS: Headline Generation as Sequence Prediction
  Using an Abstract Feature-Rich Space.}. In \bibinfo{booktitle}{{\em
  HLT-NAACL}}. \bibinfo{pages}{133--142}.
\newblock


\bibitem[\protect\citeauthoryear{Denkowski and Lavie}{Denkowski and
  Lavie}{2014}]%
        {denkowski2014meteor}
\bibfield{author}{\bibinfo{person}{Michael Denkowski} {and}
  \bibinfo{person}{Alon Lavie}.} \bibinfo{year}{2014}\natexlab{}.
\newblock \showarticletitle{Meteor universal: Language specific translation
  evaluation for any target language}. In \bibinfo{booktitle}{{\em Proceedings
  of the ninth workshop on statistical machine translation}}.
  \bibinfo{pages}{376--380}.
\newblock


\bibitem[\protect\citeauthoryear{Dorr, Zajic, and Schwartz}{Dorr
  et~al\mbox{.}}{2003}]%
        {dorr2003hedge}
\bibfield{author}{\bibinfo{person}{Bonnie Dorr}, \bibinfo{person}{David Zajic},
  {and} \bibinfo{person}{Richard Schwartz}.} \bibinfo{year}{2003}\natexlab{}.
\newblock \showarticletitle{Hedge trimmer: A parse-and-trim approach to
  headline generation}. In \bibinfo{booktitle}{{\em HLT-NAACL}}. Association
  for Computational Linguistics, \bibinfo{pages}{1--8}.
\newblock


\bibitem[\protect\citeauthoryear{Du, Shao, and Cardie}{Du
  et~al\mbox{.}}{2017}]%
        {du2017learning}
\bibfield{author}{\bibinfo{person}{Xinya Du}, \bibinfo{person}{Junru Shao},
  {and} \bibinfo{person}{Claire Cardie}.} \bibinfo{year}{2017}\natexlab{}.
\newblock \showarticletitle{Learning to Ask: Neural Question Generation for
  Reading Comprehension}. In \bibinfo{booktitle}{{\em ACL}}.
\newblock


\bibitem[\protect\citeauthoryear{Edmundson}{Edmundson}{1964}]%
        {edmundson1964problems}
\bibfield{author}{\bibinfo{person}{HP Edmundson}.}
  \bibinfo{year}{1964}\natexlab{}.
\newblock \showarticletitle{Problems in automatic abstracting}.
\newblock \bibinfo{journal}{{\it Commun. ACM}} \bibinfo{volume}{7},
  \bibinfo{number}{4} (\bibinfo{year}{1964}), \bibinfo{pages}{259--263}.
\newblock


\bibitem[\protect\citeauthoryear{Erkan and Radev}{Erkan and Radev}{2004}]%
        {erkan2004lexrank}
\bibfield{author}{\bibinfo{person}{G{\"u}nes Erkan} {and}
  \bibinfo{person}{Dragomir~R Radev}.} \bibinfo{year}{2004}\natexlab{}.
\newblock \showarticletitle{Lexrank: Graph-based lexical centrality as salience
  in text summarization}.
\newblock \bibinfo{journal}{{\em JAIR\/}}  \bibinfo{volume}{22}
  (\bibinfo{year}{2004}), \bibinfo{pages}{457--479}.
\newblock


\bibitem[\protect\citeauthoryear{Heilman and Smith}{Heilman and Smith}{2010}]%
        {heilman2010good}
\bibfield{author}{\bibinfo{person}{Michael Heilman} {and}
  \bibinfo{person}{Noah~A Smith}.} \bibinfo{year}{2010}\natexlab{}.
\newblock \showarticletitle{Good question! statistical ranking for question
  generation}. In \bibinfo{booktitle}{{\em HLT-NAACL}}.
  \bibinfo{pages}{609--617}.
\newblock


\bibitem[\protect\citeauthoryear{Hochreiter and Schmidhuber}{Hochreiter and
  Schmidhuber}{1997}]%
        {hochreiter1997long}
\bibfield{author}{\bibinfo{person}{Sepp Hochreiter} {and}
  \bibinfo{person}{J{\"u}rgen Schmidhuber}.} \bibinfo{year}{1997}\natexlab{}.
\newblock \showarticletitle{Long short-term memory}.
\newblock \bibinfo{journal}{{\em Neural computation\/}} \bibinfo{volume}{9},
  \bibinfo{number}{8} (\bibinfo{year}{1997}), \bibinfo{pages}{1735--1780}.
\newblock


\bibitem[\protect\citeauthoryear{Hu, Chen, and Zhu}{Hu et~al\mbox{.}}{2015}]%
        {hu2015lcsts}
\bibfield{author}{\bibinfo{person}{Baotian Hu}, \bibinfo{person}{Qingcai Chen},
  {and} \bibinfo{person}{Fangze Zhu}.} \bibinfo{year}{2015}\natexlab{}.
\newblock \showarticletitle{Lcsts: A large scale chinese short text
  summarization dataset}. In \bibinfo{booktitle}{{\em EMNLP}}.
\newblock


\bibitem[\protect\citeauthoryear{Jing and McKeown}{Jing and McKeown}{1999}]%
        {jing1999decomposition}
\bibfield{author}{\bibinfo{person}{Hongyan Jing} {and}
  \bibinfo{person}{Kathleen~R McKeown}.} \bibinfo{year}{1999}\natexlab{}.
\newblock \showarticletitle{The decomposition of human-written summary
  sentences}. In \bibinfo{booktitle}{{\em Proceedings of the 22nd annual
  international ACM SIGIR conference on Research and development in information
  retrieval}}. ACM, \bibinfo{pages}{129--136}.
\newblock


\bibitem[\protect\citeauthoryear{Kingma and Ba}{Kingma and Ba}{2015}]%
        {kingma2014adam}
\bibfield{author}{\bibinfo{person}{Diederik Kingma} {and}
  \bibinfo{person}{Jimmy Ba}.} \bibinfo{year}{2015}\natexlab{}.
\newblock \showarticletitle{Adam: A method for stochastic optimization}. In
  \bibinfo{booktitle}{{\em ICLR}}.
\newblock


\bibitem[\protect\citeauthoryear{Kumar, Banchs, and D'Haro~Enriquez}{Kumar
  et~al\mbox{.}}{2015}]%
        {kumar2015revup}
\bibfield{author}{\bibinfo{person}{Girish Kumar}, \bibinfo{person}{Rafael~E
  Banchs}, {and} \bibinfo{person}{Luis~Fernando D'Haro~Enriquez}.}
  \bibinfo{year}{2015}\natexlab{}.
\newblock \showarticletitle{Revup: Automatic gap-fill question generation from
  educational texts}.
\newblock


\bibitem[\protect\citeauthoryear{Lai and Farbrot}{Lai and Farbrot}{2014}]%
        {lai2014makes}
\bibfield{author}{\bibinfo{person}{Linda Lai} {and} \bibinfo{person}{Audun
  Farbrot}.} \bibinfo{year}{2014}\natexlab{}.
\newblock \showarticletitle{What makes you click? The effect of question
  headlines on readership in computer-mediated communication}.
\newblock \bibinfo{journal}{{\em Social Influence\/}} \bibinfo{volume}{9},
  \bibinfo{number}{4} (\bibinfo{year}{2014}), \bibinfo{pages}{289--299}.
\newblock


\bibitem[\protect\citeauthoryear{LaRocque}{LaRocque}{2003}]%
        {larocque2003heads}
\bibfield{author}{\bibinfo{person}{Paul LaRocque}.}
  \bibinfo{year}{2003}\natexlab{}.
\newblock \bibinfo{booktitle}{{\em Heads You Win: An Easy Guide to Better
  Headline and Caption Writing}}.
\newblock \bibinfo{publisher}{Marion Street Press, Inc.}
\newblock


\bibitem[\protect\citeauthoryear{Li, Luong, and Jurafsky}{Li
  et~al\mbox{.}}{2015}]%
        {li2015hierarchical}
\bibfield{author}{\bibinfo{person}{Jiwei Li}, \bibinfo{person}{Minh-Thang
  Luong}, {and} \bibinfo{person}{Dan Jurafsky}.}
  \bibinfo{year}{2015}\natexlab{}.
\newblock \showarticletitle{A hierarchical neural autoencoder for paragraphs
  and documents}. In \bibinfo{booktitle}{{\em ACL}}.
\newblock


\bibitem[\protect\citeauthoryear{Lin}{Lin}{2004}]%
        {lin2004rouge}
\bibfield{author}{\bibinfo{person}{Chin-Yew Lin}.}
  \bibinfo{year}{2004}\natexlab{}.
\newblock \showarticletitle{Rouge: A package for automatic evaluation of
  summaries}. In \bibinfo{booktitle}{{\em Text summarization branches out:
  Proceedings of the ACL-04 workshop}}, Vol.~\bibinfo{volume}{8}. Barcelona,
  Spain.
\newblock


\bibitem[\protect\citeauthoryear{Lindberg, Popowich, Nesbit, and
  Winne}{Lindberg et~al\mbox{.}}{2013}]%
        {lindberg2013generating}
\bibfield{author}{\bibinfo{person}{David Lindberg}, \bibinfo{person}{Fred
  Popowich}, \bibinfo{person}{John Nesbit}, {and} \bibinfo{person}{Phil
  Winne}.} \bibinfo{year}{2013}\natexlab{}.
\newblock \showarticletitle{Generating natural language questions to support
  learning on-line}.
\newblock  (\bibinfo{year}{2013}).
\newblock


\bibitem[\protect\citeauthoryear{Luhn}{Luhn}{1958}]%
        {luhn1958automatic}
\bibfield{author}{\bibinfo{person}{Hans~Peter Luhn}.}
  \bibinfo{year}{1958}\natexlab{}.
\newblock \showarticletitle{The automatic creation of literature abstracts}.
\newblock \bibinfo{journal}{{\em IBM Journal of research and development\/}}
  \bibinfo{volume}{2}, \bibinfo{number}{2} (\bibinfo{year}{1958}),
  \bibinfo{pages}{159--165}.
\newblock


\bibitem[\protect\citeauthoryear{Mannem, Prasad, and Joshi}{Mannem
  et~al\mbox{.}}{2010}]%
        {mannem2010question}
\bibfield{author}{\bibinfo{person}{Prashanth Mannem}, \bibinfo{person}{Rashmi
  Prasad}, {and} \bibinfo{person}{Aravind Joshi}.}
  \bibinfo{year}{2010}\natexlab{}.
\newblock \showarticletitle{Question generation from paragraphs at UPenn:
  QGSTEC system description}. In \bibinfo{booktitle}{{\em Proceedings of
  QG2010}}. \bibinfo{pages}{84--91}.
\newblock


\bibitem[\protect\citeauthoryear{Mathis, Rush, and Young}{Mathis
  et~al\mbox{.}}{1973}]%
        {mathis1973improvement}
\bibfield{author}{\bibinfo{person}{Betty~A Mathis}, \bibinfo{person}{James~E
  Rush}, {and} \bibinfo{person}{Carol~E Young}.}
  \bibinfo{year}{1973}\natexlab{}.
\newblock \showarticletitle{Improvement of automatic abstracts by the use of
  structural analysis}.
\newblock \bibinfo{journal}{{\em Journal of the Association for Information
  Science and Technology\/}} \bibinfo{volume}{24}, \bibinfo{number}{2}
  (\bibinfo{year}{1973}), \bibinfo{pages}{101--109}.
\newblock


\bibitem[\protect\citeauthoryear{Merino and Lema}{Merino and Lema}{2008}]%
        {merino2008needs}
\bibfield{author}{\bibinfo{person}{Gerardo~Atienza Merino} {and}
  \bibinfo{person}{Leonor~Varela Lema}.} \bibinfo{year}{2008}\natexlab{}.
\newblock \showarticletitle{Needs and demands of policy-makers}.
\newblock \bibinfo{journal}{{\em HEALTH TECHNOLOGY ASSESSMENT AND HEALTH
  POLICY-MAKING IN EUROPE\/}} (\bibinfo{year}{2008}), \bibinfo{pages}{137}.
\newblock


\bibitem[\protect\citeauthoryear{Mihalcea and Tarau}{Mihalcea and
  Tarau}{2004}]%
        {mihalcea2004textrank}
\bibfield{author}{\bibinfo{person}{Rada Mihalcea} {and} \bibinfo{person}{Paul
  Tarau}.} \bibinfo{year}{2004}\natexlab{}.
\newblock \showarticletitle{TextRank: Bringing Order into Text.}. In
  \bibinfo{booktitle}{{\em EMNLP}}, Vol.~\bibinfo{volume}{4}.
  \bibinfo{pages}{404--411}.
\newblock


\bibitem[\protect\citeauthoryear{Mostafazadeh, Misra, Devlin, Mitchell, He, and
  Vanderwende}{Mostafazadeh et~al\mbox{.}}{2016}]%
        {mostafazadeh2016generating}
\bibfield{author}{\bibinfo{person}{Nasrin Mostafazadeh}, \bibinfo{person}{Ishan
  Misra}, \bibinfo{person}{Jacob Devlin}, \bibinfo{person}{Margaret Mitchell},
  \bibinfo{person}{Xiaodong He}, {and} \bibinfo{person}{Lucy Vanderwende}.}
  \bibinfo{year}{2016}\natexlab{}.
\newblock \showarticletitle{Generating natural questions about an image}. In
  \bibinfo{booktitle}{{\em ACL}}.
\newblock


\bibitem[\protect\citeauthoryear{Nallapati, Zhai, and Zhou}{Nallapati
  et~al\mbox{.}}{2017}]%
        {nallapati2017summarunner}
\bibfield{author}{\bibinfo{person}{Ramesh Nallapati}, \bibinfo{person}{Feifei
  Zhai}, {and} \bibinfo{person}{Bowen Zhou}.} \bibinfo{year}{2017}\natexlab{}.
\newblock \showarticletitle{SummaRuNNer: A recurrent neural network based
  sequence model for extractive summarization of documents}. In
  \bibinfo{booktitle}{{\em AAAI}}.
\newblock


\bibitem[\protect\citeauthoryear{Nallapati, Zhou, Gulcehre, Xiang,
  et~al\mbox{.}}{Nallapati et~al\mbox{.}}{2016}]%
        {nallapati2016abstractive}
\bibfield{author}{\bibinfo{person}{Ramesh Nallapati}, \bibinfo{person}{Bowen
  Zhou}, \bibinfo{person}{Caglar Gulcehre}, \bibinfo{person}{Bing Xiang},
  {et~al\mbox{.}}} \bibinfo{year}{2016}\natexlab{}.
\newblock \showarticletitle{Abstractive text summarization using
  sequence-to-sequence rnns and beyond}.
\newblock \bibinfo{journal}{{\em arXiv preprint arXiv:1602.06023\/}}
  (\bibinfo{year}{2016}).
\newblock


\bibitem[\protect\citeauthoryear{Nasunee}{Nasunee}{2004}]%
        {nasunee2004analysis}
\bibfield{author}{\bibinfo{person}{Juntiga Nasunee}.}
  \bibinfo{year}{2004}\natexlab{}.
\newblock \showarticletitle{An analysis of catchy words and sentences in
  Volkswagen beetle advertisements in the United States}.
\newblock \bibinfo{journal}{{\em Unpublished master’s project.
  Srinakharinwirot University\/}} (\bibinfo{year}{2004}).
\newblock


\bibitem[\protect\citeauthoryear{Page, Brin, Motwani, and Winograd}{Page
  et~al\mbox{.}}{1999}]%
        {page1999pagerank}
\bibfield{author}{\bibinfo{person}{Lawrence Page}, \bibinfo{person}{Sergey
  Brin}, \bibinfo{person}{Rajeev Motwani}, {and} \bibinfo{person}{Terry
  Winograd}.} \bibinfo{year}{1999}\natexlab{}.
\newblock \bibinfo{booktitle}{{\em The PageRank citation ranking: Bringing
  order to the web.}}
\newblock \bibinfo{type}{{T}echnical {R}eport}. \bibinfo{institution}{Stanford
  InfoLab}.
\newblock


\bibitem[\protect\citeauthoryear{Papineni, Roukos, Ward, and Zhu}{Papineni
  et~al\mbox{.}}{2002}]%
        {papineni2002bleu}
\bibfield{author}{\bibinfo{person}{Kishore Papineni}, \bibinfo{person}{Salim
  Roukos}, \bibinfo{person}{Todd Ward}, {and} \bibinfo{person}{Wei-Jing Zhu}.}
  \bibinfo{year}{2002}\natexlab{}.
\newblock \showarticletitle{BLEU: a method for automatic evaluation of machine
  translation}. In \bibinfo{booktitle}{{\em Proceedings of the 40th annual
  meeting on association for computational linguistics}}. Association for
  Computational Linguistics, \bibinfo{pages}{311--318}.
\newblock


\bibitem[\protect\citeauthoryear{Pighin, Cornolti, Alfonseca, and
  Filippova}{Pighin et~al\mbox{.}}{2014}]%
        {pighin2014modelling}
\bibfield{author}{\bibinfo{person}{Daniele Pighin}, \bibinfo{person}{Marco
  Cornolti}, \bibinfo{person}{Enrique Alfonseca}, {and} \bibinfo{person}{Katja
  Filippova}.} \bibinfo{year}{2014}\natexlab{}.
\newblock \showarticletitle{Modelling Events through Memory-based, Open-IE
  Patterns for Abstractive Summarization.}. In \bibinfo{booktitle}{{\em ACL}}.
  \bibinfo{pages}{892--901}.
\newblock


\bibitem[\protect\citeauthoryear{Popowich and Winne}{Popowich and
  Winne}{2013}]%
        {popowich2013generating}
\bibfield{author}{\bibinfo{person}{David Lindberg~Fred Popowich} {and}
  \bibinfo{person}{John Nesbit~Phil Winne}.} \bibinfo{year}{2013}\natexlab{}.
\newblock \showarticletitle{Generating Natural Language Questions to Support
  Learning On-Line}.
\newblock \bibinfo{journal}{{\em ENLG\/}} (\bibinfo{year}{2013}),
  \bibinfo{pages}{105}.
\newblock


\bibitem[\protect\citeauthoryear{Radev and McKeown}{Radev and McKeown}{1998}]%
        {radev1998generating}
\bibfield{author}{\bibinfo{person}{Dragomir~R Radev} {and}
  \bibinfo{person}{Kathleen~R McKeown}.} \bibinfo{year}{1998}\natexlab{}.
\newblock \showarticletitle{Generating natural language summaries from multiple
  on-line sources}.
\newblock \bibinfo{journal}{{\em Computational Linguistics\/}}
  \bibinfo{volume}{24}, \bibinfo{number}{3} (\bibinfo{year}{1998}),
  \bibinfo{pages}{470--500}.
\newblock


\bibitem[\protect\citeauthoryear{Robertson and Walker}{Robertson and
  Walker}{1994}]%
        {robertson1994some}
\bibfield{author}{\bibinfo{person}{Stephen~E Robertson} {and}
  \bibinfo{person}{Steve Walker}.} \bibinfo{year}{1994}\natexlab{}.
\newblock \showarticletitle{Some simple effective approximations to the
  2-poisson model for probabilistic weighted retrieval}. In
  \bibinfo{booktitle}{{\em SIGIR}}. Springer-Verlag New York, Inc.,
  \bibinfo{pages}{232--241}.
\newblock


\bibitem[\protect\citeauthoryear{Rumelhart, Hinton, Williams,
  et~al\mbox{.}}{Rumelhart et~al\mbox{.}}{1988}]%
        {rumelhart1988learning}
\bibfield{author}{\bibinfo{person}{David~E Rumelhart},
  \bibinfo{person}{Geoffrey~E Hinton}, \bibinfo{person}{Ronald~J Williams},
  {et~al\mbox{.}}} \bibinfo{year}{1988}\natexlab{}.
\newblock \showarticletitle{Learning representations by back-propagating
  errors}.
\newblock \bibinfo{journal}{{\em Cognitive modeling\/}} \bibinfo{volume}{5},
  \bibinfo{number}{3} (\bibinfo{year}{1988}), \bibinfo{pages}{1}.
\newblock


\bibitem[\protect\citeauthoryear{Rus, Wyse, Piwek, Lintean, Stoyanchev, and
  Moldovan}{Rus et~al\mbox{.}}{2010}]%
        {rus2010first}
\bibfield{author}{\bibinfo{person}{Vasile Rus}, \bibinfo{person}{Brendan Wyse},
  \bibinfo{person}{Paul Piwek}, \bibinfo{person}{Mihai Lintean},
  \bibinfo{person}{Svetlana Stoyanchev}, {and} \bibinfo{person}{Cristian
  Moldovan}.} \bibinfo{year}{2010}\natexlab{}.
\newblock \showarticletitle{The first question generation shared task
  evaluation challenge}. In \bibinfo{booktitle}{{\em INLG}}.
\newblock


\bibitem[\protect\citeauthoryear{Rush, Chopra, and Weston}{Rush
  et~al\mbox{.}}{2015}]%
        {rush2015neural}
\bibfield{author}{\bibinfo{person}{Alexander~M Rush}, \bibinfo{person}{Sumit
  Chopra}, {and} \bibinfo{person}{Jason Weston}.}
  \bibinfo{year}{2015}\natexlab{}.
\newblock \showarticletitle{A neural attention model for abstractive sentence
  summarization}. In \bibinfo{booktitle}{{\em EMNLP}}.
\newblock


\bibitem[\protect\citeauthoryear{Salton, Singhal, Mitra, and Buckley}{Salton
  et~al\mbox{.}}{1997}]%
        {salton1997automatic}
\bibfield{author}{\bibinfo{person}{Gerard Salton}, \bibinfo{person}{Amit
  Singhal}, \bibinfo{person}{Mandar Mitra}, {and} \bibinfo{person}{Chris
  Buckley}.} \bibinfo{year}{1997}\natexlab{}.
\newblock \showarticletitle{Automatic text structuring and summarization}.
\newblock \bibinfo{journal}{{\em Information Processing \& Management\/}}
  \bibinfo{volume}{33}, \bibinfo{number}{2} (\bibinfo{year}{1997}),
  \bibinfo{pages}{193--207}.
\newblock


\bibitem[\protect\citeauthoryear{Serban, Garc{\'\i}a-Dur{\'a}n, Gulcehre, Ahn,
  Chandar, Courville, and Bengio}{Serban et~al\mbox{.}}{2016}]%
        {serban2016generating}
\bibfield{author}{\bibinfo{person}{Iulian~Vlad Serban},
  \bibinfo{person}{Alberto Garc{\'\i}a-Dur{\'a}n}, \bibinfo{person}{Caglar
  Gulcehre}, \bibinfo{person}{Sungjin Ahn}, \bibinfo{person}{Sarath Chandar},
  \bibinfo{person}{Aaron Courville}, {and} \bibinfo{person}{Yoshua Bengio}.}
  \bibinfo{year}{2016}\natexlab{}.
\newblock \showarticletitle{Generating factoid questions with recurrent neural
  networks: The 30m factoid question-answer corpus}. In
  \bibinfo{booktitle}{{\em ACL}}.
\newblock


\bibitem[\protect\citeauthoryear{Sutskever, Vinyals, and Le}{Sutskever
  et~al\mbox{.}}{2014}]%
        {sutskever2014sequence}
\bibfield{author}{\bibinfo{person}{Ilya Sutskever}, \bibinfo{person}{Oriol
  Vinyals}, {and} \bibinfo{person}{Quoc~V Le}.}
  \bibinfo{year}{2014}\natexlab{}.
\newblock \showarticletitle{Sequence to sequence learning with neural
  networks}. In \bibinfo{booktitle}{{\em NIPS}}. \bibinfo{pages}{3104--3112}.
\newblock


\bibitem[\protect\citeauthoryear{Takase, Suzuki, Okazaki, Hirao, and
  Nagata}{Takase et~al\mbox{.}}{2016}]%
        {takase2016neural}
\bibfield{author}{\bibinfo{person}{Sho Takase}, \bibinfo{person}{Jun Suzuki},
  \bibinfo{person}{Naoaki Okazaki}, \bibinfo{person}{Tsutomu Hirao}, {and}
  \bibinfo{person}{Masaaki Nagata}.} \bibinfo{year}{2016}\natexlab{}.
\newblock \showarticletitle{Neural Headline Generation on Abstract Meaning
  Representation.}. In \bibinfo{booktitle}{{\em EMNLP}}.
  \bibinfo{pages}{1054--1059}.
\newblock


\bibitem[\protect\citeauthoryear{Tan, Wan, and Xiao}{Tan et~al\mbox{.}}{2017}]%
        {tanneural}
\bibfield{author}{\bibinfo{person}{Jiwei Tan}, \bibinfo{person}{Xiaojun Wan},
  {and} \bibinfo{person}{Jianguo Xiao}.} \bibinfo{year}{2017}\natexlab{}.
\newblock \showarticletitle{From Neural Sentence Summarization to Headline
  Generation: A Coarse-to-Fine Approach}. In \bibinfo{booktitle}{{\em IJCAI}}.
\newblock


\bibitem[\protect\citeauthoryear{Wang and Jiang}{Wang and Jiang}{2016}]%
        {wang2016machine}
\bibfield{author}{\bibinfo{person}{Shuohang Wang} {and} \bibinfo{person}{Jing
  Jiang}.} \bibinfo{year}{2016}\natexlab{}.
\newblock \showarticletitle{Machine comprehension using match-lstm and answer
  pointer}.
\newblock \bibinfo{journal}{{\em arXiv preprint arXiv:1608.07905\/}}
  (\bibinfo{year}{2016}).
\newblock


\bibitem[\protect\citeauthoryear{Xu, Yang, and Lau}{Xu et~al\mbox{.}}{2010}]%
        {xu2010keyword}
\bibfield{author}{\bibinfo{person}{Songhua Xu}, \bibinfo{person}{Shaohui Yang},
  {and} \bibinfo{person}{Francis Chi-Moon Lau}.}
  \bibinfo{year}{2010}\natexlab{}.
\newblock \showarticletitle{Keyword Extraction and Headline Generation Using
  Novel Word Features.}. In \bibinfo{booktitle}{{\em AAAI}}.
  \bibinfo{pages}{1461--1466}.
\newblock


\bibitem[\protect\citeauthoryear{Yang, Yang, Dyer, He, Smola, and Hovy}{Yang
  et~al\mbox{.}}{2016}]%
        {yang2016hierarchical}
\bibfield{author}{\bibinfo{person}{Zichao Yang}, \bibinfo{person}{Diyi Yang},
  \bibinfo{person}{Chris Dyer}, \bibinfo{person}{Xiaodong He},
  \bibinfo{person}{Alexander~J Smola}, {and} \bibinfo{person}{Eduard~H Hovy}.}
  \bibinfo{year}{2016}\natexlab{}.
\newblock \showarticletitle{Hierarchical Attention Networks for Document
  Classification.}. In \bibinfo{booktitle}{{\em HLT-NAACL}}.
  \bibinfo{pages}{1480--1489}.
\newblock


\end{thebibliography}

\end{document}